\pdfoutput=1
\documentclass[journal]{IEEEtran}
\usepackage{cite}
\usepackage{amsmath,amsfonts,amsthm}
\usepackage{array}
\usepackage{bm}
\usepackage{graphicx}
\usepackage{cleveref}
\usepackage{algorithm}
\usepackage{algpseudocode}
\usepackage{algorithmicx}
\usepackage{multirow}
\usepackage{bigstrut}
\usepackage{threeparttable}
\usepackage[table]{xcolor}
\usepackage[caption=false,font=footnotesize]{subfig}
\usepackage{placeins}
\usepackage{booktabs}
\usepackage{tabularx}
\usepackage{xr}
\usepackage{url}
\usepackage[switch,pagewise]{lineno}
\newtheorem{proposition}{Proposition}
\newtheorem{theorem}{Theorem}

\newtheorem{lemma}{Lemma}
\begin{document}
\crefname{figure}{fig.}{figs.}
\Crefname{figure}{Fig.}{Figs.}
\Crefname{equation}{Eq.}{Eqs.}
\renewcommand{\algorithmicrequire}{\textbf{Input:}}
\renewcommand{\algorithmicensure}{\textbf{Output:}}

\newcommand{\fSphere}{f_\text{Sphere}}
\newcommand{\fElli}{f_\text{Elli}}
\newcommand{\fRosen}{f_\text{Rosen}}
\newcommand{\fDiscus}{f_\text{Discus}}
\newcommand{\fCigar}{f_\text{Cigar}}
\newcommand{\fDiffPow}{f_\text{DiffPow}}
\newcommand{\fRotElli}{f_\text{RotElli}}
\newcommand{\fRotRosen}{f_\text{RotRosen}}
\newcommand{\fRotDiscus}{f_\text{RotDiscus}}
\newcommand{\fRotCigar}{f_\text{RotCigar}}
\newcommand{\fRotDiffPow}{f_\text{RotDiffPow}}

\algrenewcommand{\algorithmiccomment}[1]{\small \hskip2em// \emph{#1}}

\title{MMES: Mixture Model based Evolution Strategy for Large-Scale Optimization}

\author{Xiaoyu He, Zibin Zheng, Yuren Zhou
\thanks{This work was supported by the Key-Area Research and Development Program of Guangdong Province (2018B010109001), the National Natural Science Foundation of China (62006252 and 61773410), and the Guangdong Basic and Applied Basic Research Foundation (2019A1515111154).}
\thanks{Xiaoyu He, Zibin Zheng, and Yuren Zhou are with the School of Data and Computer Science, Sun Yat-sen University, Guangzhou 510006, P. R. China. (E-mail: hexy73@mail.sysu.edu.cn (X. He), zhzibin@mail.sysu.edu.cn (Z. Zheng), zhouyuren@mail.sysu.edu.cn (Y. Zhou))} 
\thanks{* Corresponding Author: Z. Zheng}
}
\maketitle

\newcommand{\tabincell}[2]{\begin{tabular}{@{}#1@{}}#2\end{tabular}}  
\begin{abstract}
This work provides an efficient sampling method for the covariance matrix adaptation evolution strategy (CMA-ES) in large-scale settings. In contract to the Gaussian sampling in CMA-ES, the proposed method generates mutation vectors from a mixture model, which facilitates exploiting the rich variable correlations of the problem landscape within a limited time budget. We analyze the probability distribution of this mixture model and show that it approximates the Gaussian distribution of CMA-ES with a controllable accuracy. We use this sampling method, coupled with a novel method for mutation strength adaptation, to formulate the mixture model based evolution strategy (MMES) -- a CMA-ES variant for large-scale optimization. The numerical simulations show that, while significantly reducing the time complexity of CMA-ES, MMES preserves the rotational invariance, is scalable to high dimensional problems, and is competitive against the state-of-the-arts in performing global optimization.
\end{abstract}

\IEEEpeerreviewmaketitle
\section{Introduction}

Evolution strategies (ESs)~\cite{beyer_evolution_2002} are a class of powerful evolutionary algorithms for black-box real-valued optimization. ESs sample candidate solutions from a multivariate Gaussian distribution and adapt the distribution to increase the likelihood of reproducing high-quality solutions. The covariance matrix adaptation ES (CMA-ES)~\cite{6790628,nikolaus_hansen_reducing_2003}, a modern ES implementation that resembles second-order methods, adapts the covariance matrix to the shape of the function landscape by learning the linear correlations among variables. In the ideal situation when all variable correlations are learned, the CMA-ES on any convex quadratic function behaves like a standard ES on a spheric function, which exhibits a linear convergence rate~\cite{arnold_weighted_2006}. CMA-ES also has the invariance against the rotational transformations on the decision space and the order-preserving transformations on the objective function. These properties make CMA-ES a popular black-box solver in many real-world applications such as artificial intelligence~\cite{abdolmaleki_deriving_2017,chen_restart-based_2019}, engineering design~\cite{mezura-montes_empirical_2008,gregory_fast_2011}, and automatic control~\cite{hansen_method_2009,shir_efficient_2014}.

The standard CMA-ES explicitly store all variable correlations in a covariance matrix of size $n\times n$, leading to a time complexity of $O(n^2)$ per generation and a space complexity of $O(n^2)$, where $n$ is the number of variables. Thus, CMA-ES is more time-/space-consuming than other typical evolutionary algorithms, which is usually regarded as the main limitation in large-scale optimization tasks. One pioneering work to address this issue is the separable CMA-ES (sep-CMA)~\cite{Ros:2008:SMC:1431377.1431409}. sep-CMA discards all off-diagonal entries of the covariance matrix and reduces the time and space complexity to $O(n)$. However, this makes sep-CMA incapable of handling variable correlations and causes significant performance degradation on non-separable functions. Later studies generally focus on seeking a trade-off between keeping only diagonal entries and keeping the full matrix. 

One popular method for balancing the performance and efficiency of CMA-ES in large-scale optimization is based on the covariance matrix reconstruction techniques. The basic idea is to model the covariance matrix with a set of $m$ vectors, which is much computationally cheaper than explicitly maintaining a full covariance matrix, provided that $m \ll n$. The direction vectors are designed to represent promising search directions, so the algorithms learn only a few variable correlations that contribute substantially to the optimization progress. Algorithms based on this method~\cite{loshchilov_lm-cma:_2017,loshchilov_large_2018,li_simple_2017,li_fast_2020,he_large-scale_2020} generally have competitive performance and preserve the rotational invariance, but only have a limited amount of computational burden.
There also exist works~\cite{akimoto_comparison-based_2014,akimoto_projection-based_2016} that hybridize this idea with the diagonal-learning strategy, aiming to further exploit the separability on relatively simple problems. 
Incorporating CMA-ES with the cooperative co-evolution (CC) framework~\cite{yang_large_2008} is another alternative for reducing the algorithm complexity. The CC based variants~\cite{liu_scaling_2013,mei_competitive_2016,tong_model_2019,sun_decomposition_2019} implicitly maintain a set of sub-blocks of the covariance matrix and can perform well on additive separable problems. However, they receive all deficiencies of the CC framework and, for example, cannot solve fully non-separable functions~\cite{ma_survey_2018}. 
Covariance matrix modeling techniques have also been studied in the context of estimation of distribution algorithms. In~\cite{kaban_toward_2016,sanyang_heavy_2015}, the high-dimensional covariance matrix is defined as an ensemble of low-dimensional covariance matrices that are estimated from randomly projected solutions, while in~\cite{dong_latent_2019} it is reconstructed from the principal components of high-quality population members. 
These algorithms preserve certain invariance properties like CMA-ES, yet their distributions are not iteratively updated but estimated directly from the solutions. Thus, they are not fully compatible with state-of-the-art cumulation mechanisms developed for the ES family.

This work concerns the covariance matrix reconstruction based ESs due to their desirable invariance properties and the robustness in handling different kinds of problems. 
Among them, we distinguish two different schemes for reconstructing the covariance matrix, according to whether all the $m$ direction vectors are used simultaneously.
\begin{itemize}
	\item The first scheme, used in the rank-$m$ ES (Rm-ES)~\cite{li_simple_2017}, the limited-memory matrix adaptation ES (LM-MA)~\cite{loshchilov_large_2018}, and the search direction adaptation ES (SDA-ES)~\cite{he_large-scale_2020}, is to utilize all direction vectors simultaneously in generating every solution. The solutions obtained definitely obey a multivariate Gaussian distribution determined by the mean vector and the covariance matrix. Thus, these algorithms possess a probability model of high interpretability. The setting $m \ll n$, however, becomes mandatory rather than optional, since the runtime of the sampling operations scales linearly with $m$. This deteriorates the performance, because the number of degrees of freedom in the reconstructed covariance matrix cannot exceed $nm$ and the algorithms fail to capture all variable correlations.
	\item The second scheme chooses only a subset of $l$ ones from all $m$ direction vectors, where $l$ may vary for different solutions. This scheme facilitates exploring the rich characteristics of the function landscape, as $m$ can be sufficiently large while the time complexity relies mainly on $l$ (which is usually set as $l \ll m$). The limited-memory CMA-ES (LM-CMA)~\cite{loshchilov_lm-cma:_2017} and the fast CMA-ES (fast-CMA)~\cite{li_fast_2020} adopt this scheme and show significant performance improvement. On the other hand, the reconstructed Gaussian distribution is actually conditioned on the specified $l$ vectors, so the true probability model is not Gaussian. Additionally, the used vectors are usually chosen heuristically and few studies derive an explicit form for the true probability model which guides the search. Thus, we do not obtain new insights into how these variants approximate the original CMA-ES.
\end{itemize}

Considering the properties of the above two schemes, one may wonder whether it is possible to combine the merits of them. 
Therefore, the first research question that motivates this study is:

\textit{\textbf{Q1: How to reconstruct a probability distribution, with a closed form expression, from an arbitrary number of direction vectors while keeping the sampling operations efficient?}}


This work proposes a fast mixture sampling method (FMS) as an answer to \textit{\textbf{Q1}}. FMS samples solutions from a Gaussian distribution whose covariance matrix is the regularized arithmetic mean of $l$ rank-1 matrices randomly constructed from a set of $m$ direction vectors. The solutions turn out to obey the well-known Gaussian mixture model that has a closed form expression. 
Then, one may take this model one step further and ask:


\textit{\textbf{Q2: How can the reconstructed probability distribution approximate the Gaussian distribution of the standard CMA-ES?}}

Surprisingly, we find when $l \rightarrow \infty$ the reconstructed probability model converges to a multivariate Gaussian distribution which approximates the one used in the standard CMA-ES. 
Due to this property, we call the parameter $l$ ``mixing strength'' and regard it as the most important parameter in further analysis. 
However, the infinity assumption on the mixing strength is too strong, so it is natural to ask how accurate can this approximation be when the mixing strength is finite. That is, we investigate: 

\textit{\textbf{Q3: To what extent can the reconstructed probability distribution, with a finite mixing strength, approximate the Gaussian distribution of the standard CMA-ES?}}

Our analysis shows that the probability distribution in FMS and a Gaussian distribution which approximates the one in CMA-ES only differ in high order statistical information. More precisely, their difference in the high order statistical moments decreases linearly when the mixing strength increases. This demonstrates the rationality of setting $l \ll m$ and the ability of FMS to explore a large amount of variable correlations within a limited time. 

To demonstrate the effectiveness of FMS, we incorporate it into the standard ES framework and design a mixture model based evolution strategy (MMES) to handle large-scale black-box optimization problems. We also propose a simple rule, called paired test adaptation (PTA), to adjust the mutation strength. The innovations of MMES are as follows:

\begin{itemize}
	\item MMES reconstructs the probability model from an arbitrarily large number of direction vectors without increasing the time complexity. 
	\item MMES has a nice theoretic property that its probability model converges to an approximation of the Gaussian distribution in the standard CMA-ES and the asymptotic error is inversely proportional to the mixing strength.
	\item MMES adapts the mutation strength in a way that relies only on the objective function and exhibits the so-called derandomization property.
\end{itemize}

In the remainder of this paper, we first provide the basic idea of FMS in Section II. The implementation details of MMES are given in Section III. Thereafter, we present the simulation results on two benchmark suites in Section IV. Finally, Section V concludes this paper and gives some remarks for future studies.

\section{Fast Mixture Sampling (FMS)}
\label{sec:FMS}
This section describes the FMS method and analyzes how and to what extent can its underlying probability model approximate that of the standard CMA-ES.
\subsection{Target Probability Distribution}
The early implementation of CMA-ES~\cite{6790628} adapts the covariance matrix using the so-called rank-1 update:
\begin{equation}
\begin{cases}
\bm{C}^{(0)} = \bm{I}_n \\
\bm{C}^{(g+1)} = (1-c_{cov}) \bm{C}^{(g)} + c_{cov} \bm{p}^{(g)}(\bm{p}^{(g)})^T
\end{cases}
\label{eq:covariance-recursively-rank-1-update}
\end{equation}
where $\bm{C}^{(g)}$ is the maintained covariance matrix, $c_{cov} \in[0,1]$ is the learning rate, $\bm{p}^{(g)} \in R^n$ is a random vector referred to as the evolution path, $\bm{I}_n$ is the $n$-dimensional identity matrix, and the superscript $g$ denotes the generation index. 
The recursive form of~\Cref{eq:covariance-recursively-rank-1-update} requires explicitly storing the full covariance matrix, which eventually becomes the performance bottleneck. By noticing that the evolution paths are smoothed with a very small decaying coefficient, one can approximate the covariance matrix efficiently in a non-recursive manner. Such an approximation takes the following form:
\begin{equation}
\bm{C}_{a} = (1-c_a)^m\bm{I} + c_a\sum_{j=1}^m (1-c_a)^{m-j} \bm{q}_j\bm{q}_j^T
\label{eq:covariance-non-recursively-rank-1-update}
\end{equation}
where the vectors $\bm{q}_1,\cdots,\bm{q}_m\in R^n$ are designed to approximate the evolution paths and $c_a$ is a new parameter analogous to the learning rate. 
Then, one can define an approximate Gaussian distribution 
\begin{equation*}
\mathcal{P}_a:\;\;\;\mathcal{N}(\bm{0}_n,\bm{C}_a) 
\label{eq:approximated-Gaussian-distribution}
\end{equation*}
to perturb the population mean and generate candidate solutions, where $\bm{0}_n$ denotes the $n$-dimensional zero vector.

$\mathcal{P}_a$ is the target probability distribution which FMS would approximate. It has been used in various papers~\cite{li_simple_2017,li_fast_2020,jagodzinski_differential_2017} and extensive numerical results suggest that using a larger $m$ leads to a better performance.
However, it takes $O(nm)$ time to sample a solution from $\mathcal{P}_a$ and so one has to keep $m$ sufficiently small in large-scale optimization, causing the dilemma on performance versus efficiency. The goal of this section is to derive an efficient method that samples solutions approximately obey $\mathcal{P}_a$ while allowing $m$ to be arbitrarily large.

\subsection{Working Procedure of FMS}
The proposed FMS method firstly generates a random vector $\bm{i}=(i_1,\cdots,i_l)^T \in \{1,\cdots,m\}^l$ with the probability distribution 
\begin{equation*}
\mathcal{P}_{\bm{i}}:\;\;\; p(i_j = k) = \alpha_k, j\in\{1,\cdots,l\}, k\in \{1,\cdots,m\} 
\end{equation*}
where $\alpha_1,\cdots,\alpha_m \in (0,1)$ and $\sum_{k=1}^m \alpha_k = 1$. 
Then, use $i_1, \cdots, i_l$ as indexes to select $l$ ones from the $\bm{q}$ vectors and construct the following symmetric matrix
\begin{equation}
\bm{\Sigma_i} = (1-\gamma) \bm{I}_n + \frac{\gamma}{l} \sum_{j=1}^l \bm{q}_{i_j} \bm{q}_{i_j}^T
\label{eq:conditioned-covariance-matrix}
\end{equation}
where $\gamma \in (0,1)$ is a regularization parameter for ensuring positive definiteness. Finally, define a multivariate Gaussian distribution with $\bm{\Sigma_i}$ being the covariance matrix 
\begin{equation*}
\mathcal{P}_{\bm{\Sigma_i}}:\;\;\; \mathcal{N}(\bm{0}_n,\bm{\Sigma_i}) 
\end{equation*}
and use this distribution to sample a mutation vector. 

Note that, the distribution~$\mathcal{P}_{\bm{\Sigma_i}}$ is actually conditioned on the chosen $\bm{q}$ vectors, and thus, the true probability model of FMS is a mixture of one multivariate isotropic Gaussian distribution and $m$ linear Gaussian distributions. For simplicity, denote this probability model by $\mathcal{P}_m$:
\begin{equation*}
\mathcal{P}_m:\;\;\; \text{\textit{The probability model reconstructed in FMS}.}
\end{equation*}
The following lemma provides a closed form expression for $\mathcal{P}_m$.

\begin{lemma}
The distribution $\mathcal{P}_m$ is uniquely determined by its moment generating function given by
\begin{equation}
M(\bm{t}) = \exp\left(\frac{1}{2}(1-\gamma)|\bm{t}|^2\right) \left(\sum_{j=1}^m \alpha_j \exp\left(\frac{\gamma}{2l}(\bm{t}^T \bm{q}_j)^2\right)\right)^l.
\end{equation}
\label{lemma:moment-generating-function-for-FMS}
\end{lemma}

On the one hand, \Cref{lemma:moment-generating-function-for-FMS} implies that $\mathcal{P}_m$ is directly relevant to all $\bm{p}$ vectors, independent of the specification of $\bm{i}$. On the other hand, FMS takes $O(ln)$ time per solution, since any mutation vector $\bm{z}$ sampled from~$\mathcal{P}_m$ can be written as
\begin{equation}
\bm{z} = \sqrt{1-\gamma} \bm{z}_0 + \sqrt{\frac{\gamma}{l}} \sum_{j=1}^l \bm{q}_{i_j} z_j,
\label{eq:sampled-mutation-vector}
\end{equation}
where $\bm{z}_0$ is sampled from $\mathcal{N}(\bm{0}_n,\bm{I}_n)$ and $z_1,\cdots,z_l$ are independent identically distributed (i.i.d.) random variables sampled from $\mathcal{N}(0,1)$. Thus, the time complexity of FMS is independent of $m$. By setting $l$ to a small number, the running time of the sampling operations can be significantly reduced. This gives the answer to \textit{\textbf{Q1}}. 

\subsection{Approximation Property of FMS}
\label{ss:approximation-property-of-FMS}
We show how $\mathcal{P}_m$ approximates $\mathcal{P}_a$. 
Before further analysis, we specify the parameters of $\mathcal{P}_{\bm{i}}$ as follows:

\begin{equation}
\begin{cases}
\gamma & = 1-(1-c_a)^m, \\
\alpha_k & = \frac{1}{\gamma} c_a(1-c_a)^{m-k}, k=1,\cdots,m. \\
\end{cases}
\label{eq:parameter-probability-i}
\end{equation}
This setting leads to a nice theoretic property:

\begin{theorem}
$\mathcal{P}_m$ converges to $\mathcal{P}_a$ when the mixing strength approaches infinity, provided that the parameters in \Cref{eq:parameter-probability-i} are used.
\label{theorem:FMS-convergence-to-gaussian}
\end{theorem}

\Cref{theorem:FMS-convergence-to-gaussian} provides an answer to \textbf{\textit{Q2}} by showing that $\mathcal{P}_a$ can be exactly recovered from $\mathcal{P}_m$ if the mixing strength is sufficiently large. This is the approximation property in the ideal case. The infinite mixing strength assumption is, however, too strong to be practically useful as it could make FMS even more expensive than directly sampling from $\mathcal{P}_a$. Fortunately, this assumption can be relaxed without significantly affecting the similarities between $\mathcal{P}_m$ and $\mathcal{P}_a$. This will be discussed in the next subsection.

\subsection{Approximation Accuracy Analysis}
\label{ss:approximation-accuracy-analysis}
We answer $\textbf{\textit{Q3}}$ and show, with a finite mixing strength, how accurate can the approximation be. 
Since \Cref{lemma:moment-generating-function-for-FMS} states that $\mathcal{P}_m$ can be completely characterized by its moment generating function, we perform the analysis by comparing the moments of $\mathcal{P}_m$ and $\mathcal{P}_a$.

Firstly, it is easy to conclude from the symmetry property that all odd order moments of $\mathcal{P}_m$ equal to 0, being coincident with $\mathcal{P}_a$. 

Moreover, the second order moments of $\mathcal{P}_m$ and $\mathcal{P}_a$ are closely related:
\begin{theorem}
$\mathcal{P}_m$ and $\mathcal{P}_a$ have the same covariance matrix given by \Cref{eq:covariance-non-recursively-rank-1-update}, provided that the parameters in \Cref{eq:parameter-probability-i} are used.
\label{theorem:second-order-moment-for-FMS}
\end{theorem}
Thus, $\mathcal{P}_m$ approximates $\mathcal{P}_a$ in the sense that they have the identical second order moment. The special instance of \Cref{theorem:second-order-moment-for-FMS} when $l=1$ coincides with a recent study~\cite{arabas_towards_2019}. However, \Cref{theorem:second-order-moment-for-FMS} further suggests that such a property holds for any mixture strength, and so, it is a more generic result.

The above approximation properties hold for any mixing strength, so we can infer that the mixing strength only affects the higher moments of even order. The following theorem gives a rigorous statement.

\begin{theorem}
Provided that the parameters in \Cref{eq:parameter-probability-i} are used.
The difference of $\mathcal{P}_m$ and $\mathcal{P}_a$ in the $k$-th order moments is on the order of $O(1/l)$, where $k=4,6,8,\cdots$.
\label{theorem:difference-in-higher-order-moments}
\end{theorem}

\Cref{theorem:difference-in-higher-order-moments} is critical in practice. It states that there is no need to choose a large mixing strength since the approximation accuracy, measured by the difference in high order statistical moments, increases at a linear rate when increasing the mixing strength. Therefore, it is reasonable to choose a small mixing strength to reduce the runtime without significantly deteriorating the performance. The numerical studies show that the setting $l=4$ works very well in various scenarios.

The following theorem facilitates an intuitive understanding of the characteristics of $\mathcal{P}_m$. 
\begin{theorem}
The projected distribution of $\mathcal{P}_m$ onto any one-dimensional subspace has a non-negative excess kurtosis. 
\label{theorem:non-negative-excess-kurtosis}
\end{theorem}

\Cref{theorem:non-negative-excess-kurtosis} does not rely on the parameters of $\mathcal{P}_{\bm{i}}$; but when the setting in~\Cref{eq:parameter-probability-i} is used, it implies that $\mathcal{P}_m$ has tails fatter than $\mathcal{P}_a$ and is more likely to produce outliers.
In this sense, $\mathcal{P}_m$ is analogous to the Cauchy distribution or the t-distribution which have been studied in the evolutionary algorithm literatures~\cite{rudolph_adaptive_2008-1,sanyang_heavy_2015}. Nevertheless, this would not mean $\mathcal{P}_m$ belongs to the class of heavy-tailed distributions as we can infer from \Cref{lemma:moment-generating-function-for-FMS} that its tails decay exponentially fast.

\subsection{Empirical Verification}
We conduct numerical simulations to verify the above theoretical analyses regarding the approximation accuracy. In the simulations, we firstly create $N_s = 10000$ samples from $\mathcal{P}_m$ and then compare their empirical distribution to the target distribution $\mathcal{P}_a$. The sample set is denoted by $\{\bm{y}_1,\cdots, \bm{y}_{N_s} | \bm{y}_i \in R^n\}$.
Parameters of the distributions are set as $m = 200, n = 1000$, and $c_a = 0.1$. The $\bm{q}$ vectors are constructed as $\bm{q}_i = (q_{i,1},\cdots,q_{i,n})^T$ where $q_{i,i} = 10^{3\times (i-1)/m}$ and $q_{i,j} = 0$ for all $j\neq i$ ($i\in\{1,\cdots,m\}, j\in\{1,\cdots,n\}$). This setting makes sure that both $\mathcal{P}_m$ and $\mathcal{P}_a$ can be decomposed into $n$ univariate ones, which facilitates the comparison and visualization of the high order statistical moments.

We propose two metrics to qualify how $\mathcal{P}_m$ differs from $\mathcal{P}_a$.
The first metric is the normalized variance ($NV$), calculated as
\begin{equation}
	NV = \frac{1}{n \cdot N_s} \sum_{i=1}^{N_s} \|\sqrt{\bm{C}_a^{-1}}\bm{y}_i\|^2.
	\label{eq:metric-nv}
\end{equation}
According to \Cref{theorem:second-order-moment-for-FMS}, using the matrix $\bm{C}_a$ to normalize the samples will lead to a unit diagonal covariance. Consequently, the $NV$ metric given in~\Cref{eq:metric-nv} must be close to 1, regardless of the mixing strength.

Our second metric is the standardized moment error ($SME$). It measures the difference between $\mathcal{P}_m$ and $\mathcal{P}_a$, in terms of higher order moments. For the $k$-th order moment, the metric (denoted by $SME_k$) is given by
\begin{equation}
	SME_k = \frac{1}{n \cdot N_s} \sum_{i=1}^{N_s} \sum_{j=1}^n \frac{y_{i,j}^k}{\delta_j^k}-\tau_k
\end{equation}
where $\tau_k$ denotes the $k$-th order moment of the standard normal distribution, $y_{i,j}$ denotes the $j$-th element of $\bm{y}_i$, and $\delta_j$ is the standard variance of the samples along the $j$-th axis direction.
Small absolute values of $SME_k$ (i.e., $|SME_k|$) indicate that $\mathcal{P}_m$ is close to $\mathcal{P}_a$ in terms of the $k$-th order moment. According to \Cref{theorem:FMS-convergence-to-gaussian,theorem:difference-in-higher-order-moments}, $|SME_k|$ should decrease to 0 at a rate of order $O(1/l)$.

\begin{figure}[tbp]
\centering
\subfloat[Averaged variance after normalization]{\includegraphics[width=0.35\textwidth]{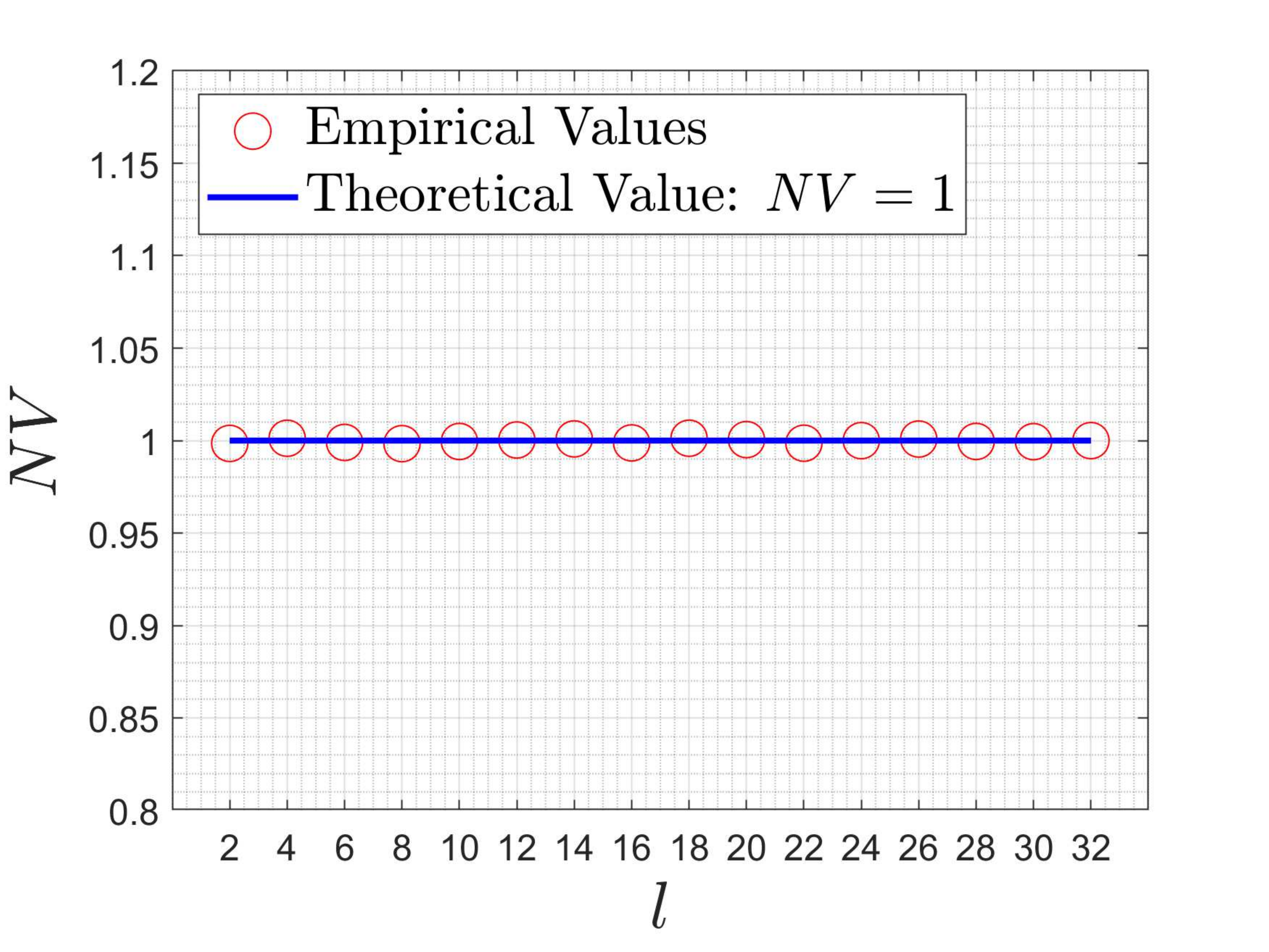} \label{subfig:nv} }
\\
\subfloat[Averaged approximation error of the 4-th order standardized moment]{\includegraphics[width=0.35\textwidth]{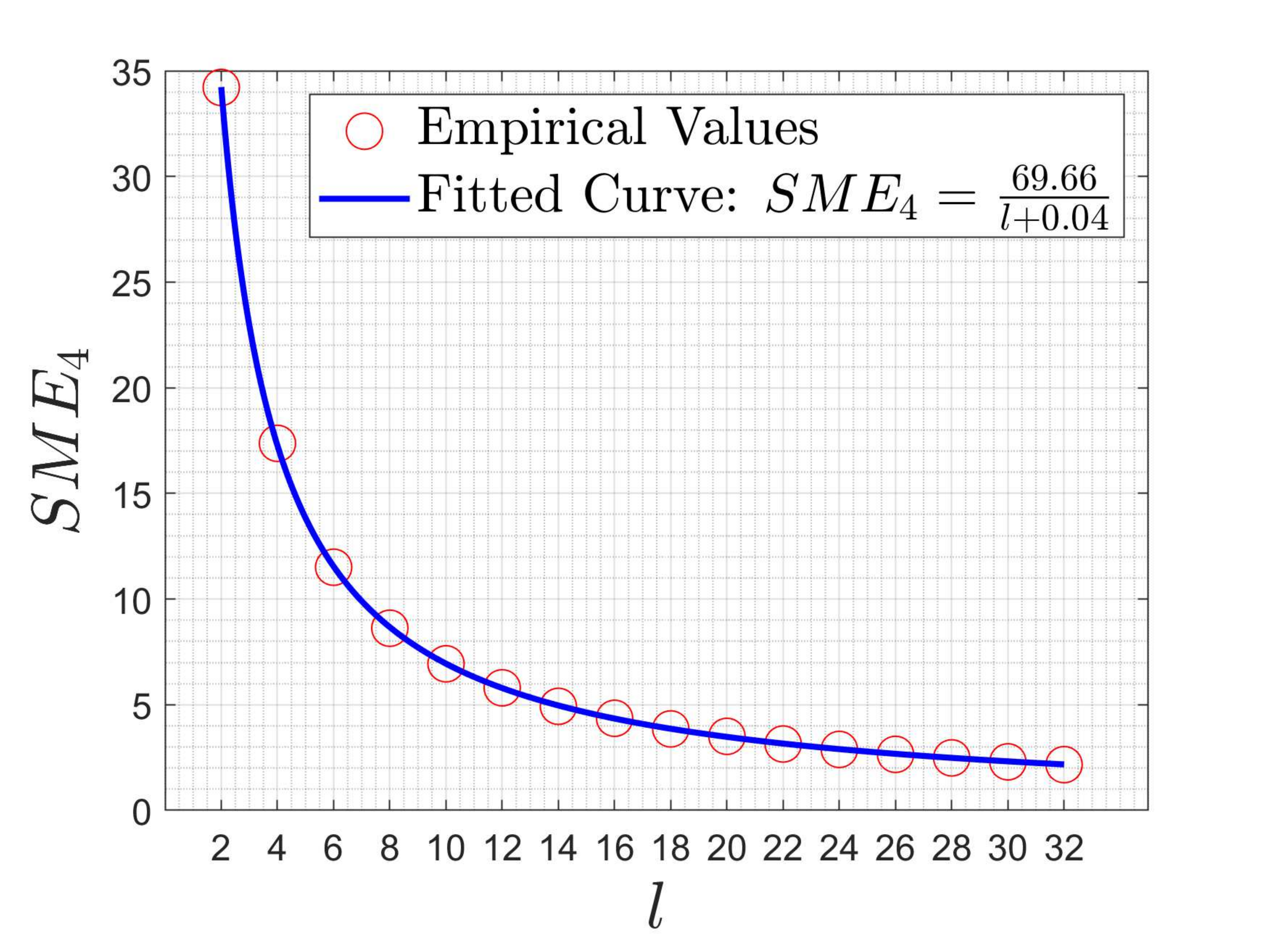} \label{subfig:m4} }
\\
\subfloat[Averaged approximation error of the 6-th order standardized moment]{\includegraphics[width=0.35\textwidth]{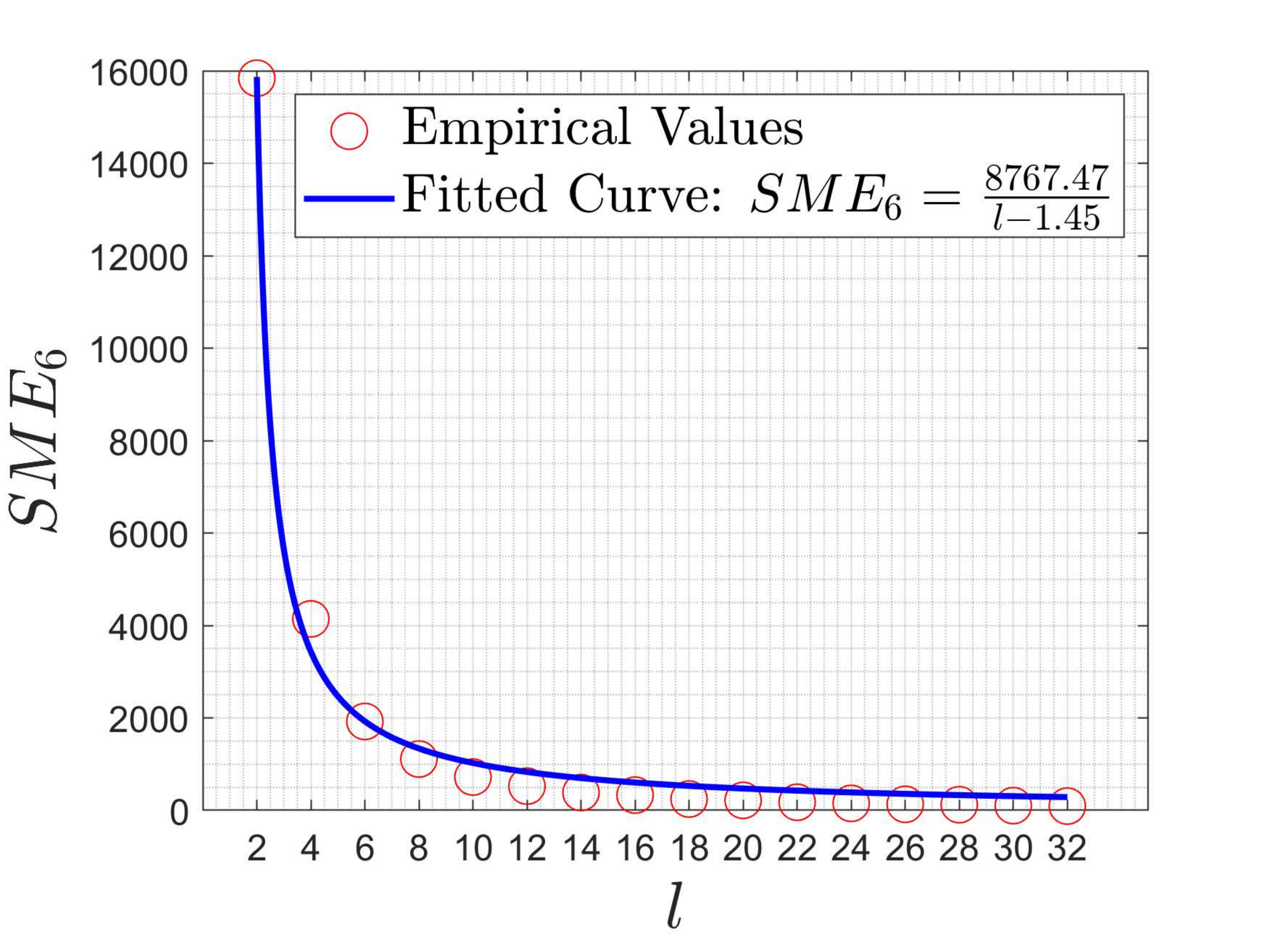} \label{subfig:m6} }
\caption{Influence of mixing strength on the approximation accuracy.}
\label{fig:empirical-verification-distribution}
\end{figure}

The simulations are performed with different mixing strengths chosen from $\{2,4,6,\cdots,32\}$. The corresponding results are plotted in \Cref{fig:empirical-verification-distribution}. From \Cref{subfig:nv}, it is found that all empirical results (depicted by red circles) are very close to the theoretical value (depicted by the blue line). Thus, $\mathcal{P}_m$ and $\mathcal{P}_a$ has similar covariance matrix, regardless of the mixing strength. This conclusion coincides with~\Cref{theorem:second-order-moment-for-FMS}.

Regarding the $SME$ metric, we focus on the cases $k=4$ and $k=6$. The results are respectively shown in \Cref{subfig:m4,subfig:m6}. In order to capture the asymptotic characteristics with different mixing strengths, we fit a rational function (shown by the blue line) using least squares regression. In \Cref{subfig:m4}, the regression curve $SME_4 = \frac{69.66}{l+0.04}$ perfectly fits all empirical results, implying that the asymptotic error given in \Cref{theorem:difference-in-higher-order-moments} is quite accurate in the 4-th order case. In \Cref{subfig:m6}, the curve $SME_6 = \frac{8786.47}{l-1.45}$ slightly overestimates the error for $l \ge 8$; but on the whole, it still exhibits the $O(1/l)$ decreasing rate, which coincides with \Cref{theorem:difference-in-higher-order-moments}. In addition, both \Cref{subfig:m4,subfig:m6} show that the $SME$ metric decreases rapidly when the mixing strength is relative small (say $l < 10$) but will approach 0 extremely slow. This supports our claim that the mixing strength needs not to be very large. 

\section{The Proposed Algorithm: MMES}
We develop the MMES algorithm by incorporating FMS into a generic $(\mu,\lambda)$-ES framework. MMES considers an unconstrained function $f:R^n \rightarrow R$ to be minimized by sampling in the $g$-th generation a population of $\lambda$ candidate solutions as
\begin{equation}
\bm{x}_i^{(g)} = \bm{m}^{(g)} + \sigma^{(g)} \bm{z}_i^{(g)}, i \in \{1,\cdots,\lambda\}
\end{equation}
where $\bm{m}^{(g)}$ is the mean vector, $\sigma^{(g)}$ is the mutation strength, and $\bm{z}_i^{(g)}$ is a mutation vector sampled from $\mathcal{P}_m$.
After evaluating the population, the best $\mu$ ones are chosen to update $\bm{m}^{(g)}$, $\sigma^{(g)}$, and $\mathcal{P}_m$. In the update, the $(\mu,\lambda)$-ES framework defines a set of weights and uses a weighting scheme to favor top ranked solutions.

\subsection{Paired Test based Mutation Strength Adaptation}
\label{ss:PTA}
Mutation strength adaptation enables ESs to achieve excellent convergence speed and reduce the adaptation time. The cumulative step-size control (CSA) in the standard CMA-ES is not applicable in MMES since it works in an implicit coordinate system defined by the covariance matrix. We present in this section an alternative to CSA, called paired test adaptation (PTA), which makes no assumption about the probability distribution and only relies on the objective function values.
An innovation of PTA is that it permits the preferences to high quality solutions and preserves the derandomization property of CSA. 

PTA is an improved version of the generalized 1/5-th success rule (GSR) in our previous work~\cite{he_large-scale_2020}. Similar to other success based methods, it adopts a simple rule: increasing the mutation strength if the optimization process is significantly successful, or decreasing it otherwise. The success is measured by the change of the objective function values evaluated in consecutive generations. 
The major distinction is that PTA uses a weighting scheme to put more importance to top ranked solutions.
Specifically, it computes a scalar $L^{(g)}$ at generation $g$ as:
\begin{equation}
L^{(g)} = \sum_{i=1}^\mu \omega_i \mathbb{I}[f(\bm{x}_{i:\lambda}^{(g-1)}) > f(\bm{x}_{i:\lambda}^{(g)})]
\end{equation}
where $\mathbb{I}[\cdot]$ is the indicator function, $\bm{x}_{i:\lambda}^{(g)}$ denotes the $i$-th best solution in the $g$-th generation, and $\omega_i$ is the weights, defined in the standard ES framework, satisfying $1 > \omega_1 > \cdots > \omega_\mu > 0$ and $\sum_{i=1}^\mu \omega_i = 1$.
The scalar $L^{(g)}$ calculates the weighted percentage of solution pairs that gain improvement of the objective values, and thus, serves as a success metric. 

To explicitly measure the success, we show that $L^{(g)}$ obeys a well-defined probability distribution under random selection, so the significance of success can be calculated from a properly designed statistical test. It is easy to see that $\mu L^{(g)}$ obeys a binomial distribution if setting $\omega_1=\cdots=\omega_\mu$. However, the case would be nontrivial when the weighting scheme is used. Fortunately, we find that with the default weight settings from the standard CMA-ES, $L^{(g)}$ converges in distribution to a Gaussian, as the population size approaches infinity.

\begin{proposition}
Set $\omega_i = \omega_i'/\sum_{j=1}^\mu \omega_j'$, $(i=1,\cdots,\mu)$, where $\omega_i' = \ln(\mu+0.5)-\ln(i)$. Assume that the solutions are randomly sorted. Then, as $\mu \rightarrow \infty$, 
\begin{equation}
L^{(g)} \xrightarrow{\text{d}} \mathcal{N}\left(\frac{1}{2},\frac{1}{4}\sum_{i=1}^\mu \omega_i^2\right).
\end{equation}
\label{proposition:distribution-of-L}
\end{proposition}

Inspired by this property, we design an exponential smoothing rule to improve the robustness of $L^{(g)}$ as

\begin{equation}
W^{(g)} = (1-c_\sigma) W^{(g-1)} + \sqrt{c_\sigma(2-c_\sigma)} \frac{2L^{(g)}-1}{\sqrt{\sum_{i=1}^\mu \omega_i^2}}
\label{eq:exponential-smoothing}
\end{equation}
where $c_\sigma\in(0,1)$ is a learning rate to average $L^{(g)}$ over generations. 
The second term on the right hand side is designed for normalization since we have $\frac{2L^{(g)}-1}{\sqrt{\sum_{i=1}^\mu \omega_i^2}}$ approximately obey $\mathcal{N}(0,1)$. Therefore, the smoother $W^{(g)}$ can be considered as an accumulated and normalized version of the success metric.
The update rule in \Cref{eq:exponential-smoothing} exhibits the derandomization property when \Cref{proposition:distribution-of-L} holds. On the one hand, the update is stable in the sense that if $W^{(g-1)} \sim \mathcal{N}(0,1)$ we also have $W^{(g)} \sim \mathcal{N}(0,1)$. On the other hand, the distribution of $W^{(g)}$ is independent of other parameters to be adapted. These coincide with the derandomization properties of CSA.

The metric $W^{(g)}$ characterizes the success of the optimization process, and in particular, $W^{(g)} > 0$ means that the algorithm generates more improving pairs than expected. Considering that it approximately obeys $\mathcal{N}(0,1)$ under random selection, we can verify whether the generation $g$ is successful with a simple $z$-test and use the result to update the mutation strength. This idea is inspired by GSR and is implemented as
\begin{equation}
\sigma^{(g+1)} = \sigma^{(g)} \exp\left(\frac{1}{d_\sigma}\left(\Phi(W^{(g)})-1+\alpha_z\right)\right)
\end{equation}
where $\Phi(\cdot)$ is the cumulative probability function of $\mathcal{N}(0,1)$, $\alpha_z$ is the significance level of the $z$-test, and $d_\sigma$ is a damping factor. When the success probability (i.e., $\Phi(W^{(g)})$) is smaller than $1-\alpha_z$, PTA decreases the mutation strength to encourage local search, which in turn helps to increase the success probability. On the contrary, when the success probability is greater than $1-\alpha_z$, PTA increases the mutation strength to enhance the global exploration ability, thereby causing a reduction in the success probability. By iteratively applying the above procedures, PTA keeps the significance of success at a $\alpha_z$ level.

\subsection{Direction Vector Adaptation and Selection}
\label{ss:direction-vector-adaptation-and-selection}
The direction vectors in reconstructing the probability model are designed to increase the likelihood of generating promising solutions. As the probability model reconstructed in FMS approximates the one in the standard CMA-ES, various well-established methods~\cite{li_simple_2017,loshchilov_lm-cma:_2017} can be applied without any modification. In this work, we adopt the one proposed in~\cite{loshchilov_lm-cma:_2017}. 
This method maintains a set of direction vectors $\bm{q}_1,\cdots,\bm{q}_m\in R^n$, a set of timestamps $t_1,\cdots,t_m \in Z^+$, and a set of logical indexes $v_1,\cdots,v_m\in\{1,\cdots,m\}$ in a way that $\bm{q}_{v_i}$ stores the evolution path generated in the $t_{v_i}$-th generation.
The adaptation starts by identifying a logic index $k^*$ given by $k^*=\underset{k\in\{2,\cdots,m\}}{\arg\min} t_{v_k} - t_{v_{k-1}}$.
The newly obtained evolution path (denoted by $\bm{p}$) then replaces $\bm{q}_{v_{k^*}}$ if $t_{v_k} - t_{v_{k-1}}$ is no larger than a threshold $T\in Z^+$, or replaces $\bm{q}_{v_1}$ otherwise. 
This preserves a certain distance, in terms of number of generations, between consecutive evolution paths to prevent the probability model from degeneracy.

When selecting random direction vectors, MMES generates the corresponding indexes in two steps. The first step is to produce a logic index $k$ from the distribution $\mathcal{P}_{\bm{i}}$. With the parameters in~\Cref{eq:parameter-probability-i}, it is equivalent to drawing the integer $m-k$ from a geometric distribution with the success probability $c_a$, conditioned on the range $[0,m-1]$. Then, the second step is to transform this index to the physical one. 
Combining these two steps together, a random index for selecting the $\bm{q}$ vectors can be calculated as
\begin{equation}
v_{m-j\%m}
\end{equation}
where $\%$ is the modulo operation and $j$ is a random integer sampled from a geometric distribution with the success probability $c_a$ (denoted by $\mathcal{G}(c_a)$). This procedure puts more importance to more recent evolution paths as they are logically stored with higher indexes.
\subsection{Detailed Implementation}
The pseudo-code of MMES is given in~\textbf{\Cref{alg:MMES}}\footnote{The source code is publically available at \url{https://github.com/hxyokokok/MMES}.}. 
In the initialization (Lines 2-6), all direction vectors are set to $\bm{0}_n$ such that the initial probability model is an isotropic Gaussian.
In the main loop, the FMS procedure is called in Lines 8-14 to generate the population. For each solution, it samples one $n$-dimensional isotropic Gaussian vector $\bm{z}_0$ and $l$ isotropic Gaussian scalars $z_1,\cdots,z_l$ in Lines 9 and 10. Then, $l$ indexes are generated in Lines 11 and 12 by sampling the geometric distribution $\mathcal{G}(c_a)$, followed by the transformation described~\Cref{ss:direction-vector-adaptation-and-selection}. Using these indexes, Line 13 chooses $l$ direction vectors to construct the mutation vector, according to \Cref{eq:sampled-mutation-vector}, with $\bm{z}_0$ and $z_1,\cdots,z_l$ being the mixture components. Finally, in Line 14, the mutation vector is rescaled and imposed on the population mean to get a candidate solution. 

Lines 16-18 are from the standard CMA-ES. All solutions are sorted according to their objective function values in Line 16. The new mean is recombined from the best $\mu$ solutions in Line 17, with a rank-based weighting strategy to favor high quality solutions. Line 18 updates the evolution path by cumulating the mutation step of the population mean, where $c_c$ is the learning rate. 

Lines 19-26 adapt the direction vectors as well as their indexes and timestamps according to~\cite{loshchilov_lm-cma:_2017}. It chooses the $k^*$-th vector having the minimal distance to the previous one (Line 19), removes it from the vector record (Lines 23 and 24), and then appends the evolution path to the record (Lines 25 and 26). The index $k^*$ is reset to 1 to drop the earliest information whenever no pair of consecutive vectors has a distance smaller than $T$ (Lines 20-22). 

In the end of the loop, Lines 27-29 implement the PTA rule. The implementation is the same as described in~\Cref{ss:PTA} except that the term $1/\sqrt{\sum_{i=1}^\mu \omega_i^2}$ in~\Cref{eq:exponential-smoothing} is replaced by $\mu_{eff}$, by convention of the standard CMA-ES. 

\begin{figure}[tb]
\begin{algorithm}[H]
	\caption{\emph{MMES}}
	\small
	\label{alg:MMES}
	\begin{algorithmic}[1]
	\Require $\bm{m}^{(0)}$: mean vector; $\sigma^{(0)}$: mutation strength; 
	\State $g = 0$
	\State $\bm{p}^{(0)} = \bm{0}_n$ 
	\State $W^{(0)} = 0$ 
	\State $\bm{q}_i^{(0)} = \bm{0}_n$ for $i=1,\cdots,m$
	\State $t_i=0$ for $i=1,\cdots,m$
	\State $v_i=i$ for $i=1,\cdots,m$
	\While {the termination criterion is not met}
		\For {$i = 1 \text{ to } \lambda$} 
			\State $\bm{z}_0\sim \mathcal{N}(\bm{0}_n,\bm{I}_n)$
			\State $z_1,\cdots,z_l\sim \mathcal{N}(0,1)$
			\State $j_1,\cdots,j_l \sim \mathcal{G}(c_a)$
			\State $j_k=v_{m-j_k\%m}$ for $k=1,\cdots,m$
			\State $\bm{z} = \sqrt{1-\gamma}\bm{z}_0 + \sqrt{\frac{\gamma}{l}} \sum_{k=1}^l z_k \bm{q}_{j_k}$
		\State $\bm{x}_i^{(g)} = \bm{m}^{(g)} + \sigma^{(g)} \bm{z}$
		\EndFor
		\State Sort the solutions such that $f(\bm{x}_{1:\lambda}^{(g)}) \le \cdots \le f(\bm{x}_{\lambda:\lambda}^{(g)})$
		\State $\bm{m}^{(g+1)} = \sum_{j=1}^{\mu} \omega_j \bm{x}_{j:\lambda}^{(g)}$ 
	    \State $\bm{p}^{(g+1)} = (1-c_c)\bm{p}^{(g)} + \sqrt{c_c(2-c_c)\mu_{eff}} \frac{\bm{m}^{(g+1)} - \bm{m}^{(g)}}{\sigma^{(g)}}$
		\State $k^*=\underset{k\in\{2,\cdots,m\}}{\arg\min} t_{v_k} - t_{v_{k-1}}$
		\If {$t_{v_k} - t_{v_{k-1}} \ge T$}
			\State $k^*=1$
		\EndIf
    	\State $v_k = v_{k+1}$ for $k=k^*,\cdots,m-1$
    	\State $v_m = v_{k^*}$ 
    	\State $t_{v_m} = g+1$
    	\State $\bm{q}_{v_m} = \bm{p}^{(g+1)}$

	    \State $L = \sum_{i=1}^\mu \omega_i \mathbb{I}[f(\bm{x}_{i:\lambda}^{(g-1)}) > f(\bm{x}_{i:\lambda}^{(g)})]$

		\State $W^{(g+1)} = (1-c_\sigma) W^{(g)} + \sqrt{c_\sigma(2-c_\sigma)\mu_{eff}} (2L-1)$
		\State $\sigma^{(g+1)} = \sigma^{(g)} \exp\left(\frac{1}{d_\sigma}\left(\Phi(W^{(g)})-1+\alpha_z\right)\right)$
		\State $g=g+1$
	\EndWhile
  \\ \Return 
 \end{algorithmic} 
 \end{algorithm}
\end{figure}

\subsection{Parameters}
All parameters of MMES are summarized in~\Cref{tab:default-parameters}. $\lambda,\mu,\omega_1,\cdots,\omega_\mu, \mu_{eff}$, and $c_c$ are from the standard CMA-ES, so they are set as suggested in~\cite{hansen_cma_2016}. Other parameters are discussed as below:
\begin{itemize}
\item $c_a$ is the success probability of the geometric distribution in choosing the direction vectors. \Cref{theorem:second-order-moment-for-FMS} states that it serves as a learning rate for the covariance matrix of the reconstructed probability model. Since a standard ES requires $O(n)$ expected time for a fixed relative improvement on a spherical function~\cite{arnold_performance_2004}, setting $c_a \ge O(1/n)$ seems to be necessary to obtain the same convergence speed for MMES on a convex quadratic function. Therefore, we use the setting $c_a = 4/n$.

\item $T$ is the minimal distance between the consecutive evolution paths maintained by MMES. As suggested in~\cite{li_simple_2017}, we set $T=\lceil 1/c_c\rceil$ which tends to keep the evolution paths uncorrelated.

\item $\gamma$ is the regularization coefficient in building the conditional Gaussian distribution $\mathcal{P}_{\bm{\Sigma_i}}$. We choose the setting $\gamma = 1 - (1-c_a)^m$ since it leads to the theoretical properties discussed in~\Cref{ss:approximation-property-of-FMS}.

\item $m$ is the number of candidate direction vectors. It constrains the number of variable correlations that can be learned. In this work, we set $m=2\lceil \sqrt{n} \rceil$. This is relatively larger than the usual settings in many CMA-ES variants, but is still affordable in most scenarios where evolutionary algorithms are applied.

\item $l$ is the mixing strength affecting the approximation accuracy of the probability model. 
As analyzed in~\Cref{ss:approximation-accuracy-analysis}, a small value is enough to attain considerable accuracy. Thus, we recommend setting $l$ to $O(1)$. In this work, we choose $l=4$ and the numerical experiments show that it works well in various optimization tasks.

\item $c_\sigma$ is the decay factor of the exponential smoothing for $W^{(g)}$ and $d_\sigma$ is the corresponding damping constant. They both control the changing rate of the mutation strength. As the distribution of $W^{(g)}$ is independent of the other parameters (see~\Cref{ss:PTA}), it is reasonable to set both $c_\sigma$ and $d_\sigma$ to $O(1)$. We choose the setting $c_\sigma=0.3, d_\sigma=1$ from our previous study~\cite{he_large-scale_2020} in which the same smoothing rule is adopted.

\item $\alpha_z$ is the target significance level in PTA. It is essentially a parameter of the $z$-test rather than of MMES itself. We set $\alpha_z$ to 0.05, the most popular setting in statistical tests.
\end{itemize}

\begin{table}[tbp]
\centering
\caption{Parameters for MMES}
\renewcommand\arraystretch{1}
\label{tab:default-parameters}
\begin{tabular}{l}
\toprule
$\lambda = 4+\lfloor3 \ln n\rfloor$, \; $\mu = \lfloor \frac{\lambda}{2}\rfloor$, \; $m=2\lceil \sqrt{n} \rceil$, \; $c_a = \frac{4}{n}$, \\
$c_c = \frac{0.4}{\sqrt{n}}$, \; $T = \lceil \frac{1}{c_c} \rceil$, \; $\gamma = 1 - (1-c_a)^m$, \\
$\omega_i=\frac{\ln(\mu+0.5)-\ln(i)}{\mu \ln(\mu+0.5) - \sum_{j=1}^\mu \ln(i)}$, \; $\mu_{eff} = \frac{1}{\sum_{i=1}^\mu \omega_i^2}$, \\
$c_\sigma = 0.3$, \; $d_\sigma=1$, \; $\alpha_z = 0.05$. \\
\bottomrule
\end{tabular}%
\end{table}%

\subsection{Complexity}
\label{ss:complexity-analysis}
MMES stores $\lambda$ solutions and $m$ direction vectors and so its space complexity is $O(n^{1.5})$.
The most time-consuming steps of MMES are in the FMS procedure. Precisely, it requires $O(n)$ operations for sampling the multivariate isotropic Gaussian vector in Line 9, $O(ln)$ operations for constructing the mutation vector in Line 13, and $O(n)$ operations for calculating the candidate solution in Line 14. All other lines can be performed in $O(n)$ time. Thus, when adopting the parameters in~\Cref{tab:default-parameters}, the time complexity of MMES is $O(n)$ per solution.

\section{Comparative Studies}
We investigate the performance of MMES and other eleven large-scale evolutionary algorithms on two benchmark sets. 
\subsection{Experimental Settings}
\subsubsection{Test Problems}
Two sets of test problems are selected in the numerical study. Their objective functions and brief descriptions are summarized in~\Cref{tab:test-problems}. 

\begin{table}[tbp]
\centering
\caption{Test Problems}
\renewcommand\arraystretch{1}
\setlength{\tabcolsep}{2pt}
\label{tab:test-problems}
\begin{threeparttable}
\begin{tabular}{ll}
\toprule
\multicolumn{2}{c}{Set 1: Basic Test Problems} \\
\midrule
Name & Function \\
\midrule
Sphere & \scriptsize $\fSphere(\bm{x}) = \sum_{i=1}^n x_i^2$ \\
Ellipsoid & \scriptsize $\fElli(\bm{x}) = \sum_{i=1}^n 10^{\alpha\frac{i-1}{n-1}}x_i^2$ \\
Rosenbrock & \scriptsize $\fRosen(\bm{x}) = \sum_{i=1}^{n-1}(100(x_i^2-x_{i+1})^2+(x_i-1)^2)$ \\
Discus & \scriptsize $\fDiscus(\bm{x}) = 10^6x_1^2+\sum_{i=2}^n x_i^2$ \\
Cigar & \scriptsize $\fCigar(\bm{x}) = x_1^2 + 10^6\sum_{i=2}^n x_i^2$ \\
Different Powers & \scriptsize $\fDiffPow = \sum_{i=1}^n |x_i|^{2+4\frac{i-1}{n-1}}$ \\ 
Rotated Ellipsoid & \scriptsize $\fRotElli(\bm{x}) = \fElli(\bm{R}\bm{x})$ \\
Rotated Rosenbrock & \scriptsize $\fRotRosen(\bm{x}) = \fRosen(\bm{R}\bm{x})$ \\
Rotated Discus & \scriptsize $\fRotDiscus(\bm{x}) = \fDiscus(\bm{R}\bm{x})$ \\
Rotated Cigar & \scriptsize $\fRotCigar(\bm{x}) = \fCigar(\bm{R}\bm{x})$ \\
Rotated Different Powers & \scriptsize $\fRotDiffPow(\bm{x}) = \fDiffPow(\bm{R}\bm{x})$ \\
\midrule
\midrule
\multicolumn{2}{c}{Set 2: CEC'2010 LSGO Problems} \\
\midrule
Function & Description \\
\midrule
$f_{1}$ to $f_{3}$ & Fully separable \\
$f_{4}$ to $f_{8}$ & Non-separable in a single group \\
$f_{9}$ to $f_{13}$ & Non-separable in 10 groups \\
$f_{14}$ to $f_{18}$ & Non-separable in all 20 groups \\
$f_{19}$ to $f_{20}$ & Fully non-separable \\
\bottomrule
\end{tabular}%
\begin{tablenotes}
\item[1] $\bm{R}$ is an orthogonal matrix generated by applying the Gram-Schmidt procedure on a random matrix with standard normally distributed entries.
\item[2] The condition number of $\fElli$ is customizable by tuning the parameter $\alpha$. Unless stated otherwise, we choose $\alpha = 6$ to render the problem landscape ill-conditioned.
\end{tablenotes}
\end{threeparttable}
\end{table}

\paragraph{Set 1: Basic Test Problems}
The first test set is intended to test the algorithms' rotational invariance and scalability, the most favorable properties in designing new ESs. It contains 11 basic test problems, all of which have the global minimum 0. $\fSphere$ is the simplest and serves as a base in performance analysis. $\fElli$, $\fRosen$, $\fDiscus$, $\fCigar$, and $\fDiffPow$ are separable or only have very weak variable correlations. Based on them, another five fully non-separable problems, namely $\fRotElli$, $\fRotRosen$, $\fRotDiscus$, $\fRotCigar$, and $\fRotDiffPow$, are constructed respectively by imposing a rotational transformation on the decision space. These problems differ in the landscape characteristics, and hence, mainly challenge the ESs' adaptation ability in different optimization tasks. For example, the Rosenbrock problem and the Cigar problem have a low-rank structured landscape in the sense that there exists a significant eigengap in the Hessian such that the landscape shape can be well approximated by only a few direction vectors. Thus, ESs based on the probability model reconstruction techniques are expected to solve these problems easily, provided that the used direction vectors are appropriately adapted. On the contrary, the Ellipsoid, the Discus, and the Different Powers problems do not have this property, and they are likely to cause considerable difficulties to the ESs chosen in the comparative study. 

\paragraph{Set 2: CEC'2010 LSGO Problems}
The second set is the benchmark suite for the CEC'2010 competition on large-scale global optimization (LSGO). It contains 20 test problems and covers a variety of difficulties such as multimodality, non-separability, and boundary constraints. We choose this set to verify the overall performance of the algorithms. A distinct feature of this test set is that its non-separability can be explicitly controlled. Specifically, $f_1$ to $f_3$ are fully separable, $f_{19}$ to $f_{20}$ are fully non-separable, and the rest are partially non-separable. In the partially non-separable problems, the variables are uniformly divided into 20 groups, some of which are chosen and then made non-separable, such that the variable correlations only exist in certain groups and there are no correlations between groups or in unchosen groups. For $f_4$ to $f_8$, $f_9$ to $f_{13}$, and $f_{14}$ to $f_{18}$, the numbers of non-separable groups are set to 1, 10, and 20, respectively. For their detailed definitions and properties, please refer to~\cite{tang_benchmark_2009}.



\subsubsection{Algorithms for Comparison}



CMA-ES~\cite{nikolaus_hansen_reducing_2003} and its five large-scale variants, namely sep-CMA~\cite{Ros:2008:SMC:1431377.1431409}, LM-CMA~\cite{loshchilov_lm-cma:_2017}, LM-MA~\cite{loshchilov_large_2018}, Rm-ES~\cite{li_simple_2017}, SDA-ES~\cite{he_large-scale_2020} are used in the first test set. In all these algorithms, as well as in MMES, the initial mean vector is uniformly sampled in the range $[-5, 5]^n$ and the initial mutation strength is set to 3. 
We benchmark these algorithms with their default parameter settings.

Another five large-scale evolutionary algorithms that are not based on the ES framework are used in the second test set. They include DECC-G~\cite{yang_large_2008}, MA-SW~\cite{molina_memetic_2011}, MOS~\cite{latorre_large_2013}, CCPSO2~\cite{li_cooperatively_2012}, and DECC-DG~\cite{omidvar_cooperative_2014}. DECC-G, DECC-DG, and CCPSO2 are CC-based algorithms and are chosen to show the advantages and disadvantages of the ES framework over the CC framework. MA-SW and MOS are state-of-the-art memetic algorithms. The former is the winner of the CEC'2010 LSGO competition while the latter wins all the CEC competitions hold during 2013-2017. 
For the above five competitors, the results are directly from~\cite{li_simple_2017,latorre_comprehensive_2015,yang_level-based_2017,he_large-scale_2020}, measured in the standard settings for the CEC competitions.

\subsubsection{Performance Metrics}
The experiments on the first test set mainly concern the convergence ability of the algorithms. Thus, the number of function evaluations required to converge (denoted by $FEs$) is utilized as the performance indicator. We consider an algorithm as converged if it finds an objective function value smaller than $10^{-8}$ before $FEs$ reaches a pre-defined threshold $maxFEs$. For those fail to converge, their $FEs$ values are directly set to $maxFEs$. $maxFEs$ is set to $10^8$ for $n=1000$ and $2\times 10^8$ for $n>1000$. All algorithms are independently run 20 times and the median results are reported. 

For the experiments on the second test set, the performance of algorithms are measured by the best objective function values found within a computational budget of $3\times 10^6$ function evaluations. 
MMES is independently run 25 times, as recommended for the CEC competition. 

On each test instance, we use the Wilcoxon rank sum test~\cite{wilcoxon1945individual} to verify whether the results of MMES and those of the others are significantly different. For MA-SW, MOS, CCPSO, and DECC-DG where only the median results are available, the Wilcoxon signed rank test is used instead. To have an overall view of the performance in a certain set of test instances, we calculate for each algorithm the rank averaged over all test instances, according to the Friedman test~\cite{derrac_practical_2011}. The significance of difference between MMES and the other competitors are reported, with a correction by the Bonferroni procedure~\cite{dunn_multiple_1961} to eliminate the family-wise error.

\subsection{Effectiveness of PTA}

We verify the potential of PTA as an alternative to CSA and other methods for mutation strategy adaptation. To this end, we compare PTA with the CSA from LM-MA, the population success rule (PSR) from LM-CMA, the rank success rule (RSR) from Rm-ES, and the GSR from SDA-ES. For a fair comparison, these methods are extracted from the corresponding algorithms and incorporated into the standard ES. 

\begin{figure}[tbp]
\centering
\subfloat[$\fSphere$]{\includegraphics[width=0.24\textwidth]{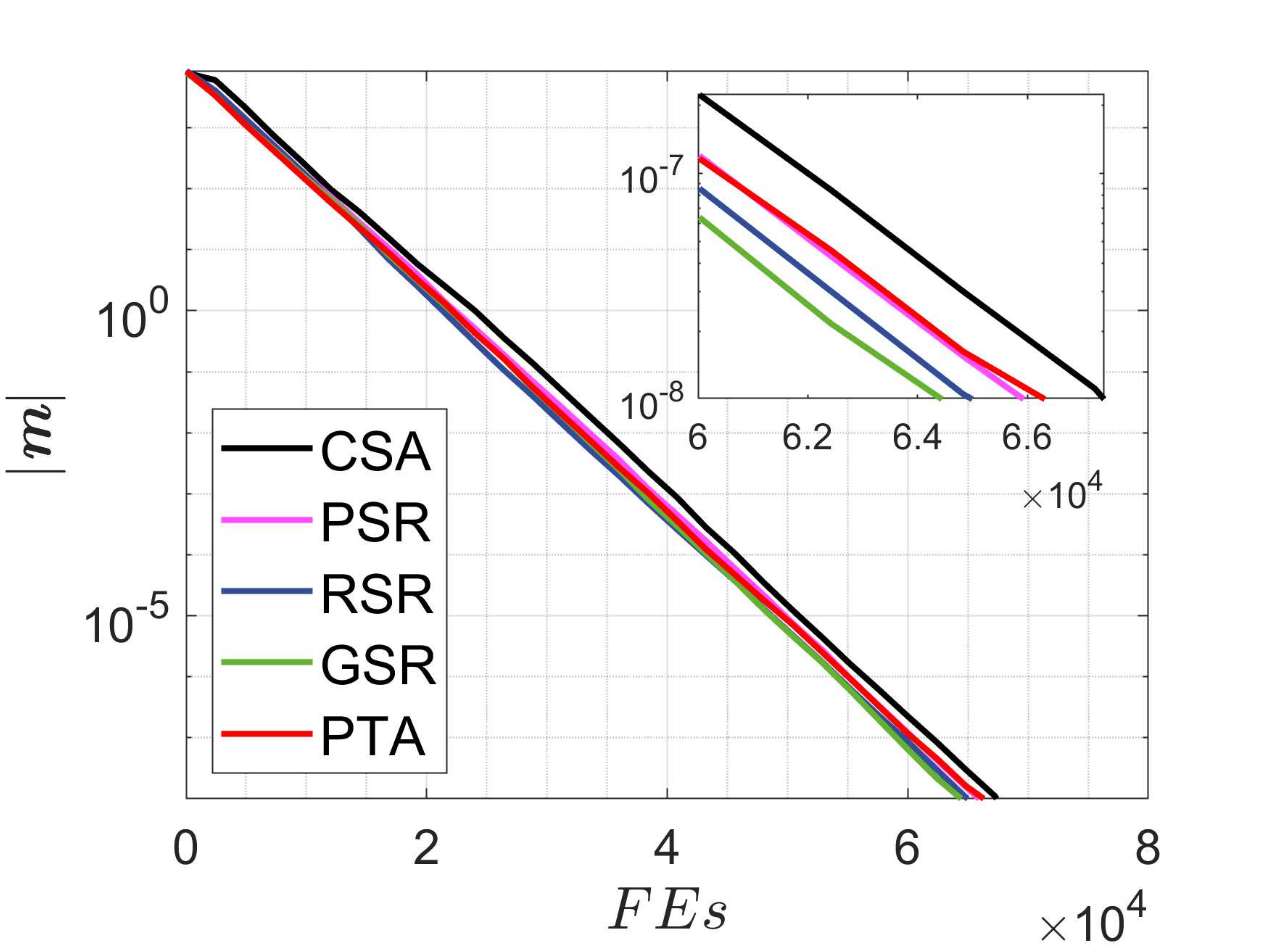} \label{subfig:1000D-Sphere} }
\subfloat[$\fElli, \alpha = 1$]{\includegraphics[width=0.24\textwidth]{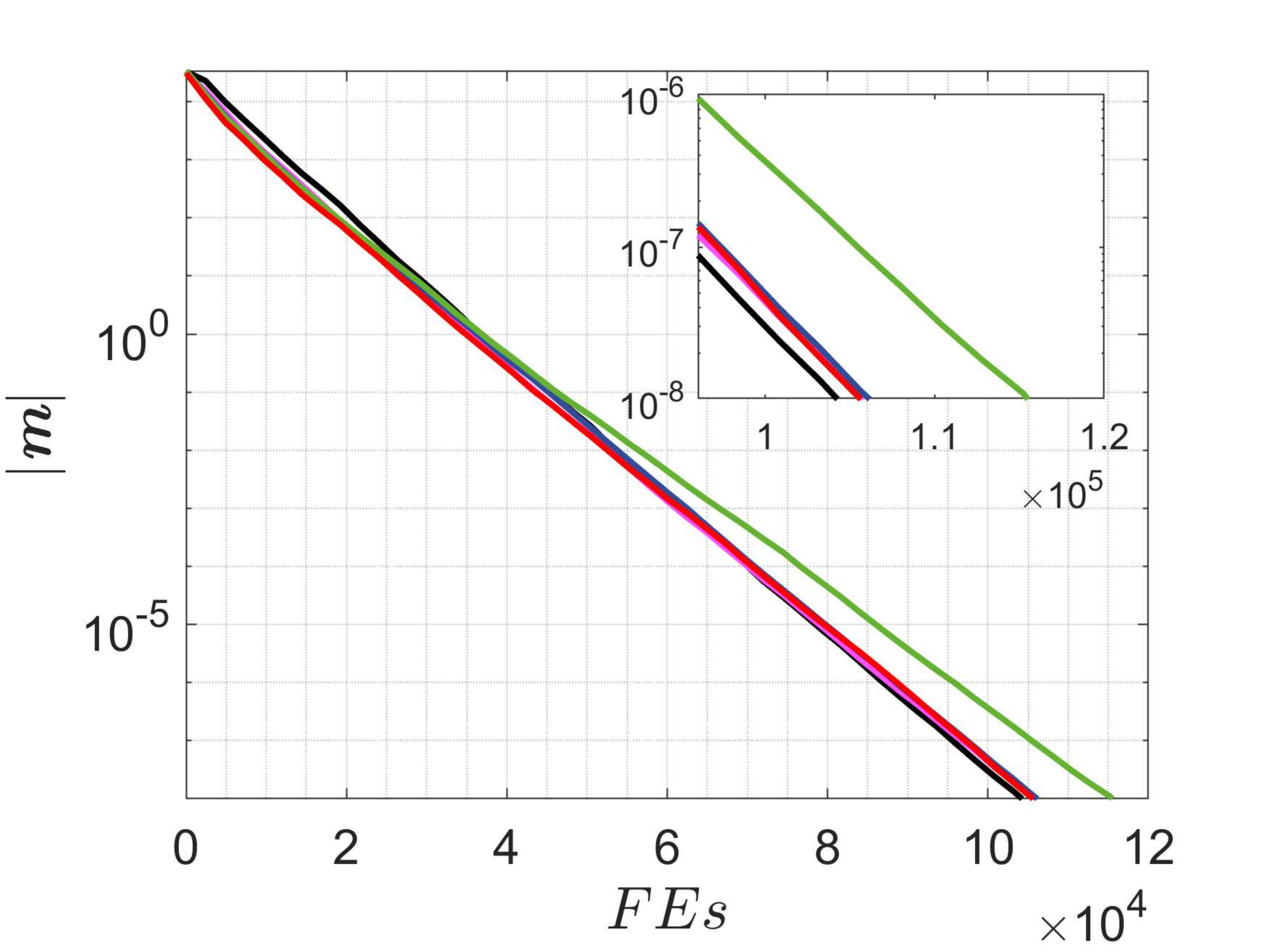} \label{subfig:1000D-Elli-cn1} }
\hfil
\subfloat[$\fElli, \alpha = 2$]{\includegraphics[width=0.24\textwidth]{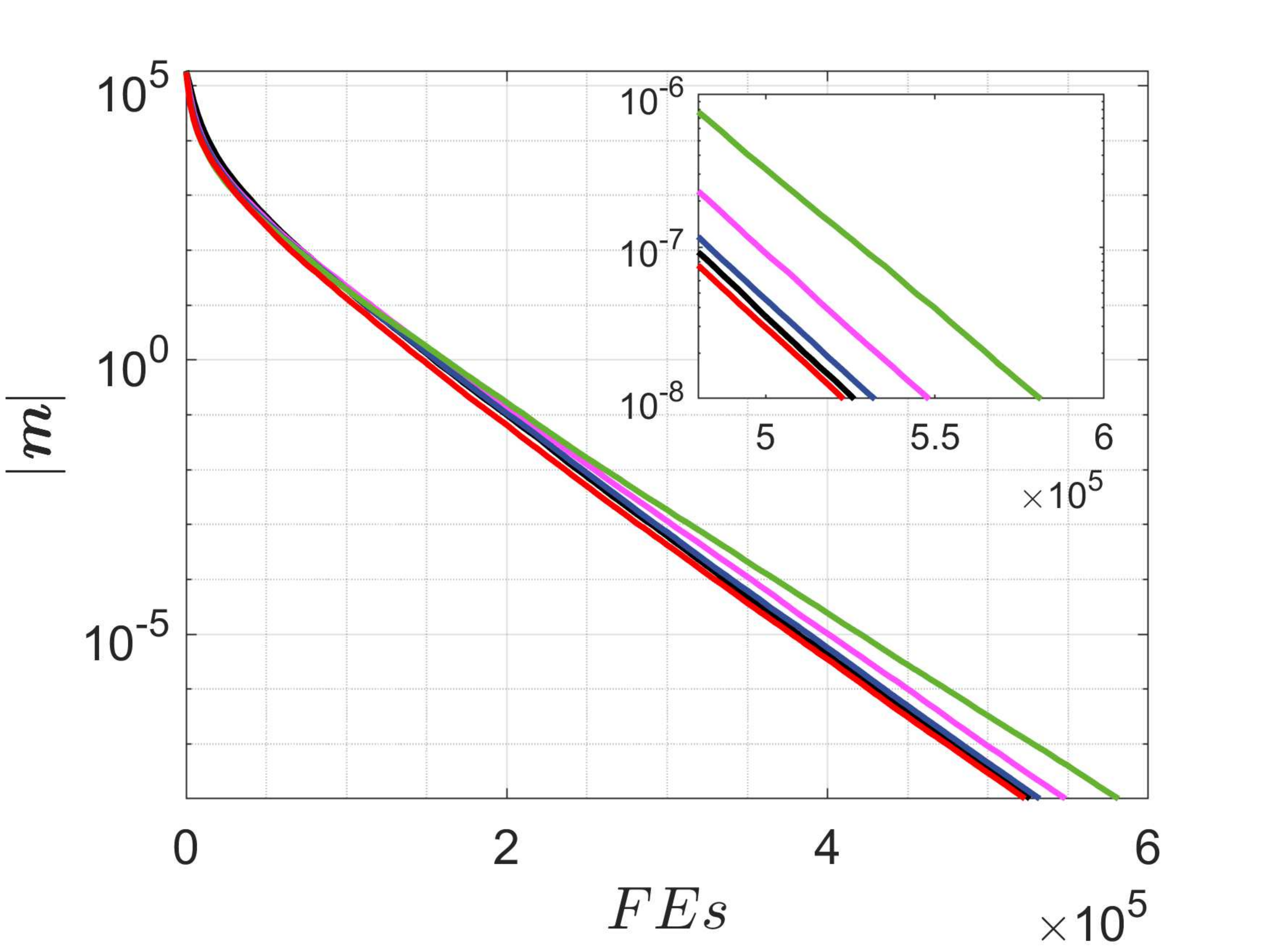} \label{subfig:1000D-Elli-cn2} }
\subfloat[$\fElli, \alpha = 3$]{\includegraphics[width=0.24\textwidth]{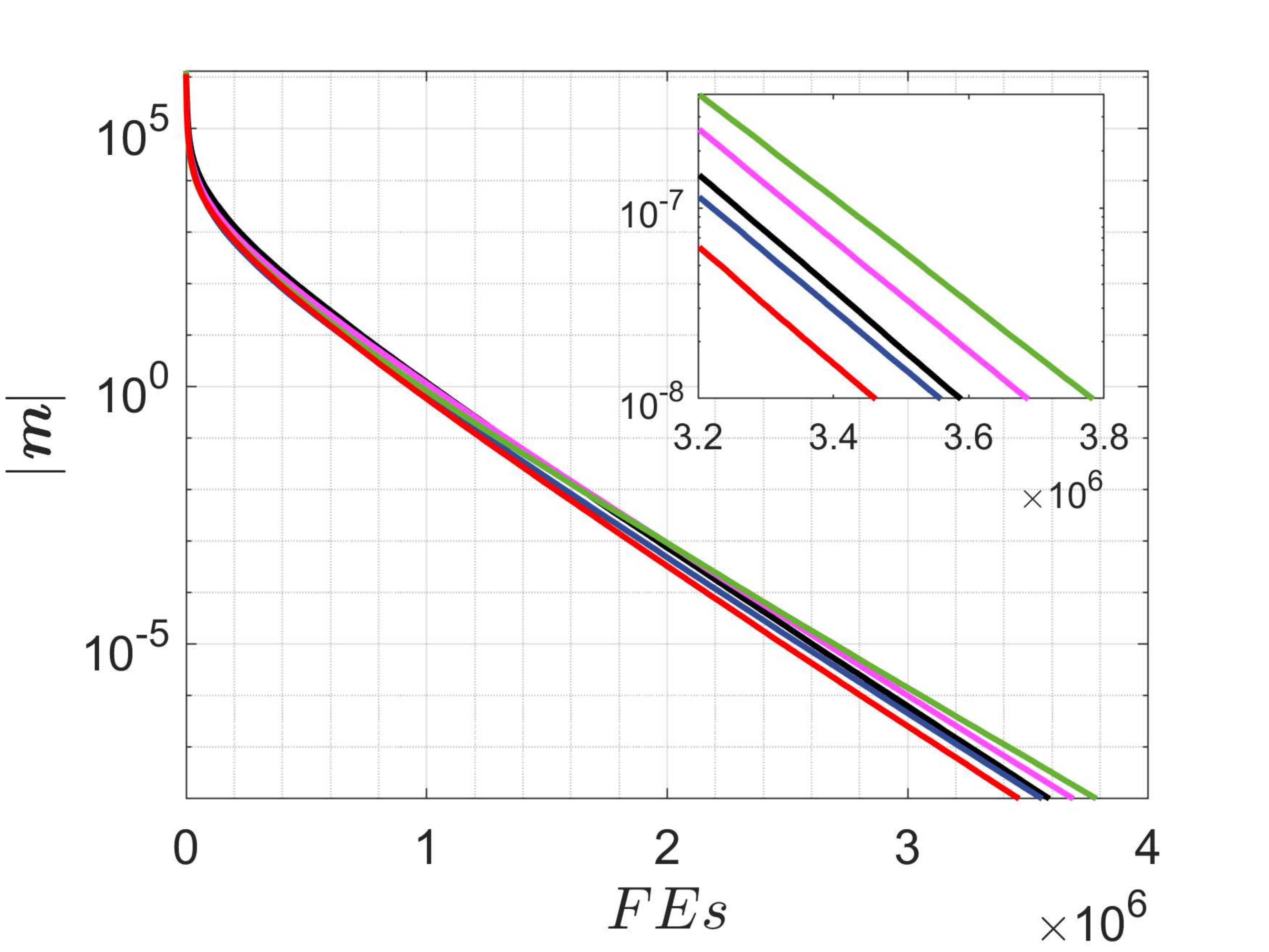} \label{subfig:1000D-Elli-cn3} }
\caption{Evolutionary trajectories on the 1000-dimensional Sphere problem and the Ellipsoid problems with different condition numbers. The vertical axis is the distance of the mean to the global optimum, calculated as $|\textbf{\textit{m}}|$, and the horizontal axis is the corresponding number of function evaluations required. The plots are also zoomed in on the end phase of the optimization process.}
\label{fig:convergence-curve-1000D-sphere}
\end{figure}

We first choose 1000-dimensional $\fSphere$ as the test problem. Since the function landscape can be perfectly fitted by the density contours of the isotropic Gaussian model, this test is to reveal the behavior of PTA in isolation from the covariance matrix adaptation or the sampling procedure. \Cref{subfig:1000D-Sphere} presents the Euclidean distance between the mean and the global optimum, calculated as $|\bm{m}|$, versus the number of function evaluations. We can observe from this semi-log plot that PTA enables ES to attain the linear convergence speed, as its trajectory is rendered by a descending line. This is consistent with the theoretical results about classic methods such as the 1/5-th success rule~\cite{morinaga_generalized_2019} and the CSA without cumulation~\cite{auger_linear_2016}. Other comparative methods also exhibit linear convergence and no significant difference is observed in terms of the convergence rate.

Compared with the simplest Sphere problem, well-conditioned Ellipsoid problem is more interesting because it simulates a real scenario where the optimal covariance matrix cannot be learned.
We thus consider 1000-dimensional $\fElli$ with the conditioning parameter $\alpha$ varying in $\{1,2,3\}$. The results are presented respectively in \Cref{subfig:1000D-Elli-cn1,subfig:1000D-Elli-cn2,subfig:1000D-Elli-cn3}. It is seen that all algorithms converge linearly except in the early stage. GSR exhibits the slowest convergence rate and is significantly surpassed by PTA, which is exactly opposite to the results on $\fSphere$.
The performance degradation of GSR is probability due to that the solutions give less information than on $\fSphere$ about how success the progress is made, and hence, GSR tends to accept small mutation strengths leading to a slow adaptation rate. PTA seems to alleviate this problem by using a weighting scheme to favor top ranked solutions. The above tests demonstrate that PTA can serve as an alternative to existing methods for mutation strength adaptation and its weighting scheme may also save computational effort on problems with small or moderate condition numbers.



\subsection{Rotational Invariance}
\label{ss:rotational-invariance}
Rotational invariance refers to the property of an algorithm that its performance does not change after rotating the decision space, provided that the algorithm is properly initialized. An algorithm possessing this property is robust to the non-separability of the problem, as the variable correlations can be linearly approximated by the rotational transformations on the decision space.
For the standard CMA-ES, the rotational invariance is built-in. It explicitly maintains all pairwise linear correlations in a covariance matrix such that the rotations can be completely captured. On the contrary, sep-CMA does not possess this property since it does not try to explore the variable correlations at all. Other comparative algorithms including MMES use only a set of direction vectors to reconstruct the probability model and therefore cannot learn all variable correlations. Thus, in this subsection, we verify whether they are invariant against rotations with numeric simulations.

\begin{figure}[tb]
\centering
\subfloat[$\fElli$]{\includegraphics[width=0.24\textwidth]{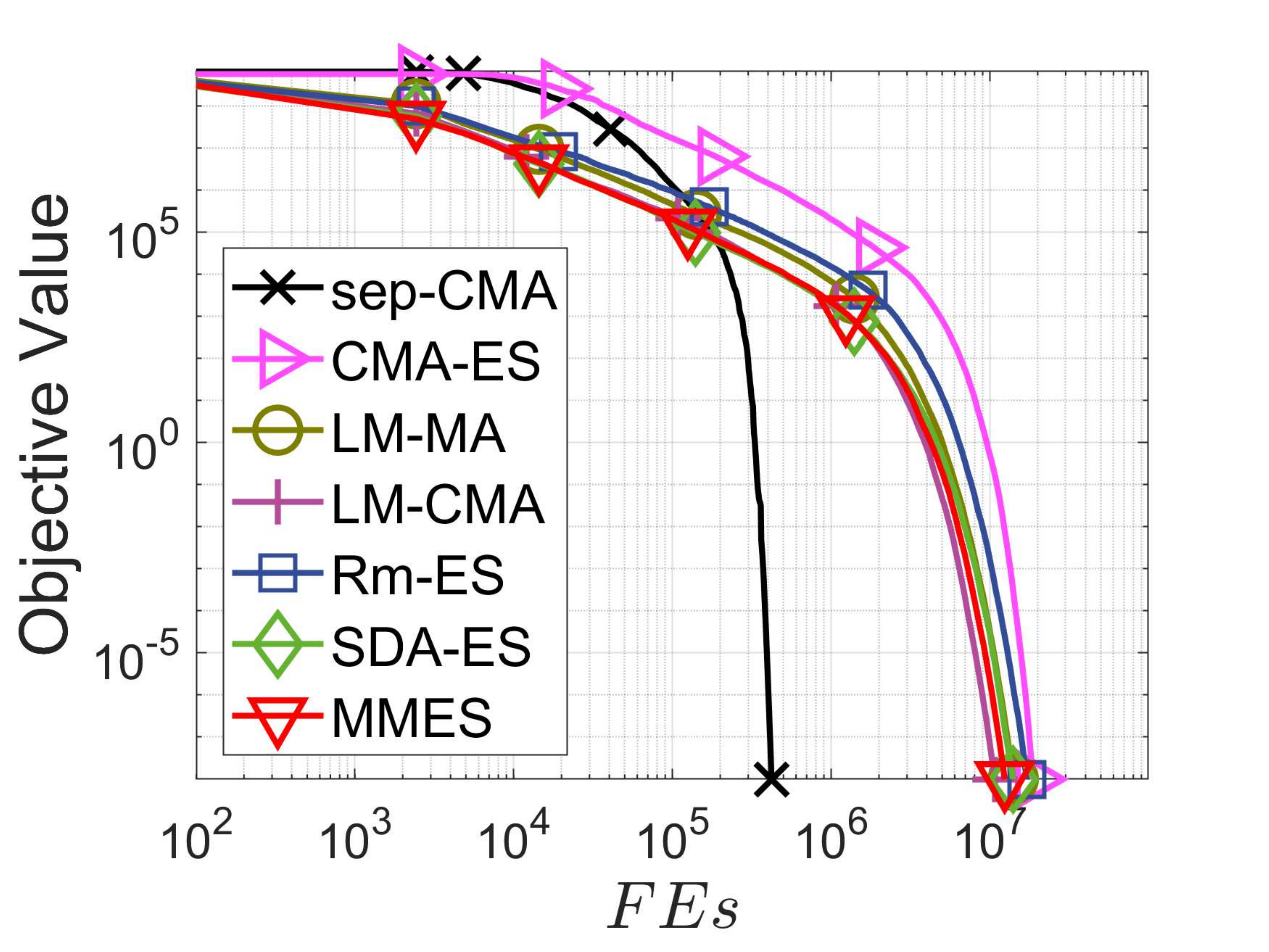} \label{subfig:1000D-Elli} }
\subfloat[$\fRotElli$]{\includegraphics[width=0.24\textwidth]{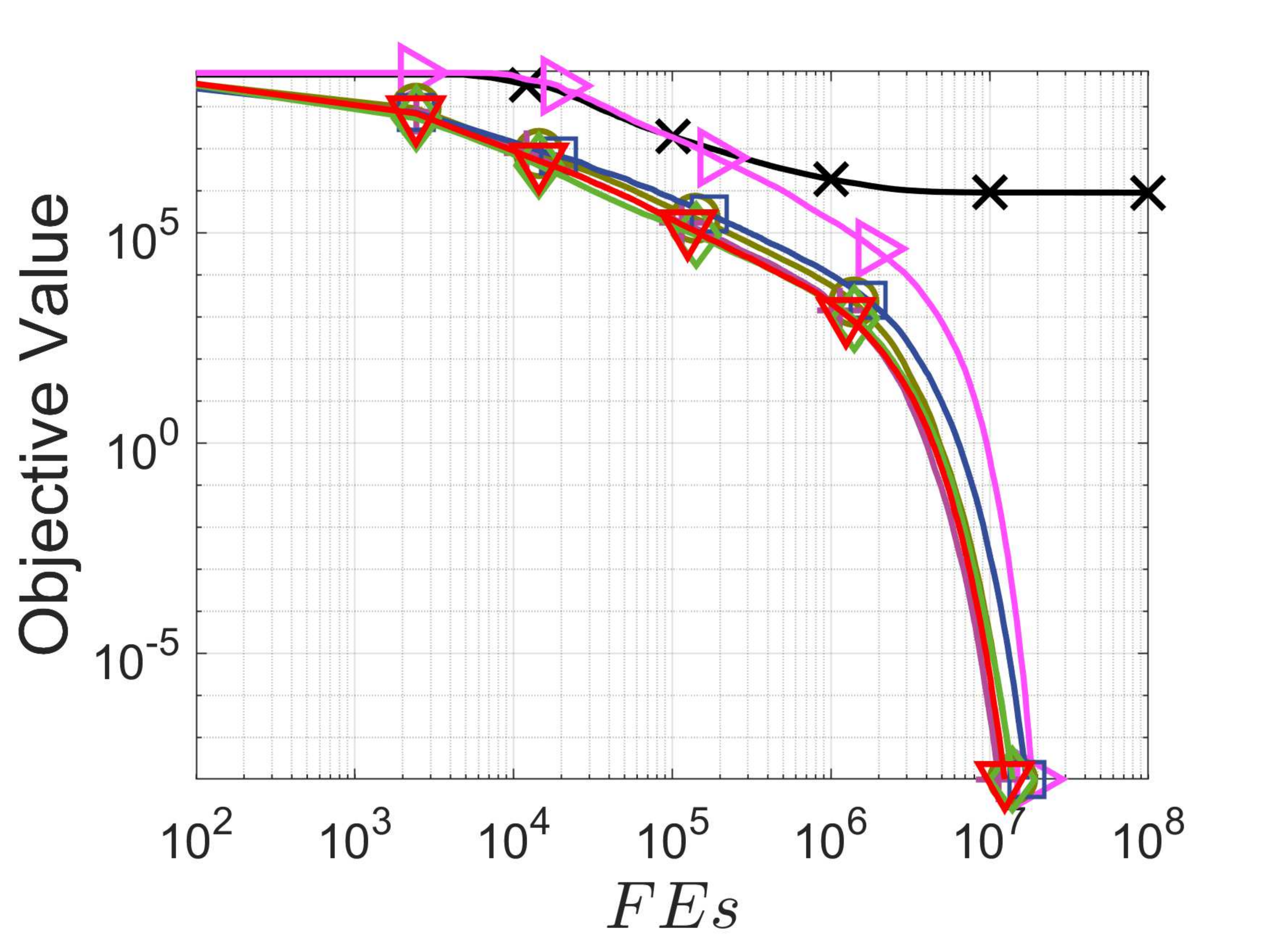} \label{subfig:1000D-RotElli} }
\hfil
\subfloat[$\fRosen$]{\includegraphics[width=0.24\textwidth]{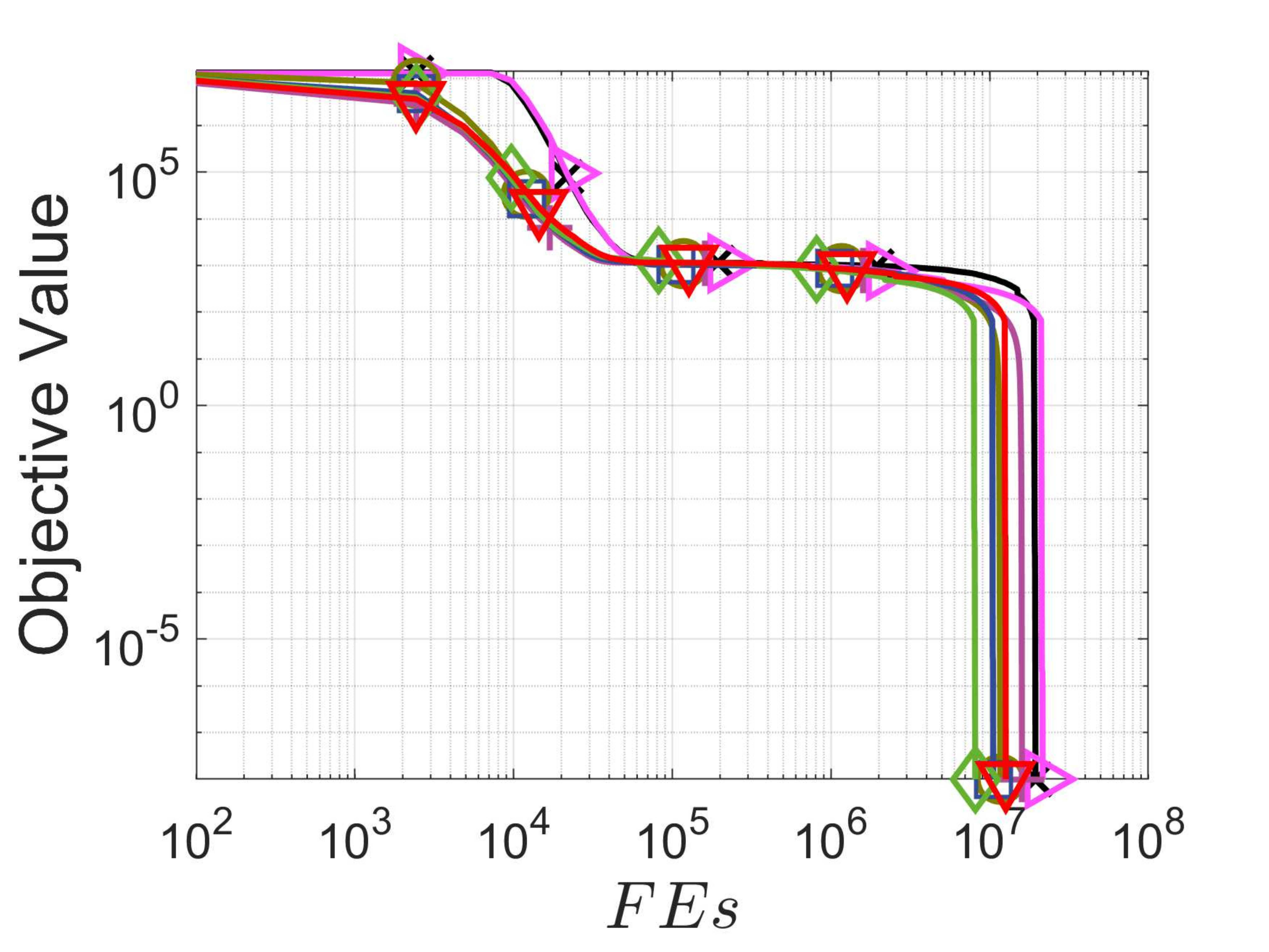} \label{subfig:1000D-Rosen} }
\subfloat[$\fRotRosen$]{\includegraphics[width=0.24\textwidth]{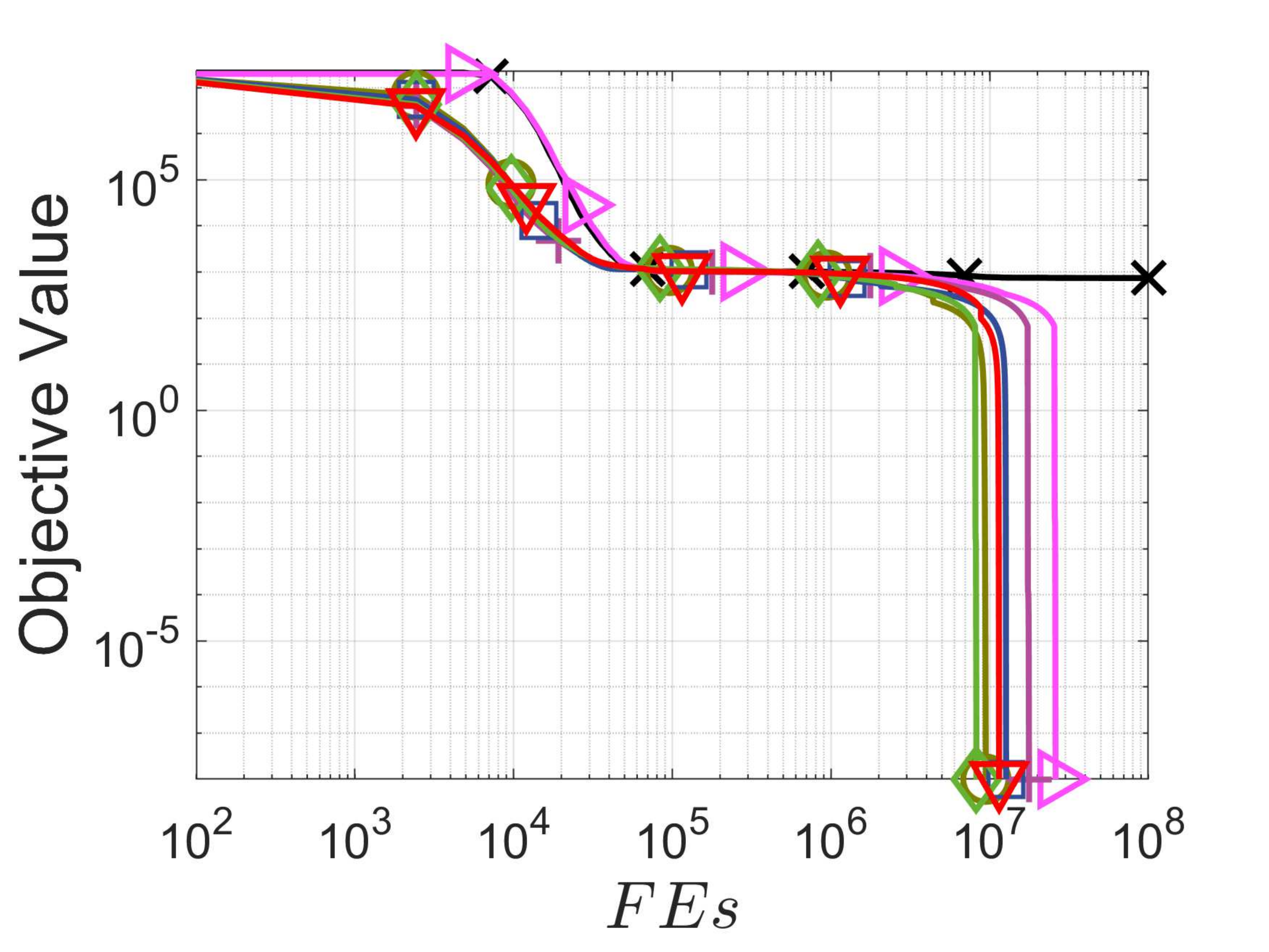} \label{subfig:1000D-RotRosen} }
\hfil
\subfloat[$\fDiscus$]{\includegraphics[width=0.24\textwidth]{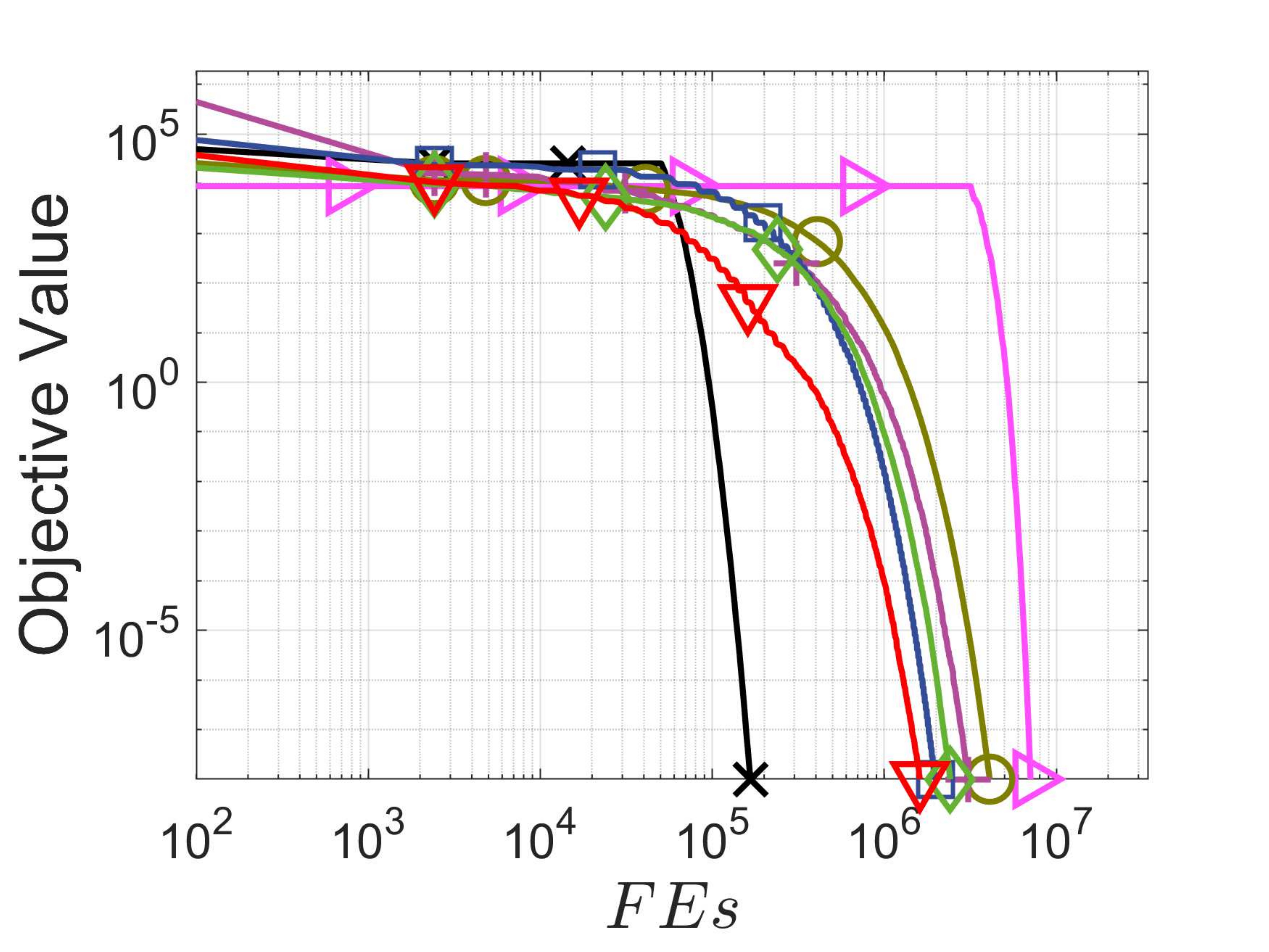} \label{subfig:1000D-Discus} }
\subfloat[$\fRotDiscus$]{\includegraphics[width=0.24\textwidth]{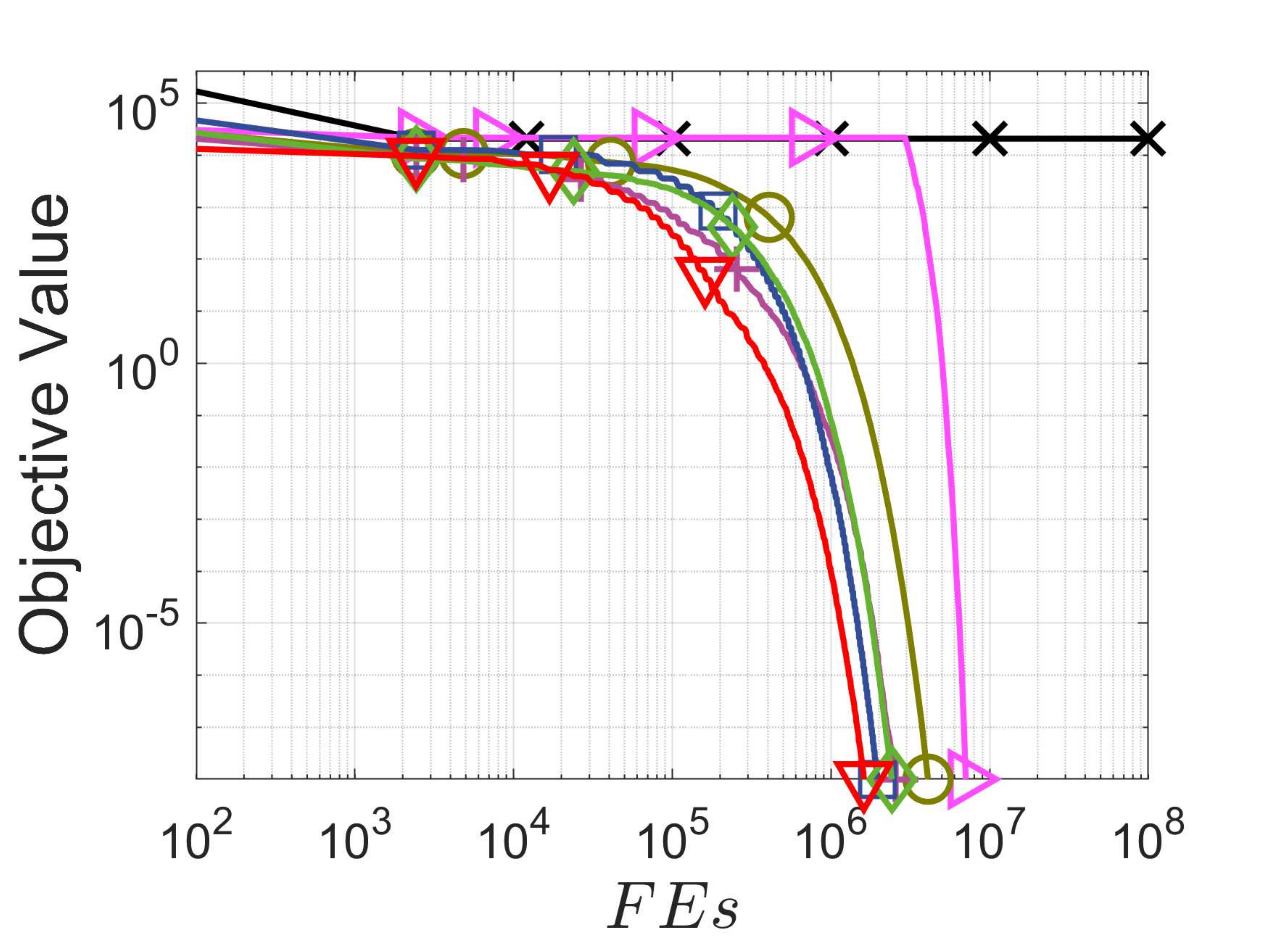} \label{subfig:1000D-RotDiscus} }
\hfil
\subfloat[$\fCigar$]{\includegraphics[width=0.24\textwidth]{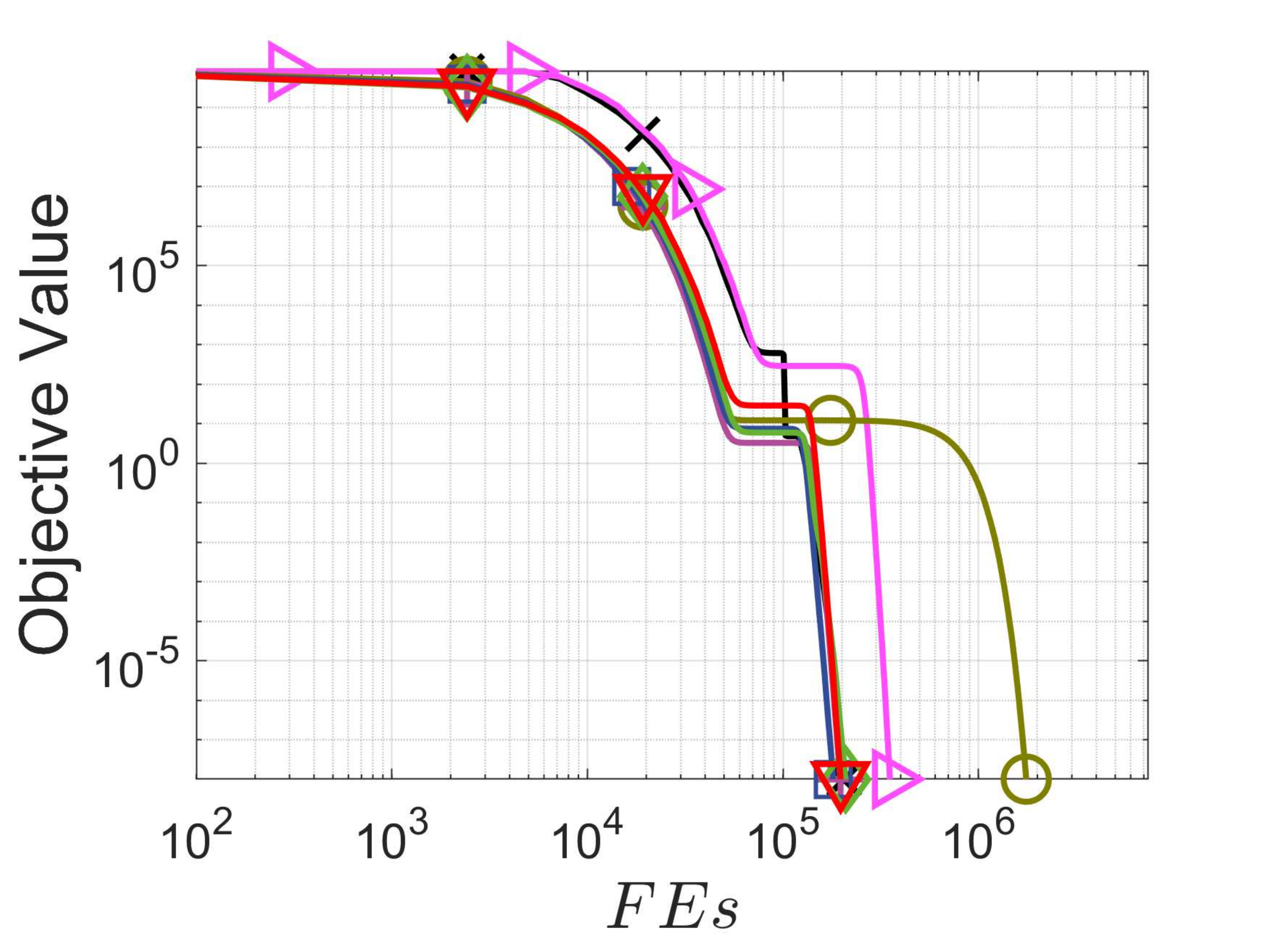} \label{subfig:1000D-Cigar} }
\subfloat[$\fRotCigar$]{\includegraphics[width=0.24\textwidth]{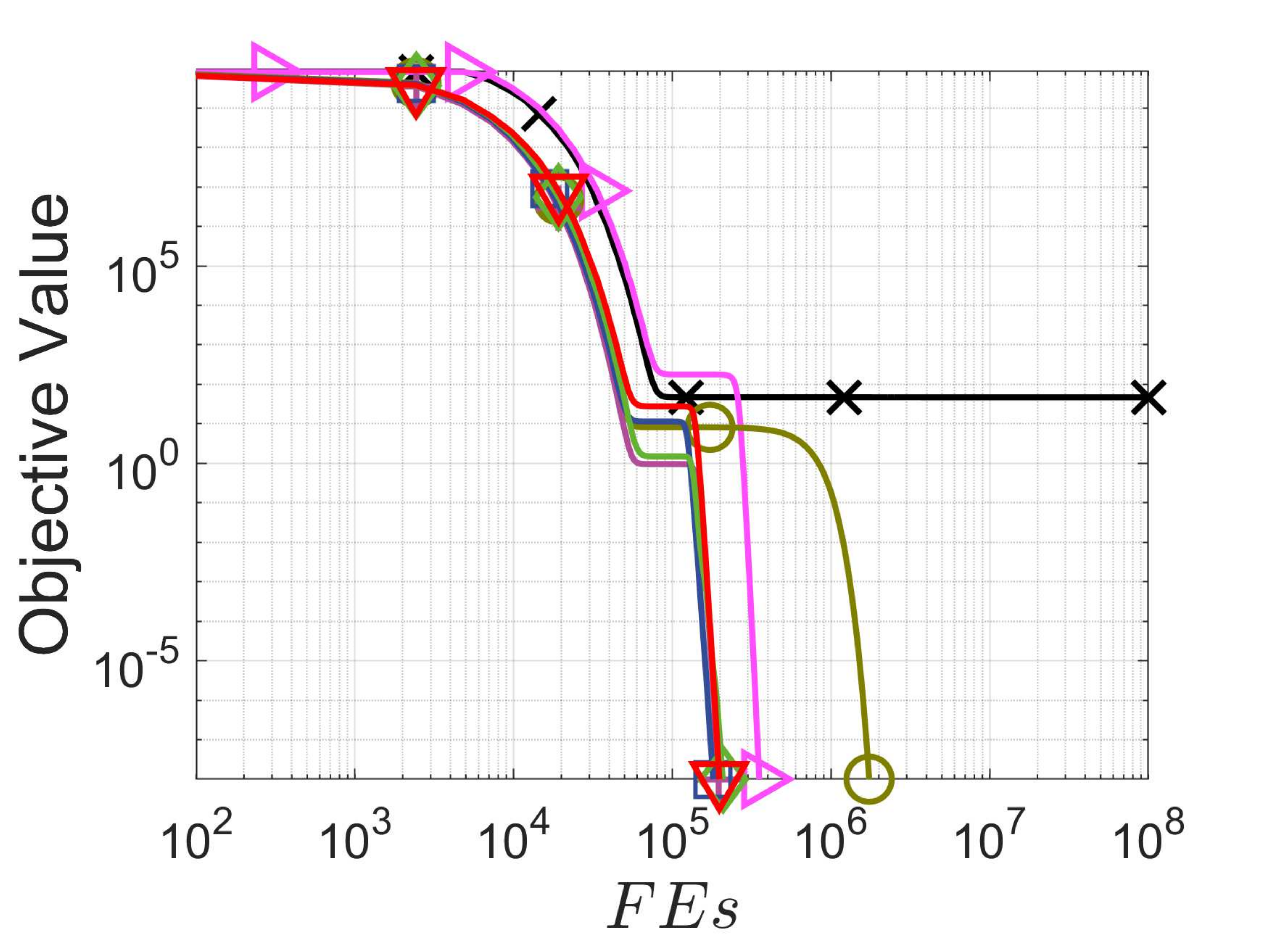} \label{subfig:1000D-RotCigar} }
\hfil
\subfloat[$\fDiffPow$]{\includegraphics[width=0.24\textwidth]{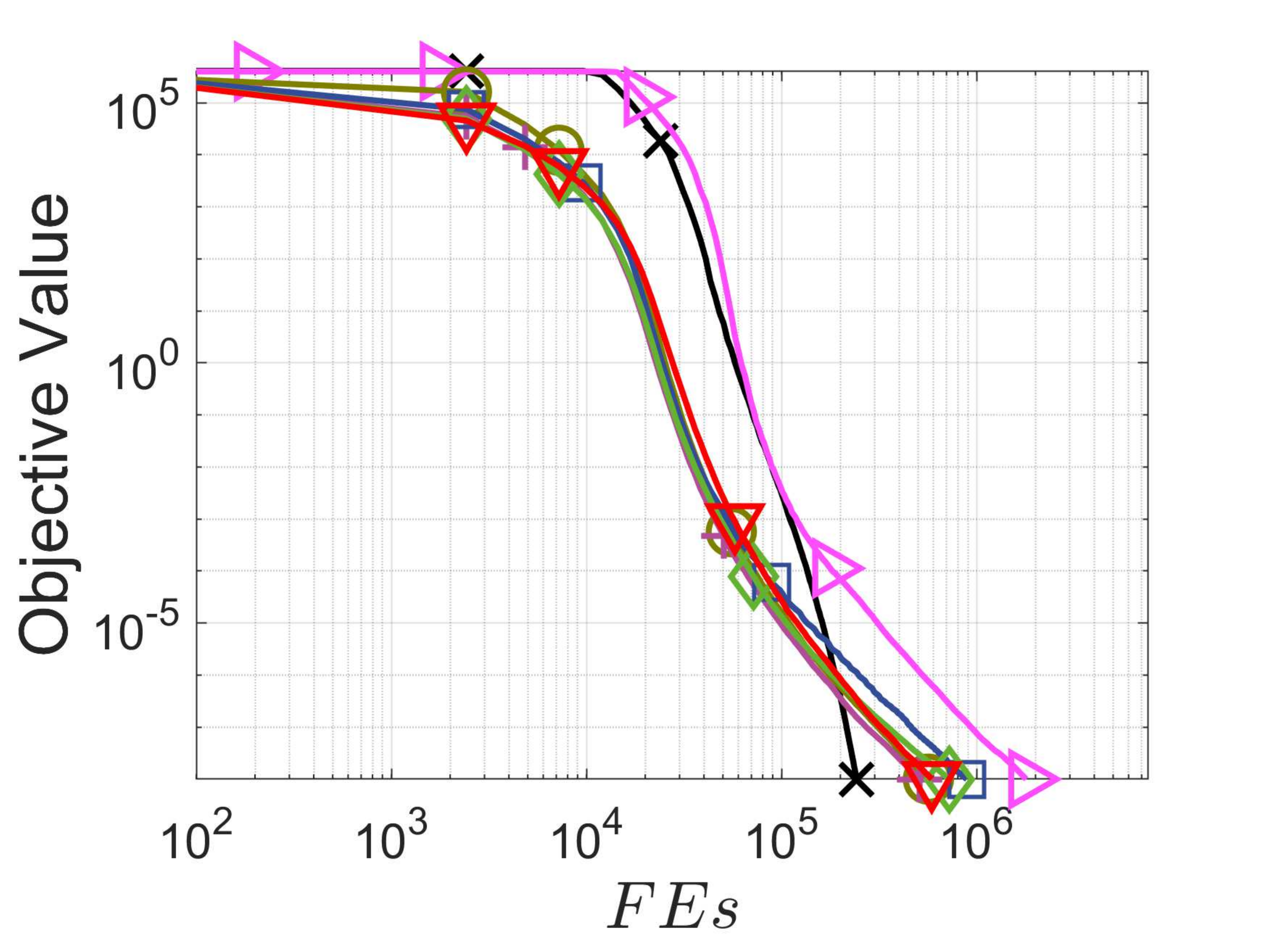} \label{subfig:1000D-DiffPow} }
\subfloat[$\fRotDiffPow$]{\includegraphics[width=0.24\textwidth]{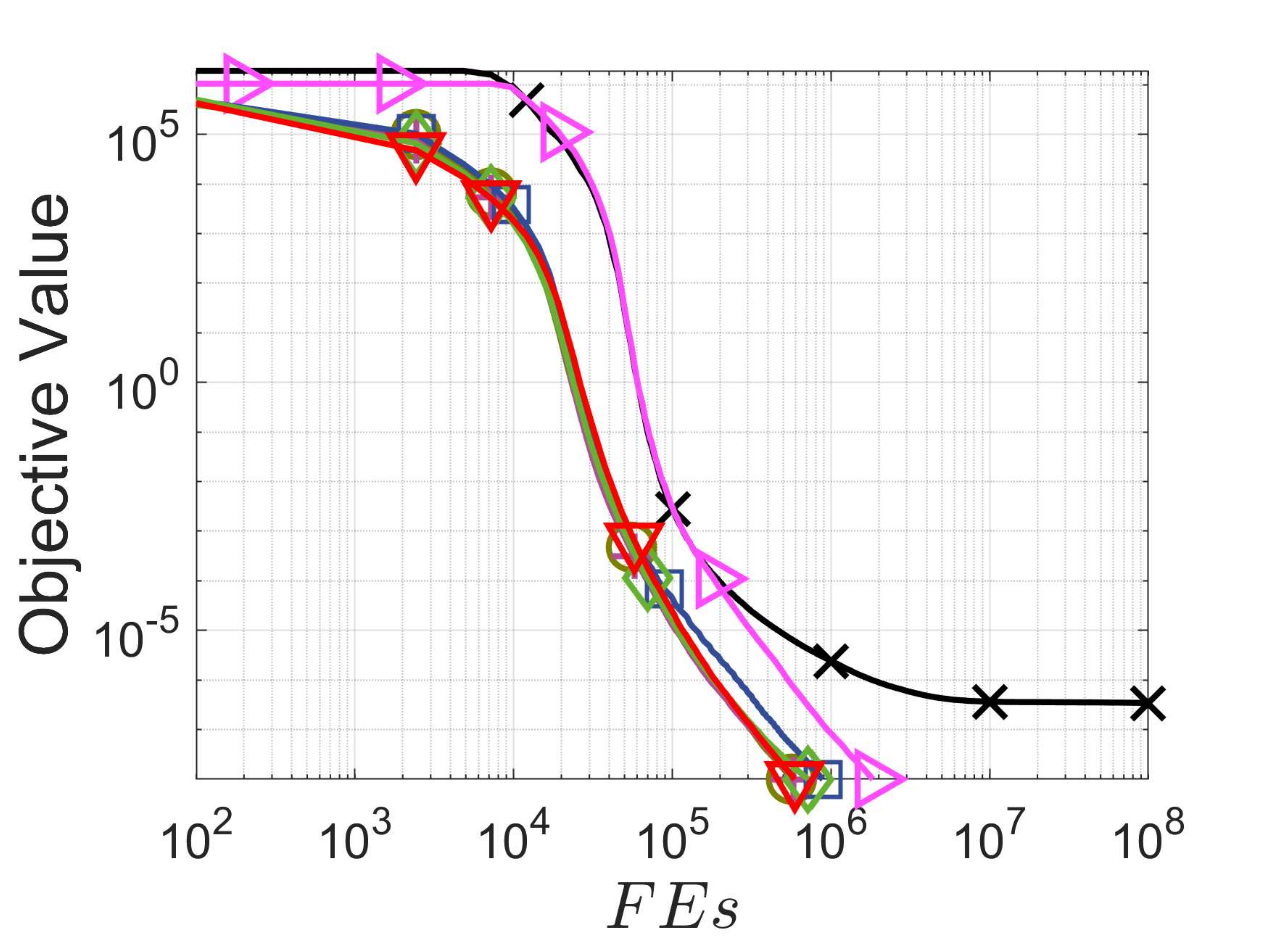} \label{subfig:1000D-RotDiffPow} }
\caption{Median results on the 1000-dimensional basic test problems with and without rotation, shown by evolutionary trajectories.}
\label{fig:test-rotational-invariance}
\end{figure}

\Cref{fig:test-rotational-invariance} shows the convergence behaviors of each algorithm on the 1000-dimensional unrotated basic test problems and on the corresponding rotated versions. It is evident that sep-CMA is sensitive to the rotations, as it performs well on most unrotated problems but fails to converge in all rotated problems. Such a performance deterioration is not seen for MMES or other modern CMA-ES variants, demonstrating that they are rotationally invariant. In fact, for MMES, the rotational invariance is guaranteed by design: both of its two key components, FMS and PTA, contain purely linear operations  (see Lines 8-15 and 27-29 in~\Cref{alg:MMES}) and are independent of certain Euclidean coordinates which may be changed by the rotations. The change in performance (if exists) is mainly caused by the non-invariant initialization and seems to be negligible according to the experiments.

\Cref{tab:1000-dimensional-summary} presents the ranking results for the algorithms that are invariant against rotations. 
The standard CMA-ES are surpassed by its variants, probably due to the large number of strategy parameters required to be adapted. Applying more advanced adaptation and sampling schemes (e.g.,~\cite{jastrebski_improving_2006,wang_mirrored_2019}) or carefully tuning the hyperparameters (e.g.,~\cite{krause_qualitative_2017,rowe_choice_2014}) is likely to address this issue.
MMES performs consistently well on all test problems and ranks first according to the Friedman test. The win/loss/tie record of the Wilcoxon test also shows it is competitive with LM-CMA and superior to the others on the majority of the problems. However, the statistical difference between MMES and the four variants are insignificant as reported by the Bonferroni post hoc procedure. Thus, no algorithm has an absolute advantage over the others. More detailed numerical results are found in Table S-I in the supplement. 

\begin{table}[htbp]
\centering
\scriptsize
\caption{Ranks of rotationally invariant algorithms on the 1000-dimensional basic test problems, in terms of the number of function evaluations required to reach the accuracy $10^{-8}$. Better results are highlighted with darker gray background. }
\renewcommand\arraystretch{1}
\setlength{\tabcolsep}{4pt}
\label{tab:1000-dimensional-summary}
\begin{threeparttable}
\begin{tabular}{ccccccc}
\toprule
  & CMA-ES & LM-MA & LM-CMA & Rm-ES & SDA-ES & MMES \\
\midrule
$\fElli$ & \cellcolor[rgb]{ .98,  .98,  .98} 6$\;\;\bullet$ & \cellcolor[rgb]{ .839,  .839,  .839} 4$\;\;\bullet$ & \cellcolor[rgb]{ .627,  .627,  .627} 1$\;\;\circ$ & \cellcolor[rgb]{ .91,  .91,  .91} 5$\;\;\bullet$ & \cellcolor[rgb]{ .769,  .769,  .769} 3$\;\;\bullet$ & \cellcolor[rgb]{ .698,  .698,  .698} 2 \\
$\fRosen$ & \cellcolor[rgb]{ .98,  .98,  .98} 6$\;\;\bullet$ & \cellcolor[rgb]{ .839,  .839,  .839} 4$\;\;\dagger$ & \cellcolor[rgb]{ .91,  .91,  .91} 5$\;\;\bullet$ & \cellcolor[rgb]{ .769,  .769,  .769} 3$\;\;\dagger$ & \cellcolor[rgb]{ .627,  .627,  .627} 1$\;\;\circ$ & \cellcolor[rgb]{ .698,  .698,  .698} 2 \\
$\fDiscus$ & \cellcolor[rgb]{ .98,  .98,  .98} 6$\;\;\bullet$ & \cellcolor[rgb]{ .91,  .91,  .91} 5$\;\;\bullet$ & \cellcolor[rgb]{ .839,  .839,  .839} 4$\;\;\bullet$ & \cellcolor[rgb]{ .698,  .698,  .698} 2$\;\;\bullet$ & \cellcolor[rgb]{ .769,  .769,  .769} 3$\;\;\bullet$ & \cellcolor[rgb]{ .627,  .627,  .627} 1 \\
$\fCigar$ & \cellcolor[rgb]{ .91,  .91,  .91} 5$\;\;\bullet$ & \cellcolor[rgb]{ .98,  .98,  .98} 6$\;\;\bullet$ & \cellcolor[rgb]{ .698,  .698,  .698} 2$\;\;\dagger$ & \cellcolor[rgb]{ .627,  .627,  .627} 1$\;\;\circ$ & \cellcolor[rgb]{ .839,  .839,  .839} 4$\;\;\bullet$ & \cellcolor[rgb]{ .769,  .769,  .769} 3 \\
$\fDiffPow$ & \cellcolor[rgb]{ .98,  .98,  .98} 6$\;\;\bullet$ & \cellcolor[rgb]{ .698,  .698,  .698} 2$\;\;\circ$ & \cellcolor[rgb]{ .627,  .627,  .627} 1$\;\;\circ$ & \cellcolor[rgb]{ .91,  .91,  .91} 5$\;\;\bullet$ & \cellcolor[rgb]{ .839,  .839,  .839} 4$\;\;\bullet$ & \cellcolor[rgb]{ .769,  .769,  .769} 3 \\
$\fRotElli$ & \cellcolor[rgb]{ .98,  .98,  .98} 6$\;\;\bullet$ & \cellcolor[rgb]{ .769,  .769,  .769} 3$\;\;\bullet$ & \cellcolor[rgb]{ .627,  .627,  .627} 1$\;\;\circ$ & \cellcolor[rgb]{ .91,  .91,  .91} 5$\;\;\bullet$ & \cellcolor[rgb]{ .839,  .839,  .839} 4$\;\;\bullet$ & \cellcolor[rgb]{ .698,  .698,  .698} 2 \\
$\fRotRosen$ & \cellcolor[rgb]{ .98,  .98,  .98} 6$\;\;\bullet$ & \cellcolor[rgb]{ .698,  .698,  .698} 2$\;\;\circ$ & \cellcolor[rgb]{ .91,  .91,  .91} 5$\;\;\bullet$ & \cellcolor[rgb]{ .839,  .839,  .839} 4$\;\;\dagger$ & \cellcolor[rgb]{ .627,  .627,  .627} 1$\;\;\circ$ & \cellcolor[rgb]{ .769,  .769,  .769} 3 \\
$\fRotDiscus$ & \cellcolor[rgb]{ .98,  .98,  .98} 6$\;\;\bullet$ & \cellcolor[rgb]{ .91,  .91,  .91} 5$\;\;\bullet$ & \cellcolor[rgb]{ .839,  .839,  .839} 4$\;\;\bullet$ & \cellcolor[rgb]{ .698,  .698,  .698} 2$\;\;\bullet$ & \cellcolor[rgb]{ .769,  .769,  .769} 3$\;\;\bullet$ & \cellcolor[rgb]{ .627,  .627,  .627} 1 \\
$\fRotCigar$ & \cellcolor[rgb]{ .91,  .91,  .91} 5$\;\;\bullet$ & \cellcolor[rgb]{ .98,  .98,  .98} 6$\;\;\bullet$ & \cellcolor[rgb]{ .698,  .698,  .698} 2$\;\;\dagger$ & \cellcolor[rgb]{ .627,  .627,  .627} 1$\;\;\circ$ & \cellcolor[rgb]{ .839,  .839,  .839} 4$\;\;\bullet$ & \cellcolor[rgb]{ .769,  .769,  .769} 3 \\
$\fRotDiffPow$ & \cellcolor[rgb]{ .98,  .98,  .98} 6$\;\;\bullet$ & \cellcolor[rgb]{ .627,  .627,  .627} 1$\;\;\circ$ & \cellcolor[rgb]{ .698,  .698,  .698} 2$\;\;\circ$ & \cellcolor[rgb]{ .91,  .91,  .91} 5$\;\;\bullet$ & \cellcolor[rgb]{ .839,  .839,  .839} 4$\;\;\bullet$ & \cellcolor[rgb]{ .769,  .769,  .769} 3 \\
\midrule
$\bullet$ / $\circ$ / $\dagger$ & 10 / 0 / 0 & 6 / 3 / 1 & 4 / 4 / 2 & 6 / 2 / 2 & 8 / 2 / 0 &  \\
\midrule
Avg Rank & 5.8 & 3.8 & 2.7 & 3.3 & 3.1 & 2.3 \\
\midrule
$p$-Value & 0.00 & 1 & 1 & 1 & 1 &  \\
\bottomrule
\end{tabular}%

\begin{tablenotes}
\item[1] ``$\bullet$'' indicates that MMES significantly outperforms the peer algorithm at a 0.05 significance level by the Wilcoxon test, whereas ``$\circ$'' indicates the opposite. If no significant difference is detected, it will be marked by the symbol ``$\dagger$''. 
\item[2] ``Avg Rank'' denotes the ranking results averaged over all problems according to the Friedman test.
\item[3] ``$p$-Value'' denotes the significance of difference between the averaged ranks of MMES and the pair algorithms, corrected by the Bonferroni procedure. 
\end{tablenotes}
\end{threeparttable}
\end{table}

\subsection{Scalability}
\label{ss:scalability}
\begin{figure}[tb]
\centering
\subfloat[$f_{\text{Elli}}$]{\includegraphics[width=0.24\textwidth]{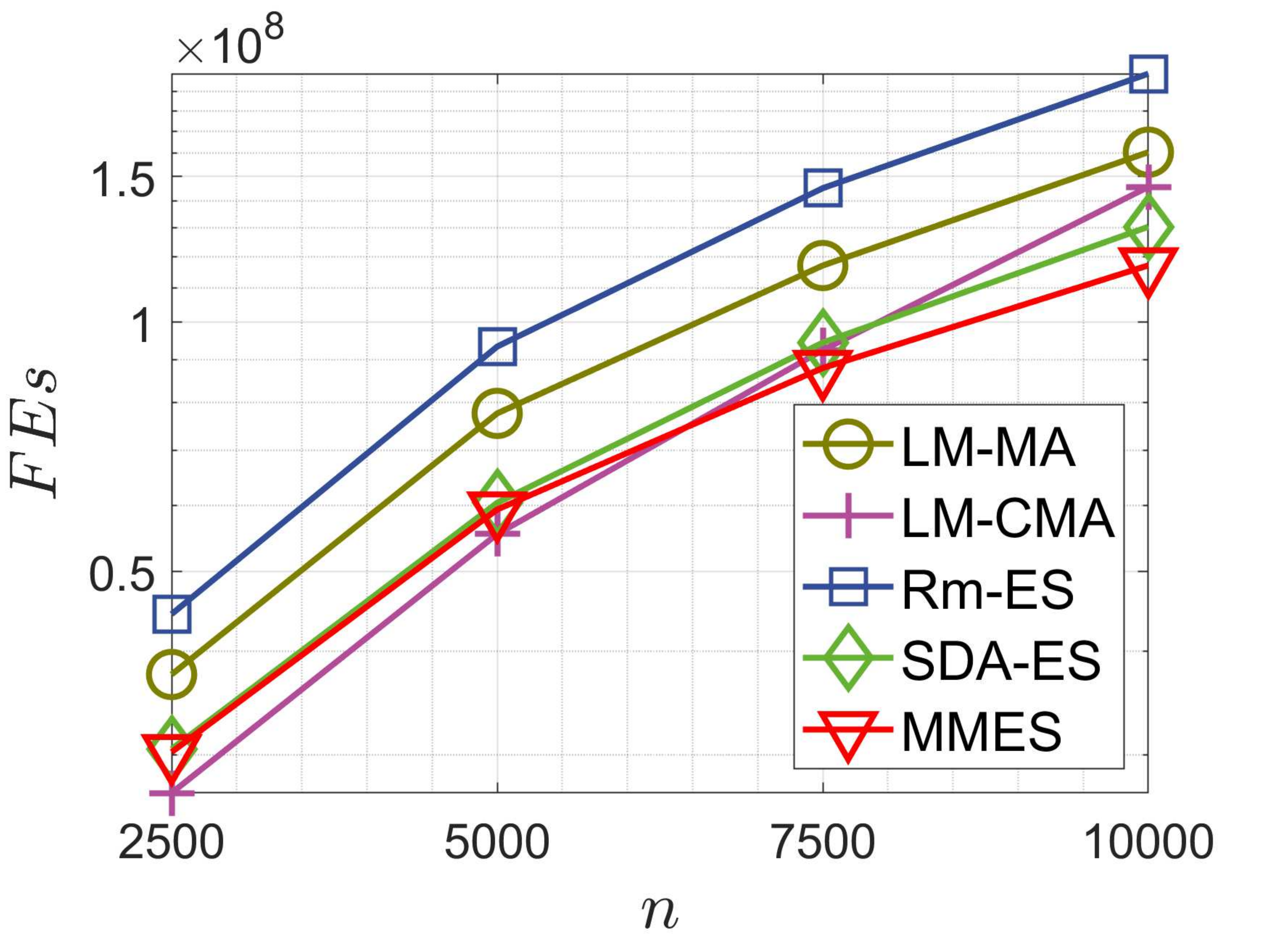} \label{subfig:scalability-Elli}}
\hfil
\subfloat[$f_{\text{Rosen}}$]{\includegraphics[width=0.24\textwidth]{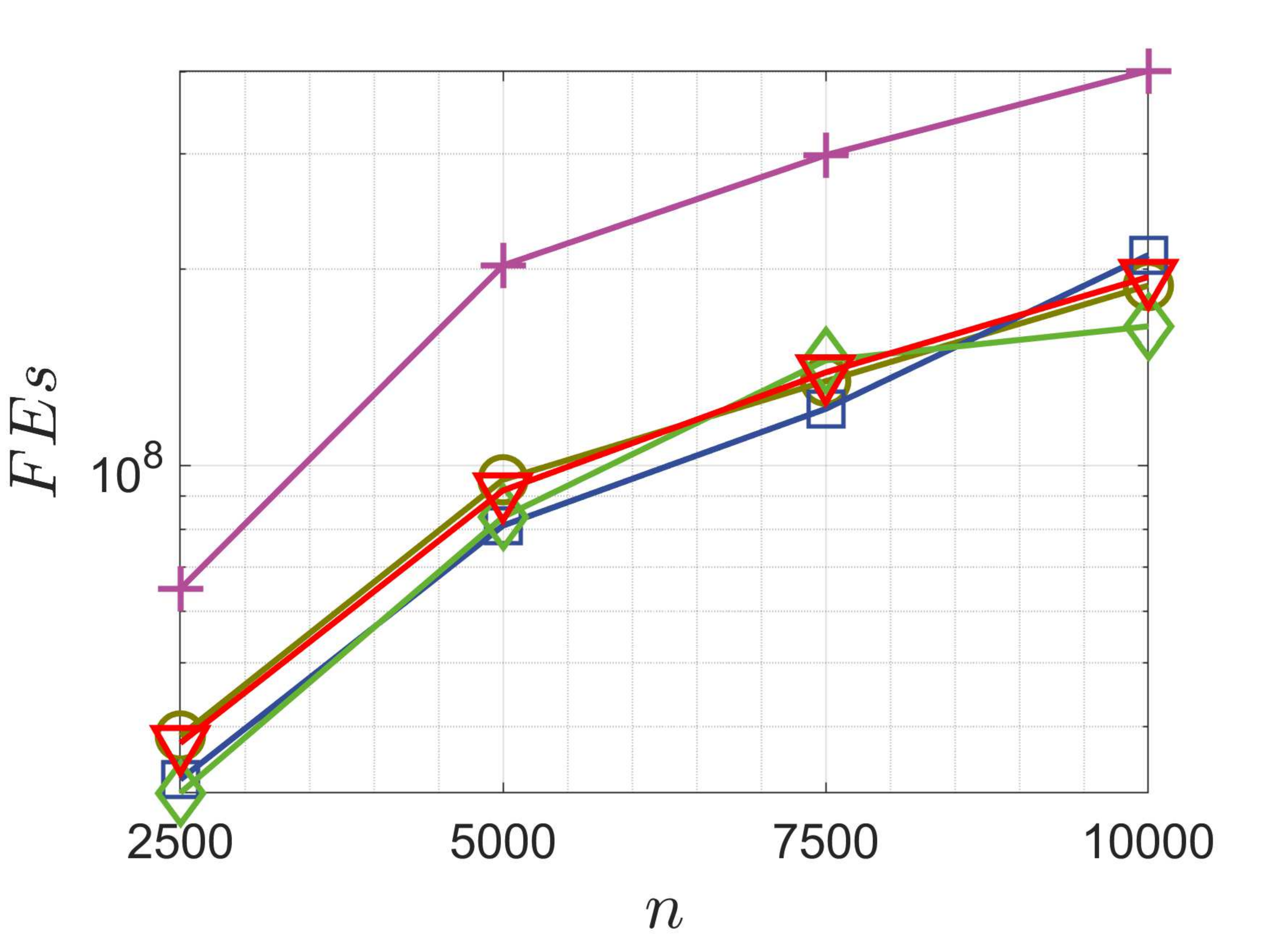} \label{subfig:scalability-Rosen}}
\subfloat[$f_{\text{Discus}}$]{\includegraphics[width=0.24\textwidth]{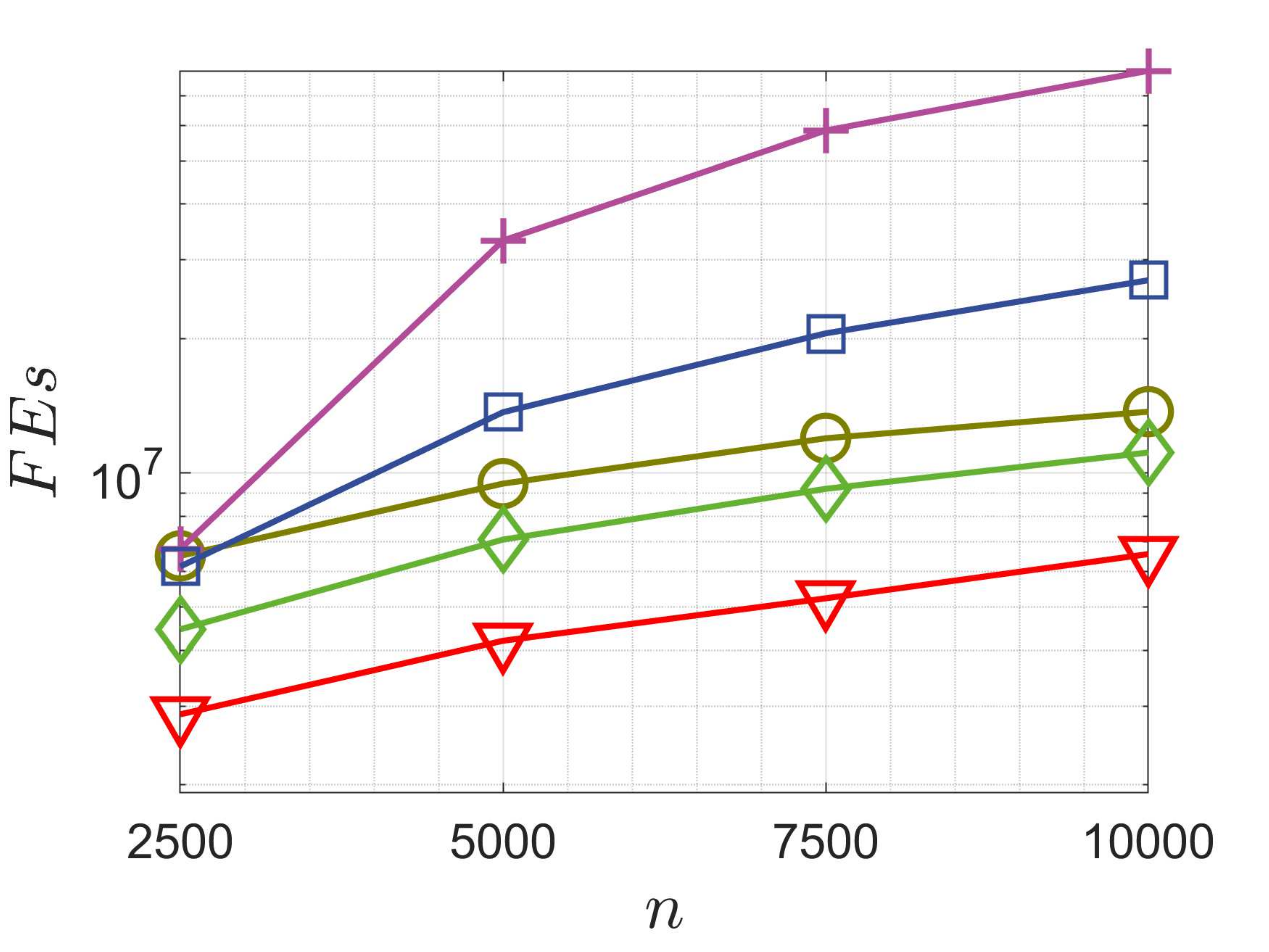} \label{subfig:scalability-Discus}}
\hfil
\subfloat[$f_{\text{Cigar}}$]{\includegraphics[width=0.24\textwidth]{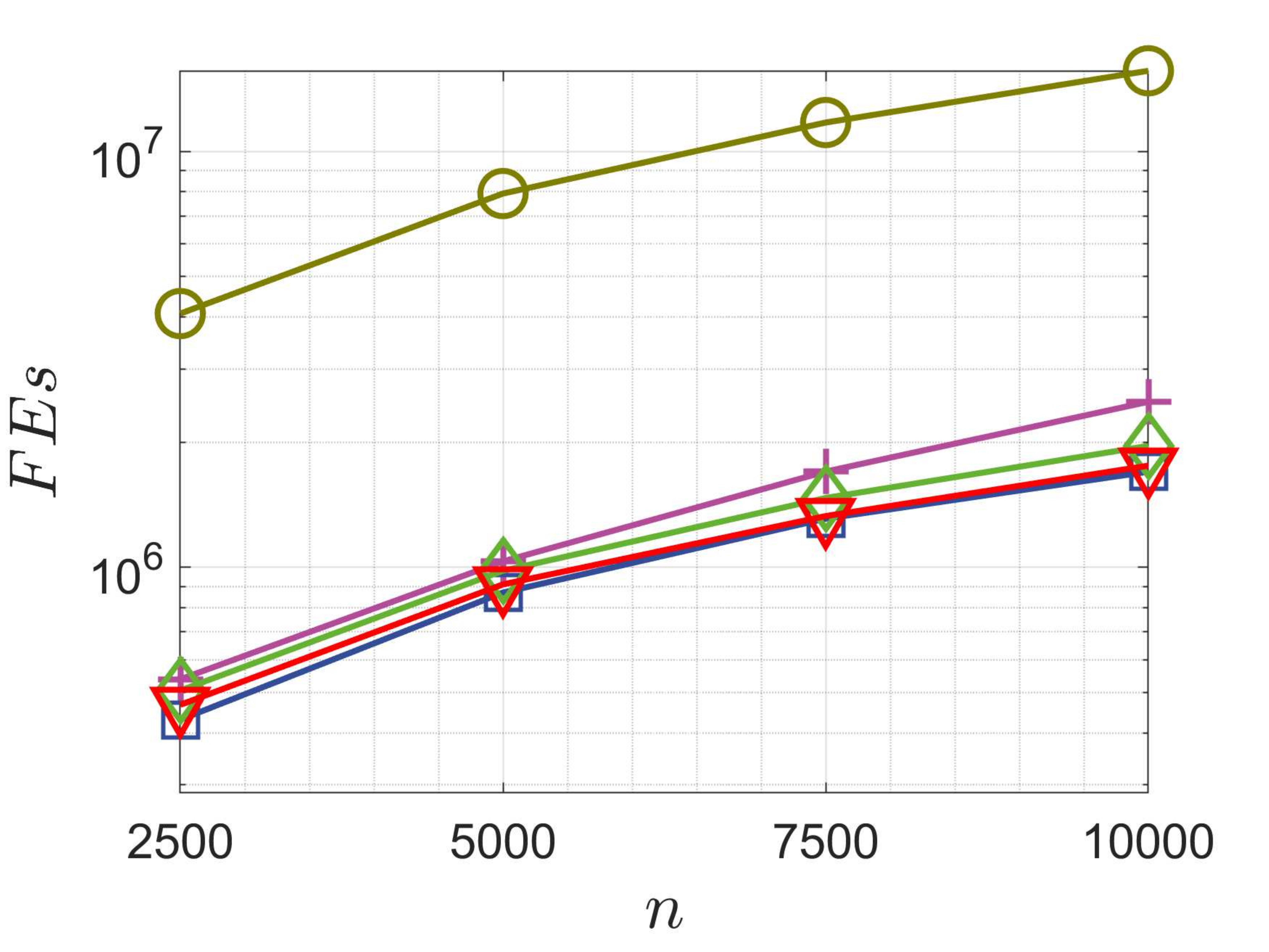} \label{subfig:scalability-Cigar}}
\subfloat[$f_{\text{DiffPow}}$]{\includegraphics[width=0.24\textwidth]{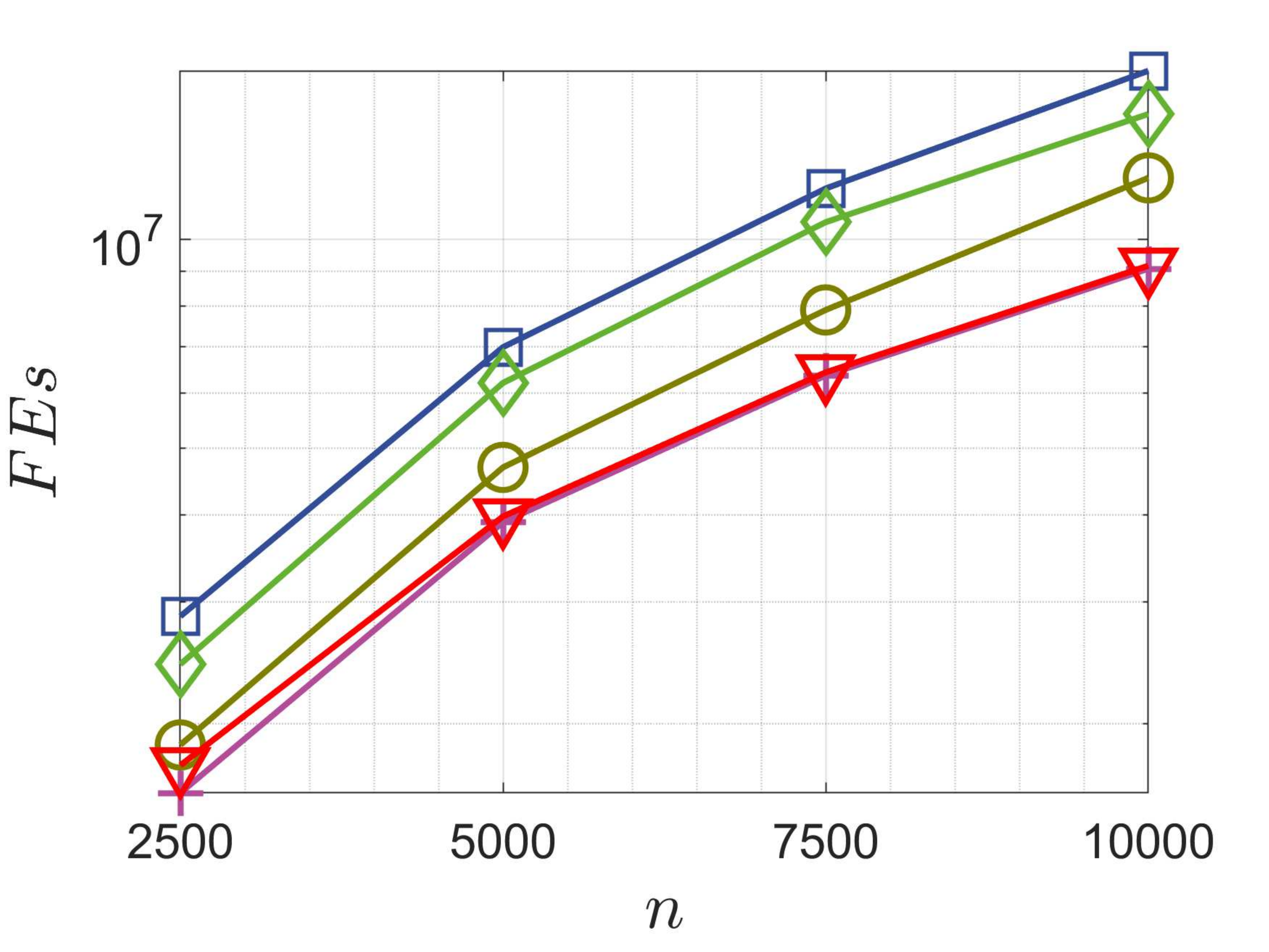} \label{subfig:scalability-DiffPow}}
\caption{Median results on the 2500-, 5000-, 7500-, and 10000-dimensional non-rotated problems. The curves present the number of function evaluations required to reach the accuracy $10^{-8}$. }
\label{fig:scalability-test}
\end{figure}

We investigate the scalability of MMES on the unrotated basic test problems with $n=2500, 5000, 7500$, and 1000. The number of function evaluations required to reach the accuracy $10^{-8}$ are plotted in~\Cref{fig:scalability-test}. 

On the Ellipsoid problem $\fElli$, MMES outperforms all except LM-CMA on the 2500- and 5000-dimensional instances and is the best performer on the higher-dimensional instances. This problem is characterized by a Hessian with uniformly spread eigenvalues in the log scale, and hence, its landscape cannot be well learned unless there are sufficiently many direction vectors for probability model reconstruction. The good performance of MMES demonstrates its advantage in exploiting the rich correlations with no extra time cost. All the other algorithms scale linearly with a constant factor varying with different methods for probability model adaptation; the linear scaling is because of the fixed number of direction vectors to be adapted. An interesting observation is that MMES also scales linearly although the number of direction vectors increases sub-linearly with the increasing dimension, possibly because of the non-linear setting for the adaptation rate. 

MMES is better than or as good as LM-CMA and LM-MA on $\fRosen$. This problem is much more difficult than $\fElli$ as the solutions have to pass through a long and narrow valley which cannot be rendered by a single linear transformation~\cite{shang_note_2006}. Rm-ES and SDA-ES achieve better performance because of the fewer direction vectors to be adapted in exploiting the landscape structure. However, some comparison results are insignificant because the landscape is multimodal and the algorithms seem to suffer from premature convergence.

On $\fCigar$, MMES achieves competitive results compared with the best performer, Rm-ES. This result is different from that observed on $\fRosen$, although their landscapes are both low-rank structured. In fact, $\fCigar$ is much simpler in that it is convex quadratic and the shape of the landscape can be completely described by only one direction vector. The relatively good performance of MMES suggests that it sustains the capability of capturing the most promising search direction even when the used direction vectors are more than required. 

MMES produces surprisingly good results on $\fDiscus$. The spectrum of the Hessian suggests that the optimal covariance matrix for capturing the variable correlations should have $n-1$ identical eigenvalues and a single one that is one million times smaller. Thus, the ESs have to learn all the difference in scalings between the most sensitive direction (i.e., $x_1$) and the other $n-1$ ones, which is, obviously, not the case for the algorithms chosen in the experiments. The good performance of MMES may be contributed to the fatter tails of the probability distribution which leads to a higher probability of generating long jumps on the insensitive directions. A parameter sensitiveness test presented later will verify this statement.

The problem $\fDiffPow$ is similar to $\fElli$ except that the variable correlations are non-linear and vary with position. The closer to the optimum, the more difficult an algorithm gets to approach it further. MMES and LM-CMA both perform well on this problem and demonstrate their local exploitation ability.

The ranking results of all the algorithms are summarized in \Cref{tab:scability-ranks}. It suggests that MMES ranks firstly according to the Friedman test and significantly surpasses LM-MA, LM-CMA, and Rm-ES. More detailed results are found in Table S-II in the supplement.

\begin{table}[ht]
\centering
\scriptsize
\caption{Ranks of large-scale CMA-ES variants on the 2500-, 5000-, 7500-, and 10000-dimensional unrotated problems in terms of the number of function evaluations required to reach the accuracy $10^{-8}$. Better results are highlighted with darker gray background.}
\renewcommand\arraystretch{1}
\setlength{\tabcolsep}{4pt}
\label{tab:scability-ranks}
\begin{tabular}{ccccccc}
\toprule
  & $n$ & LM-MA & LM-CMA & Rm-ES & SDA-ES & MMES \\
\midrule 
$\fElli$ & 2500 & \cellcolor[rgb]{ .894,  .894,  .894} 4$\;\;\bullet$ & \cellcolor[rgb]{ .627,  .627,  .627} 1$\;\;\circ$ & \cellcolor[rgb]{ .98,  .98,  .98} 5$\;\;\bullet$ & \cellcolor[rgb]{ .804,  .804,  .804} 3$\;\;\bullet$ & \cellcolor[rgb]{ .718,  .718,  .718} 2 \\
  & 5000 & \cellcolor[rgb]{ .894,  .894,  .894} 4$\;\;\bullet$ & \cellcolor[rgb]{ .627,  .627,  .627} 1$\;\;\circ$ & \cellcolor[rgb]{ .98,  .98,  .98} 5$\;\;\bullet$ & \cellcolor[rgb]{ .804,  .804,  .804} 3$\;\;\bullet$ & \cellcolor[rgb]{ .718,  .718,  .718} 2 \\
  & 7500 & \cellcolor[rgb]{ .894,  .894,  .894} 4$\;\;\bullet$ & \cellcolor[rgb]{ .718,  .718,  .718} 2$\;\;\bullet$ & \cellcolor[rgb]{ .98,  .98,  .98} 5$\;\;\bullet$ & \cellcolor[rgb]{ .804,  .804,  .804} 3$\;\;\bullet$ & \cellcolor[rgb]{ .627,  .627,  .627} 1 \\
  & 10000 & \cellcolor[rgb]{ .894,  .894,  .894} 4$\;\;\bullet$ & \cellcolor[rgb]{ .804,  .804,  .804} 3$\;\;\bullet$ & \cellcolor[rgb]{ .98,  .98,  .98} 5$\;\;\bullet$ & \cellcolor[rgb]{ .718,  .718,  .718} 2$\;\;\bullet$ & \cellcolor[rgb]{ .627,  .627,  .627} 1 \\
\midrule 
$\fRosen$ & 2500 & \cellcolor[rgb]{ .894,  .894,  .894} 4$\;\;\dagger$ & \cellcolor[rgb]{ .98,  .98,  .98} 5$\;\;\bullet$ & \cellcolor[rgb]{ .718,  .718,  .718} 2$\;\;\dagger$ & \cellcolor[rgb]{ .627,  .627,  .627} 1$\;\;\dagger$ & \cellcolor[rgb]{ .804,  .804,  .804} 3 \\
  & 5000 & \cellcolor[rgb]{ .894,  .894,  .894} 4$\;\;\dagger$ & \cellcolor[rgb]{ .98,  .98,  .98} 5$\;\;\bullet$ & \cellcolor[rgb]{ .627,  .627,  .627} 1$\;\;\dagger$ & \cellcolor[rgb]{ .718,  .718,  .718} 2$\;\;\dagger$ & \cellcolor[rgb]{ .804,  .804,  .804} 3 \\
  & 7500 & \cellcolor[rgb]{ .718,  .718,  .718} 2$\;\;\dagger$ & \cellcolor[rgb]{ .98,  .98,  .98} 5$\;\;\bullet$ & \cellcolor[rgb]{ .627,  .627,  .627} 1$\;\;\dagger$ & \cellcolor[rgb]{ .894,  .894,  .894} 4$\;\;\dagger$ & \cellcolor[rgb]{ .804,  .804,  .804} 3 \\
  & 10000 & \cellcolor[rgb]{ .718,  .718,  .718} 2$\;\;\dagger$ & \cellcolor[rgb]{ .98,  .98,  .98} 5$\;\;\bullet$ & \cellcolor[rgb]{ .894,  .894,  .894} 4$\;\;\dagger$ & \cellcolor[rgb]{ .627,  .627,  .627} 1$\;\;\circ$ & \cellcolor[rgb]{ .804,  .804,  .804} 3 \\
\midrule 
$\fDiscus$ & 2500 & \cellcolor[rgb]{ .894,  .894,  .894} 4$\;\;\bullet$ & \cellcolor[rgb]{ .98,  .98,  .98} 5$\;\;\bullet$ & \cellcolor[rgb]{ .804,  .804,  .804} 3$\;\;\bullet$ & \cellcolor[rgb]{ .718,  .718,  .718} 2$\;\;\bullet$ & \cellcolor[rgb]{ .627,  .627,  .627} 1 \\
  & 5000 & \cellcolor[rgb]{ .804,  .804,  .804} 3$\;\;\bullet$ & \cellcolor[rgb]{ .98,  .98,  .98} 5$\;\;\bullet$ & \cellcolor[rgb]{ .894,  .894,  .894} 4$\;\;\bullet$ & \cellcolor[rgb]{ .718,  .718,  .718} 2$\;\;\bullet$ & \cellcolor[rgb]{ .627,  .627,  .627} 1 \\
  & 7500 & \cellcolor[rgb]{ .804,  .804,  .804} 3$\;\;\bullet$ & \cellcolor[rgb]{ .98,  .98,  .98} 5$\;\;\bullet$ & \cellcolor[rgb]{ .894,  .894,  .894} 4$\;\;\bullet$ & \cellcolor[rgb]{ .718,  .718,  .718} 2$\;\;\bullet$ & \cellcolor[rgb]{ .627,  .627,  .627} 1 \\
  & 10000 & \cellcolor[rgb]{ .804,  .804,  .804} 3$\;\;\bullet$ & \cellcolor[rgb]{ .98,  .98,  .98} 5$\;\;\bullet$ & \cellcolor[rgb]{ .894,  .894,  .894} 4$\;\;\bullet$ & \cellcolor[rgb]{ .718,  .718,  .718} 2$\;\;\bullet$ & \cellcolor[rgb]{ .627,  .627,  .627} 1 \\
\midrule 
$\fCigar$ & 2500 & \cellcolor[rgb]{ .98,  .98,  .98} 5$\;\;\bullet$ & \cellcolor[rgb]{ .894,  .894,  .894} 4$\;\;\bullet$ & \cellcolor[rgb]{ .627,  .627,  .627} 1$\;\;\circ$ & \cellcolor[rgb]{ .804,  .804,  .804} 3$\;\;\bullet$ & \cellcolor[rgb]{ .718,  .718,  .718} 2 \\
  & 5000 & \cellcolor[rgb]{ .98,  .98,  .98} 5$\;\;\bullet$ & \cellcolor[rgb]{ .894,  .894,  .894} 4$\;\;\bullet$ & \cellcolor[rgb]{ .627,  .627,  .627} 1$\;\;\circ$ & \cellcolor[rgb]{ .804,  .804,  .804} 3$\;\;\bullet$ & \cellcolor[rgb]{ .718,  .718,  .718} 2 \\
  & 7500 & \cellcolor[rgb]{ .98,  .98,  .98} 5$\;\;\bullet$ & \cellcolor[rgb]{ .894,  .894,  .894} 4$\;\;\bullet$ & \cellcolor[rgb]{ .627,  .627,  .627} 1$\;\;\circ$ & \cellcolor[rgb]{ .804,  .804,  .804} 3$\;\;\bullet$ & \cellcolor[rgb]{ .718,  .718,  .718} 2 \\
  & 10000 & \cellcolor[rgb]{ .98,  .98,  .98} 5$\;\;\bullet$ & \cellcolor[rgb]{ .894,  .894,  .894} 4$\;\;\bullet$ & \cellcolor[rgb]{ .627,  .627,  .627} 1$\;\;\circ$ & \cellcolor[rgb]{ .804,  .804,  .804} 3$\;\;\bullet$ & \cellcolor[rgb]{ .718,  .718,  .718} 2 \\
\midrule 
$\fDiffPow$ & 2500 & \cellcolor[rgb]{ .804,  .804,  .804} 3$\;\;\bullet$ & \cellcolor[rgb]{ .627,  .627,  .627} 1$\;\;\circ$ & \cellcolor[rgb]{ .98,  .98,  .98} 5$\;\;\bullet$ & \cellcolor[rgb]{ .894,  .894,  .894} 4$\;\;\bullet$ & \cellcolor[rgb]{ .718,  .718,  .718} 2 \\
  & 5000 & \cellcolor[rgb]{ .804,  .804,  .804} 3$\;\;\bullet$ & \cellcolor[rgb]{ .627,  .627,  .627} 1$\;\;\circ$ & \cellcolor[rgb]{ .98,  .98,  .98} 5$\;\;\bullet$ & \cellcolor[rgb]{ .894,  .894,  .894} 4$\;\;\bullet$ & \cellcolor[rgb]{ .718,  .718,  .718} 2 \\
  & 7500 & \cellcolor[rgb]{ .804,  .804,  .804} 3$\;\;\bullet$ & \cellcolor[rgb]{ .627,  .627,  .627} 1$\;\;\circ$ & \cellcolor[rgb]{ .98,  .98,  .98} 5$\;\;\bullet$ & \cellcolor[rgb]{ .894,  .894,  .894} 4$\;\;\bullet$ & \cellcolor[rgb]{ .718,  .718,  .718} 2 \\
  & 10000 & \cellcolor[rgb]{ .804,  .804,  .804} 3$\;\;\bullet$ & \cellcolor[rgb]{ .627,  .627,  .627} 1$\;\;\circ$ & \cellcolor[rgb]{ .98,  .98,  .98} 5$\;\;\bullet$ & \cellcolor[rgb]{ .894,  .894,  .894} 4$\;\;\bullet$ & \cellcolor[rgb]{ .718,  .718,  .718} 2 \\
\midrule
$\bullet$ / $\circ$ / $\dagger$ &   & 16 / 0 / 4 & 14 / 6 / 0 & 12 / 4 / 4 & 16 / 1 / 3 &  \\
\midrule
Avg Rank &   & 3.65 & 3.35 & 3.35 & 2.75 & 1.9 \\
\midrule
$p$-Value &   & 0.00 & 0.04 & 0.04 & 0.89 &  \\
\bottomrule
\end{tabular}%

\end{table}

\subsection{Runtime}
\label{ss:runtime}
We verify the practical efficiency of MMES in terms of runtime. The runtime of an algorithm refers to its consumed CPU time divided by the number of function evaluations. We carry out the simulation on $\fElli$. However, to make the results independent of the specification of the test problems, the processing time for function evaluations are excluded from the measurement. 

\begin{figure}[htbp]
\centering
\includegraphics[width=0.4\textwidth]{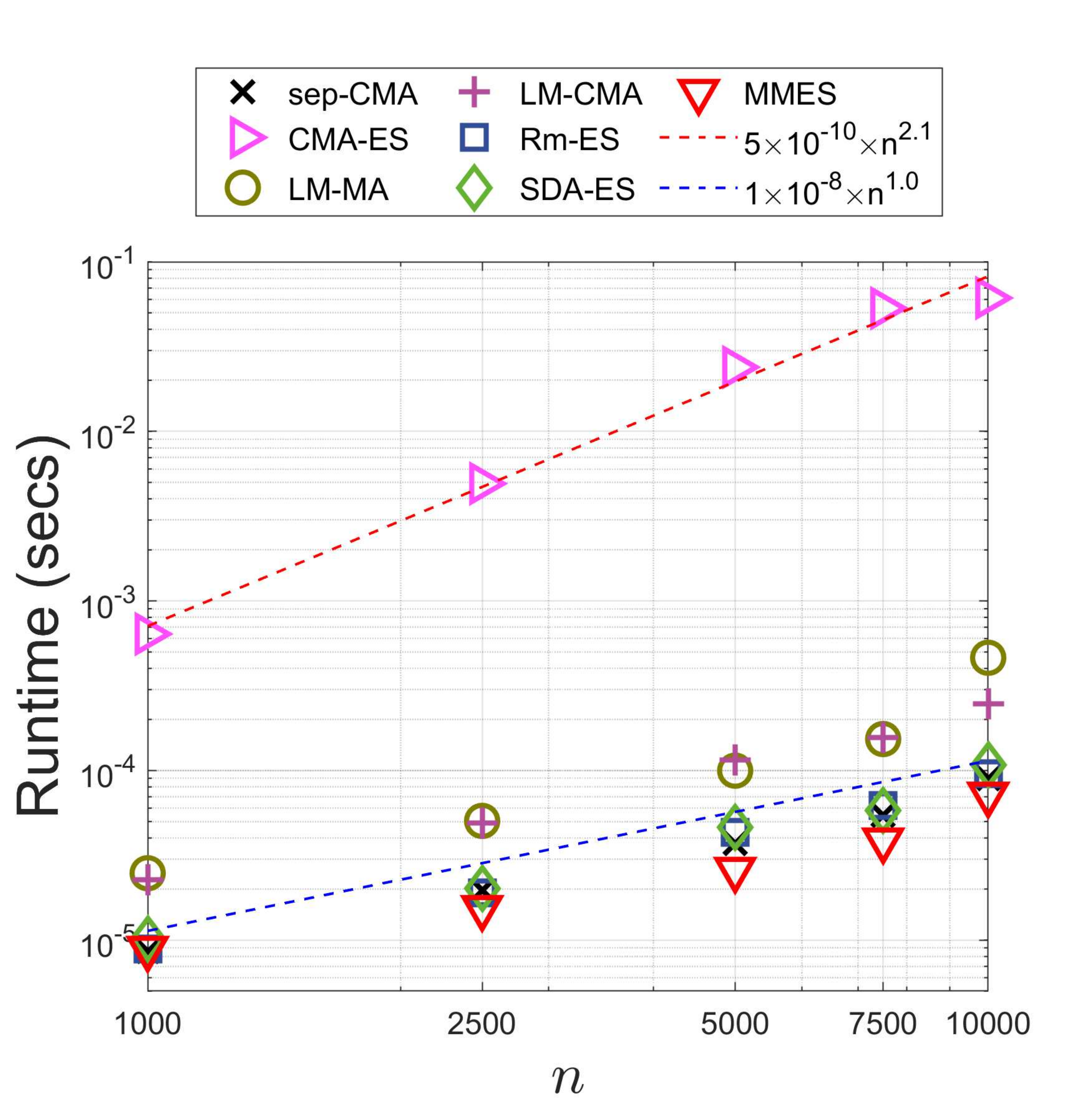} 
\caption{Runtime results on the $f_{\text{Elli}}$ problem. The dotted lines are obtained by the log-log linear regression.} 
\label{fig:running-time}
\end{figure}

\Cref{fig:running-time} provides a visual comparison for the runtime results obtained on a PC with a 3.10-GHz Intel Core i5-10500 CPU. It is clear that CMA-ES is the slowest. The regression line fitted by its associated data points, depicted by a red dotted line, suggests that its runtime grows at the order of $O(n^{2.1})$, being consistent with the fact that a full covariance matrix are required to be adapted. sep-CMA, RmES, SDA-ES, and MMES are the fastest on all dimension settings and are about 500 times faster than CMA-ES in the 10000-dimensional case. 
LM-MA and LM-CMA are slower, but their runtimes only increase by a small constant. The regression line fitted for all these variants, shown by a blue dotted line, states that their asymptotic runtimes can be well approximated by a polynomial of order 1.0. This leads to the conclusion that CMA-ES scales approximately quadratically while MMES and all the others scale approximately linearly. But note that the obtained results can heavily depend on programming skills and operating environments. In general, all the considered CMA-ES variants are time-efficient in large-scale settings.

\subsection{Sensitiveness to Mixing Strength}
\label{ss:sensitiveness-to-mixing-strength}
The mixing strength $l$ is a key parameter in the proposed MMES algorithm. A larger mixing strength leads to a more accurate probability model, but in the meanwhile, makes the runtime longer. The analysis in~\Cref{ss:approximation-accuracy-analysis} shows that a relatively small $l$ is enough for practical use and thus we verify this statement with simulations.

\begin{figure}[tb]
\centering
\subfloat[$\fDiscus$]{\includegraphics[width=0.24\textwidth]{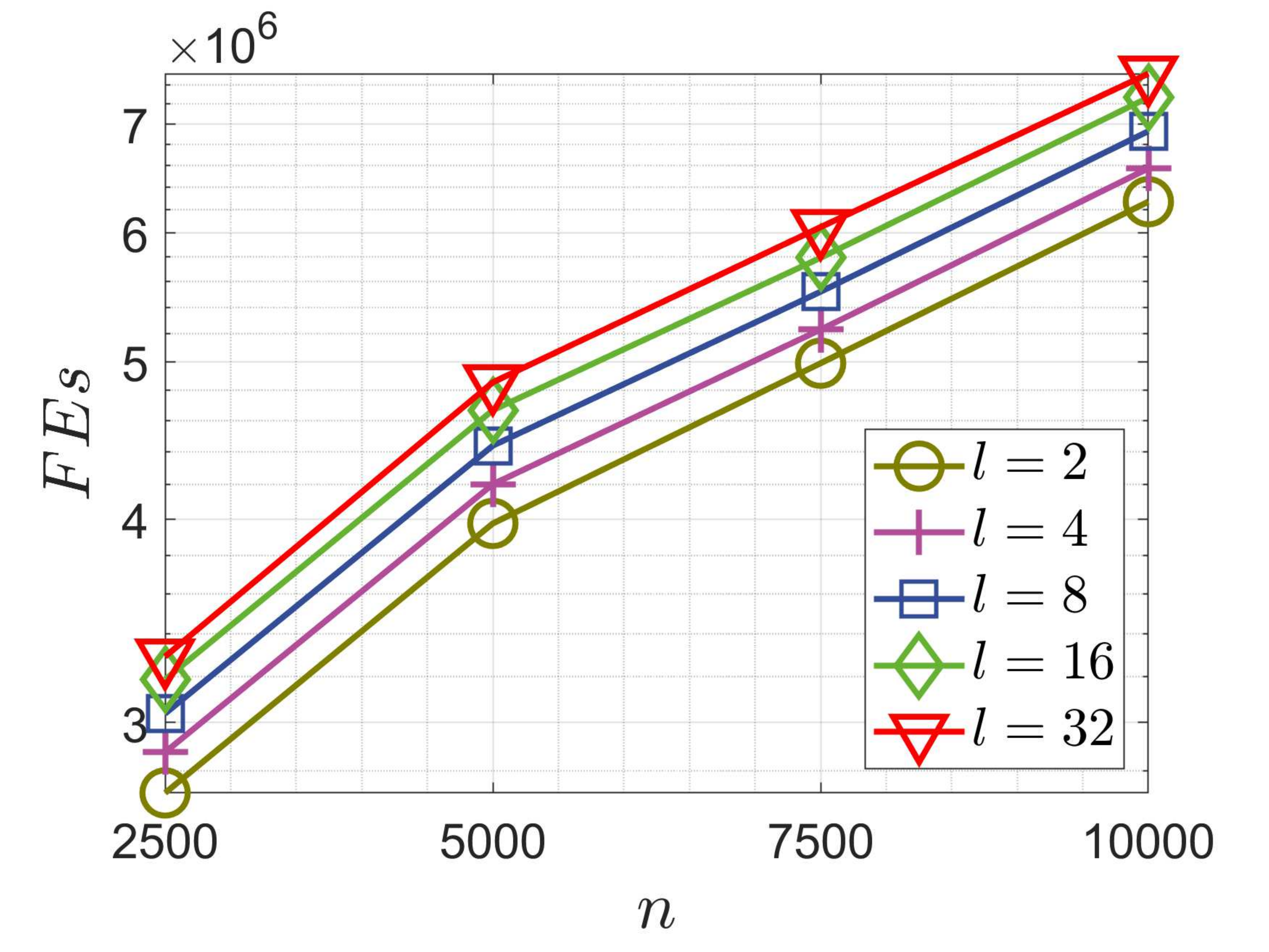} \label{subfig:parameter-test-Discus}}
\subfloat[$\fElli$]{\includegraphics[width=0.24\textwidth]{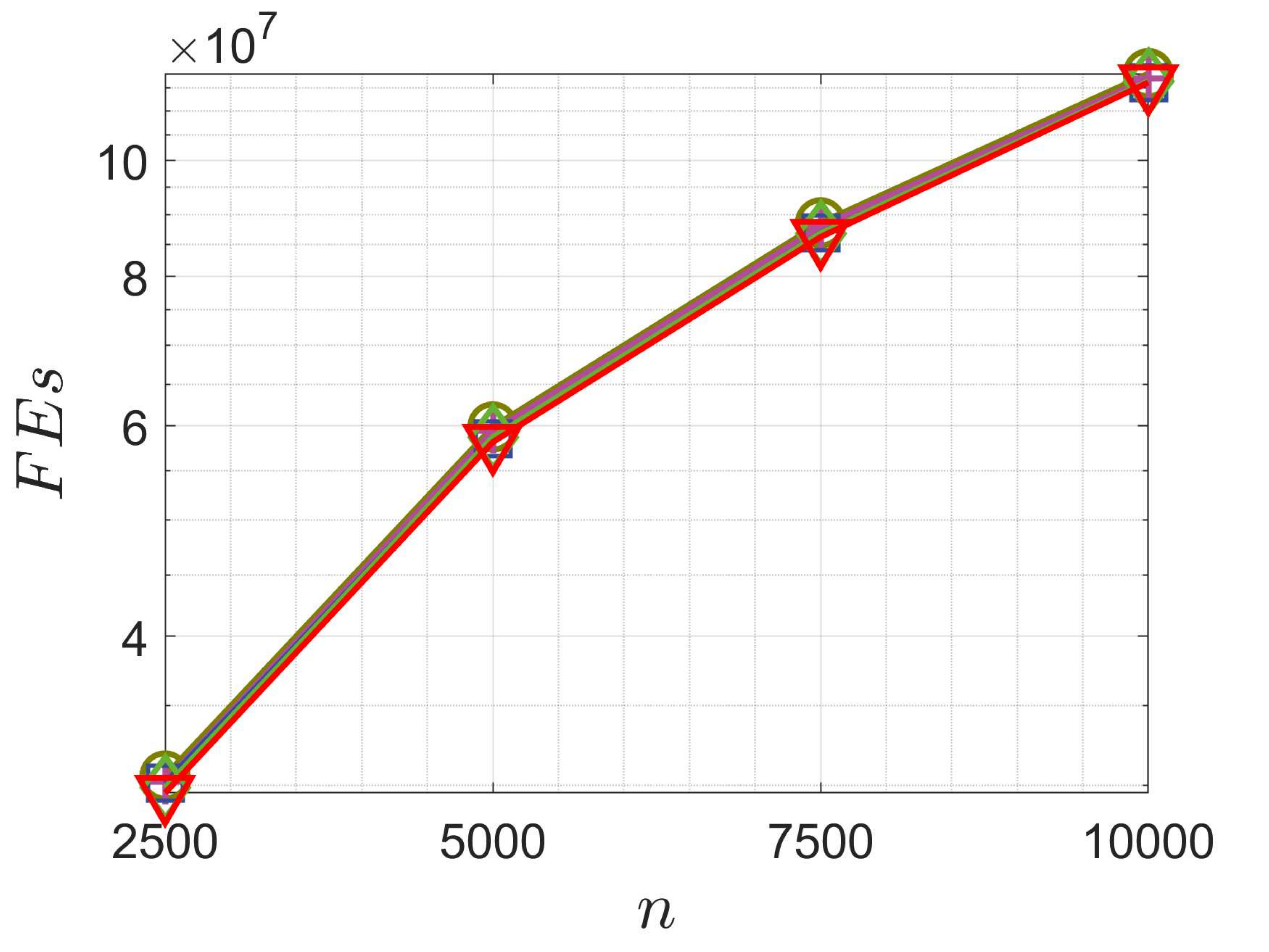} \label{subfig:parameter-test-Elli}}
\hfil
\subfloat[$\fCigar$]{\includegraphics[width=0.24\textwidth]{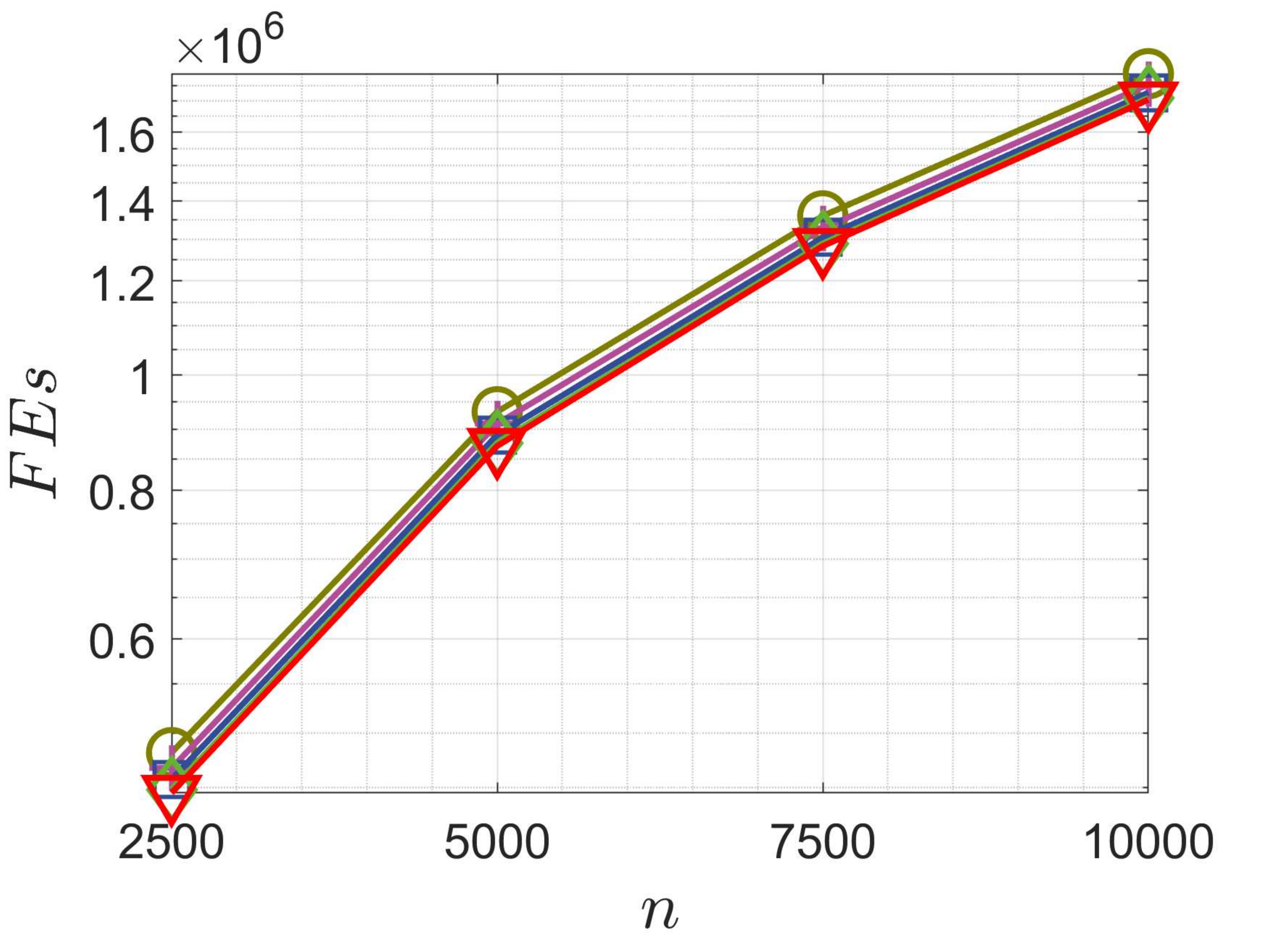} \label{subfig:parameter-test-Cigar}}
\subfloat[$\fDiffPow$]{\includegraphics[width=0.24\textwidth]{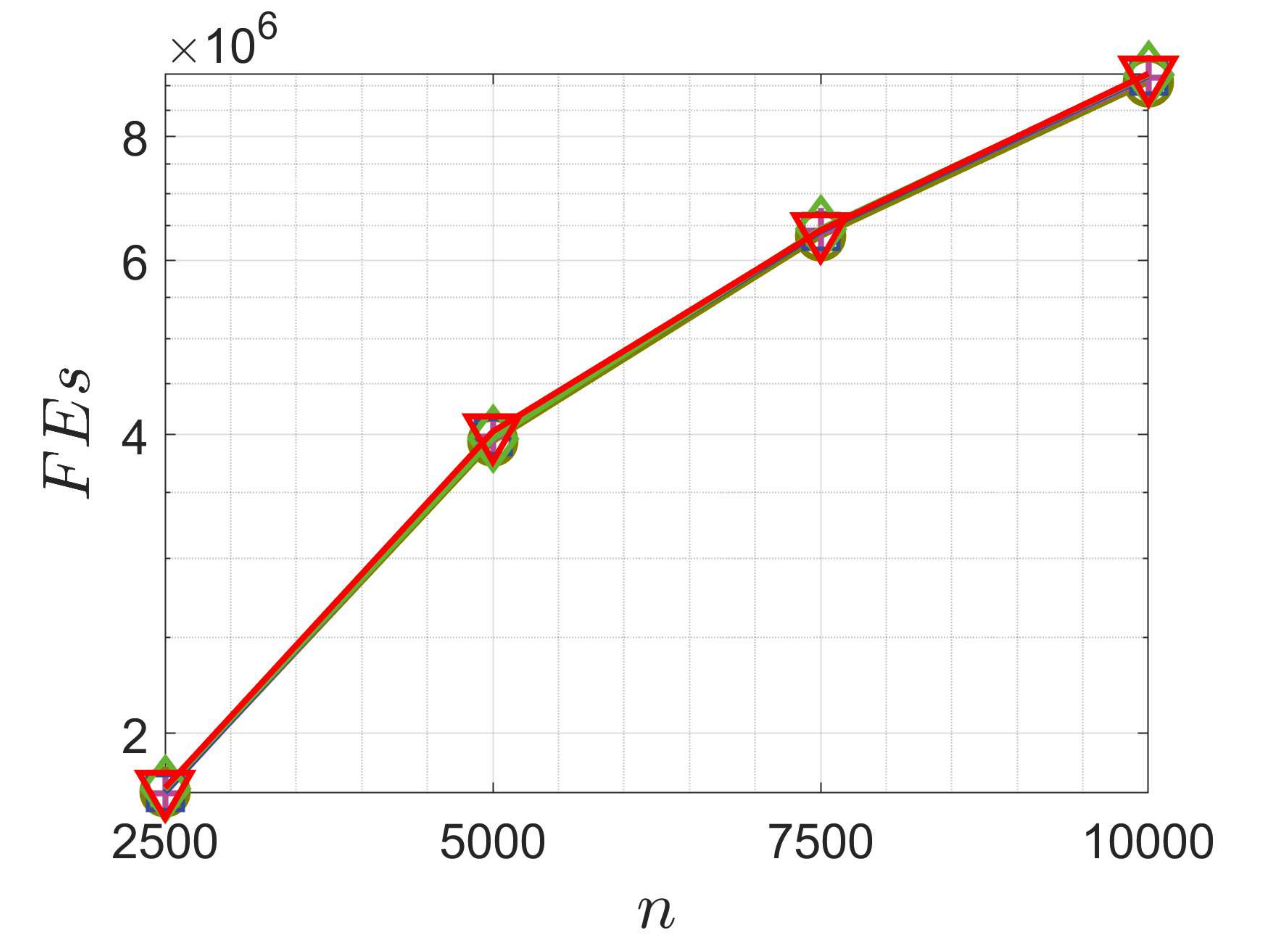} \label{subfig:parameter-test-DiffPow}}
\caption{Sensitiveness of MMES to different settings of $l$. The curves present the number of function evaluations required to reach the accuracy $10^{-8}$.}
\label{fig:parameter-test}
\end{figure}

We choose four unimodal test problems, namely $\fElli$, $\fDiscus$, $\fCigar$, and $\fDiffPow$, and investigate the scalability of MMES when different settings of $l$ are used. 
\Cref{fig:parameter-test} shows the median results produced by MMES when $l$ is chosen from $\{2, 4, 8, 16, 32\}$. We observe that $l=2$ is sufficient to solve $\fDiscus$ while increasing $l$ can even deteriorate the performance. As discussed in~\Cref{ss:scalability}, this is probably due to the fat-tailed distribution that has a higher probability of generating long jumps on insensitive search directions. 
This is not the case for $\fDiffPow$ where the settings of $l$ seem to have no influence on the performance. 
On $\fElli$ and $\fCigar$ whose landscapes can be well learned by the approximate distribution, it is observed that the performance can be improved by increasing $l$. This is consistent with the theoretical result that a larger mixing strength leads to a better approximation. Nevertheless, this does not mean a larger mixing strength is necessary, as the performance improvement seems to be quite insignificant. Thus, the default setting $l=4$ would be a reasonable choice for balancing the approximation accuracy and the sampling efficiency.

\subsection{Performance on CEC'2010 LSGO Problems}
The CEC'2010 problems from the second test set are characterized with controllable non-separability and multimodality. The aim of the experiments on this test set is to 1) verify the potential of MMES as a generic global optimizer and 2) demonstrate its pros and cons compared with non-ES-based algorithms. Note that, ESs are usually designed for local search and should be used with restart strategies in order to solve multimodal problems~\cite{auger_restart_2005}. Thus, in this part of experiments, we restart MMES whenever a stagnation is detected. An instance of MMES is considered as stagnated if the improvement of the objective values during the last $n$ generation is smaller than $10^{-8}$. At each restart, we increase both the population size and the damping factor by a factor of 2, according to the suggestions from~\cite{li_simple_2017}. 
This encourages MMES to explore the global structure of the landscape and prevents being trapped into local optima, thereby improving the global exploration ability.

\Cref{tab:CEC2010} summarizes the ranking results on the 1000-dimensional CEC'2010 LSGO test problems. The multiple comparison based on the Friedman test and the Bonferroni post hoc procedure shows that MMES ranks first among all algorithms and is superior to the three CC-based algorithms (i.e., DECC-G, CCPSO2, and DECC-DG). The Wilcoxon test also shows that it is significantly better than the other competitors on the majority of the test problems. In particular, MMES performs the best or the second best on highly multimodal problems with a huge number of local optima (e.g., $f_{15}$ and $f_{16}$). These results suggest that the MMES with restarts can serve as a competitive alternative to existing large-scale global optimization algorithms.

The simulation provides some hints about how MMES behaves differently from the other algorithms that are not based on the ES framework. For example, MMES is more robust to the non-separability than the CC-based algorithms, as the latter perform well on the fully separable problems but suffer from an obvious performance degradation on the other problems. 
The effectiveness of the CC-based algorithms is also limited by the degrees of the non-separability, as we see on the fully separable or fully non-separable problems that a more accurate grouping strategy (i.e., that in DECC-DG) does not guarantee achieving a better performance than the purely random grouping strategy (i.e., that in DECC-G). 
MA-SW and MOS perform generally better than the CC-based algorithms and can produce quite good results on the fully separable problems. This is due to their local search strategies performed individually on each variable in the way that the difference in the scalings of the coordinates can be easily explored. Nevertheless, they exhibit the strong dependence on the separable topology of the problems. MMES, on the contrary, does not explore the separable structure of the problem landscape and is less sensitive to the non-separability. In fact, the more non-separable subgroups are, the greater advantage over the others MMES possesses. This can be observed on the Ellipsoid problem where MMES performs the worst on the fully separable instance (i.e., $f_1$) but becomes the best performer whenever the non-separability is introduced (i.e., on $f_4, f_9$, or $f_{14}$).

The detailed numerical results are found in Table S-III in the supplement.

\begin{table}[htb]
\centering
\scriptsize
\caption{Ranks of different algorithms, in terms of the final objective values, on the 1000-dimensional CEC'2010 LSGO problems. Better results are highlighted with darker gray background.}
\setlength{\tabcolsep}{4pt}
\renewcommand\arraystretch{1}
\begin{tabular}{ccccccc}
\toprule
  & DECC-G & MA-SW & MOS & CCPSO2 & DECC-DG & MMES \\
\midrule
$f_{1}$ & \cellcolor[rgb]{ .769,  .769,  .769} 3$\;\;\circ$ & \cellcolor[rgb]{ .698,  .698,  .698} 2$\;\;\circ$ & \cellcolor[rgb]{ .627,  .627,  .627} 1$\;\;\circ$ & \cellcolor[rgb]{ .839,  .839,  .839} 4$\;\;\circ$ & \cellcolor[rgb]{ .91,  .91,  .91} 5$\;\;\circ$ & \cellcolor[rgb]{ .98,  .98,  .98} 6 \\
$f_{2}$ & \cellcolor[rgb]{ .91,  .91,  .91} 5$\;\;\bullet$ & \cellcolor[rgb]{ .839,  .839,  .839} 4$\;\;\bullet$ & \cellcolor[rgb]{ .698,  .698,  .698} 2$\;\;\circ$ & \cellcolor[rgb]{ .627,  .627,  .627} 1$\;\;\circ$ & \cellcolor[rgb]{ .98,  .98,  .98} 6$\;\;\bullet$ & \cellcolor[rgb]{ .769,  .769,  .769} 3 \\
$f_{3}$ & \cellcolor[rgb]{ .839,  .839,  .839} 4$\;\;\bullet$ & \cellcolor[rgb]{ .698,  .698,  .698} 2$\;\;\bullet$ & \cellcolor[rgb]{ .91,  .91,  .91} 5$\;\;\bullet$ & \cellcolor[rgb]{ .769,  .769,  .769} 3$\;\;\bullet$ & \cellcolor[rgb]{ .98,  .98,  .98} 6$\;\;\bullet$ & \cellcolor[rgb]{ .627,  .627,  .627} 1 \\
$f_{4}$ & \cellcolor[rgb]{ .98,  .98,  .98} 6$\;\;\bullet$ & \cellcolor[rgb]{ .769,  .769,  .769} 3$\;\;\bullet$ & \cellcolor[rgb]{ .698,  .698,  .698} 2$\;\;\bullet$ & \cellcolor[rgb]{ .839,  .839,  .839} 4$\;\;\bullet$ & \cellcolor[rgb]{ .91,  .91,  .91} 5$\;\;\bullet$ & \cellcolor[rgb]{ .627,  .627,  .627} 1 \\
$f_{5}$ & \cellcolor[rgb]{ .839,  .839,  .839} 4$\;\;\bullet$ & \cellcolor[rgb]{ .769,  .769,  .769} 3$\;\;\bullet$ & \cellcolor[rgb]{ .98,  .98,  .98} 6$\;\;\bullet$ & \cellcolor[rgb]{ .91,  .91,  .91} 5$\;\;\bullet$ & \cellcolor[rgb]{ .698,  .698,  .698} 2$\;\;\bullet$ & \cellcolor[rgb]{ .627,  .627,  .627} 1 \\
$f_{6}$ & \cellcolor[rgb]{ .839,  .839,  .839} 4$\;\;\bullet$ & \cellcolor[rgb]{ .627,  .627,  .627} 1$\;\;\circ$ & \cellcolor[rgb]{ .98,  .98,  .98} 6$\;\;\bullet$ & \cellcolor[rgb]{ .91,  .91,  .91} 5$\;\;\bullet$ & \cellcolor[rgb]{ .698,  .698,  .698} 2$\;\;\circ$ & \cellcolor[rgb]{ .769,  .769,  .769} 3 \\
$f_{7}$ & \cellcolor[rgb]{ .98,  .98,  .98} 6$\;\;\bullet$ & \cellcolor[rgb]{ .698,  .698,  .698} 2$\;\;\circ$ & \cellcolor[rgb]{ .627,  .627,  .627} 1$\;\;\circ$ & \cellcolor[rgb]{ .91,  .91,  .91} 5$\;\;\bullet$ & \cellcolor[rgb]{ .769,  .769,  .769} 3$\;\;\circ$ & \cellcolor[rgb]{ .839,  .839,  .839} 4 \\
$f_{8}$ & \cellcolor[rgb]{ .98,  .98,  .98} 6$\;\;\bullet$ & \cellcolor[rgb]{ .769,  .769,  .769} 3$\;\;\bullet$ & \cellcolor[rgb]{ .627,  .627,  .627} 1$\;\;\circ$ & \cellcolor[rgb]{ .91,  .91,  .91} 5$\;\;\bullet$ & \cellcolor[rgb]{ .839,  .839,  .839} 4$\;\;\bullet$ & \cellcolor[rgb]{ .698,  .698,  .698} 2 \\
$f_{9}$ & \cellcolor[rgb]{ .98,  .98,  .98} 6$\;\;\bullet$ & \cellcolor[rgb]{ .769,  .769,  .769} 3$\;\;\bullet$ & \cellcolor[rgb]{ .698,  .698,  .698} 2$\;\;\bullet$ & \cellcolor[rgb]{ .91,  .91,  .91} 5$\;\;\bullet$ & \cellcolor[rgb]{ .839,  .839,  .839} 4$\;\;\bullet$ & \cellcolor[rgb]{ .627,  .627,  .627} 1 \\
$f_{10}$ & \cellcolor[rgb]{ .98,  .98,  .98} 6$\;\;\bullet$ & \cellcolor[rgb]{ .698,  .698,  .698} 2$\;\;\bullet$ & \cellcolor[rgb]{ .91,  .91,  .91} 5$\;\;\bullet$ & \cellcolor[rgb]{ .839,  .839,  .839} 4$\;\;\bullet$ & \cellcolor[rgb]{ .769,  .769,  .769} 3$\;\;\bullet$ & \cellcolor[rgb]{ .627,  .627,  .627} 1 \\
$f_{11}$ & \cellcolor[rgb]{ .769,  .769,  .769} 3$\;\;\dagger$ & \cellcolor[rgb]{ .839,  .839,  .839} 4$\;\;\bullet$ & \cellcolor[rgb]{ .98,  .98,  .98} 6$\;\;\bullet$ & \cellcolor[rgb]{ .91,  .91,  .91} 5$\;\;\bullet$ & \cellcolor[rgb]{ .627,  .627,  .627} 1$\;\;\circ$ & \cellcolor[rgb]{ .698,  .698,  .698} 2 \\
$f_{12}$ & \cellcolor[rgb]{ .98,  .98,  .98} 6$\;\;\bullet$ & \cellcolor[rgb]{ .769,  .769,  .769} 3$\;\;\bullet$ & \cellcolor[rgb]{ .627,  .627,  .627} 1$\;\;\dagger$ & \cellcolor[rgb]{ .91,  .91,  .91} 5$\;\;\bullet$ & \cellcolor[rgb]{ .839,  .839,  .839} 4$\;\;\bullet$ & \cellcolor[rgb]{ .627,  .627,  .627} 1 \\
$f_{13}$ & \cellcolor[rgb]{ .91,  .91,  .91} 5$\;\;\bullet$ & \cellcolor[rgb]{ .698,  .698,  .698} 2$\;\;\bullet$ & \cellcolor[rgb]{ .769,  .769,  .769} 3$\;\;\bullet$ & \cellcolor[rgb]{ .839,  .839,  .839} 4$\;\;\bullet$ & \cellcolor[rgb]{ .98,  .98,  .98} 6$\;\;\bullet$ & \cellcolor[rgb]{ .627,  .627,  .627} 1 \\
$f_{14}$ & \cellcolor[rgb]{ .98,  .98,  .98} 6$\;\;\bullet$ & \cellcolor[rgb]{ .769,  .769,  .769} 3$\;\;\bullet$ & \cellcolor[rgb]{ .698,  .698,  .698} 2$\;\;\bullet$ & \cellcolor[rgb]{ .839,  .839,  .839} 4$\;\;\bullet$ & \cellcolor[rgb]{ .91,  .91,  .91} 5$\;\;\bullet$ & \cellcolor[rgb]{ .627,  .627,  .627} 1 \\
$f_{15}$ & \cellcolor[rgb]{ .91,  .91,  .91} 5$\;\;\bullet$ & \cellcolor[rgb]{ .698,  .698,  .698} 2$\;\;\bullet$ & \cellcolor[rgb]{ .98,  .98,  .98} 6$\;\;\bullet$ & \cellcolor[rgb]{ .839,  .839,  .839} 4$\;\;\bullet$ & \cellcolor[rgb]{ .769,  .769,  .769} 3$\;\;\bullet$ & \cellcolor[rgb]{ .627,  .627,  .627} 1 \\
$f_{16}$ & \cellcolor[rgb]{ .769,  .769,  .769} 3$\;\;\bullet$ & \cellcolor[rgb]{ .839,  .839,  .839} 4$\;\;\bullet$ & \cellcolor[rgb]{ .91,  .91,  .91} 5$\;\;\bullet$ & \cellcolor[rgb]{ .91,  .91,  .91} 5$\;\;\bullet$ & \cellcolor[rgb]{ .627,  .627,  .627} 1$\;\;\circ$ & \cellcolor[rgb]{ .698,  .698,  .698} 2 \\
$f_{17}$ & \cellcolor[rgb]{ .98,  .98,  .98} 6$\;\;\bullet$ & \cellcolor[rgb]{ .769,  .769,  .769} 3$\;\;\bullet$ & \cellcolor[rgb]{ .698,  .698,  .698} 2$\;\;\bullet$ & \cellcolor[rgb]{ .91,  .91,  .91} 5$\;\;\bullet$ & \cellcolor[rgb]{ .839,  .839,  .839} 4$\;\;\bullet$ & \cellcolor[rgb]{ .627,  .627,  .627} 1 \\
$f_{18}$ & \cellcolor[rgb]{ .91,  .91,  .91} 5$\;\;\bullet$ & \cellcolor[rgb]{ .698,  .698,  .698} 2$\;\;\bullet$ & \cellcolor[rgb]{ .839,  .839,  .839} 4$\;\;\bullet$ & \cellcolor[rgb]{ .769,  .769,  .769} 3$\;\;\bullet$ & \cellcolor[rgb]{ .98,  .98,  .98} 6$\;\;\bullet$ & \cellcolor[rgb]{ .627,  .627,  .627} 1 \\
$f_{19}$ & \cellcolor[rgb]{ .839,  .839,  .839} 4$\;\;\bullet$ & \cellcolor[rgb]{ .769,  .769,  .769} 3$\;\;\bullet$ & \cellcolor[rgb]{ .698,  .698,  .698} 2$\;\;\bullet$ & \cellcolor[rgb]{ .91,  .91,  .91} 5$\;\;\bullet$ & \cellcolor[rgb]{ .98,  .98,  .98} 6$\;\;\bullet$ & \cellcolor[rgb]{ .627,  .627,  .627} 1 \\
$f_{20}$ & \cellcolor[rgb]{ .91,  .91,  .91} 5$\;\;\bullet$ & \cellcolor[rgb]{ .769,  .769,  .769} 3$\;\;\bullet$ & \cellcolor[rgb]{ .698,  .698,  .698} 2$\;\;\bullet$ & \cellcolor[rgb]{ .839,  .839,  .839} 4$\;\;\bullet$ & \cellcolor[rgb]{ .98,  .98,  .98} 6$\;\;\bullet$ & \cellcolor[rgb]{ .627,  .627,  .627} 1 \\
\midrule
$\bullet$ / $\circ$ / $\dagger$ & 18 / 1 / 1 & 17 / 3 / 0 & 15 / 4 / 1 & 18 / 2 / 0 & 15 / 5 / 0 &  \\
\midrule
Avg Rank & 4.9 & 2.7 & 3.25 & 4.28 & 4.1 & 1.77 \\
\midrule
$p$-Value & 0 & 1 & 0.188 & 0.0003 & 0.0012 &  \\
\bottomrule
\end{tabular}%

\label{tab:CEC2010}%
\end{table}%

\section{Conclusion}
In this paper, we develop a large-scale CMA-ES variant, named MMES, based on a mixture modeling method for solution sampling and a paired test method for mutation strength adaptation. MMES permits reconstructing the probability model from an arbitrarily large number of direction vectors in limited operation time such that the rich variable correlations can be exploited efficiently. The theoretical analyses show that MMES well approximates CMA-ES in terms of the underlying probability model while the numerical simulations suggest that MMES demonstrates state-of-the-art performance in both local and global optimization.

Further development of MMES will be two-directional. The first direction is to develop new mechanisms for adapting the direction vectors. The current version of MMES does not involve this aspect but directly applies well established techniques. However, the proposed mixture modeling method should not be limited in the sampling process; using it to accelerate the distribution adaptation would also be an interesting idea. The second direction is to establish the convergence theorem for MMES. The PTA method in MMES relies on the comparison of the populations in consecutive generations, and thus, implies some types of elitism. We plan to search for an efficient indicator to quantify the population which may guarantee sufficient descents and lead to a global convergence theorem. 

\ifCLASSOPTIONcaptionsoff
  \newpage
\fi

\bibliographystyle{IEEEtran}
\bibliography{my}

 \appendices

\section{Proof of Lemma 1}
\begin{proof} ~\
Denote 
\begin{itemize}
	\item $p(\bm{z})$ as the probability density function (p.d.f.) value of a mutation vector $\bm{z}$ sampled from $\mathcal{P}_m$,
	\item $\phi(\bm{z}|\bm{\Sigma_i})$ as the conditional p.d.f. of $\mathcal{P}_{\bm{\Sigma_i}}$,
	\item $p(\bm{i})=\prod_{j=1}^l p(i_j)$ as the joint probability of choosing $i_1,\cdots,i_l$ from $\mathcal{P}_{\bm{i}}$,
	\item $[m]$ as the set $\{1,2,\cdots,m\}$.
\end{itemize}

By the law of total probability, $\mathcal{P}_m$ can be expressed with $\mathcal{P}_{\bm{\Sigma_i}}$ and $\mathcal{P}_{\bm{i}}$ using the relation
\begin{equation}
p(\bm{z}) = \sum_{\bm{i} \in [m]^l} p(\bm{i})\phi(\bm{z}|\bm{\Sigma_i})
\label{eq:pdf-z}
\end{equation} 

Then, we have
\begin{equation*}
\begin{split}
M(\bm{t}) &= E[\exp(\bm{t}^T \bm{z})] \\
		&= \int \exp(\bm{t}^T \bm{z}) p(\bm{z}) d\bm{z} \\
		&= \sum_{\bm{i} \in [m]^l} p(\bm{i}) \int \exp(\bm{t}^T \bm{z}) \phi(\bm{z}|\bm{\Sigma_i}) d\bm{z} \\
		&= \sum_{\bm{i} \in [m]^l} p(\bm{i}) E_{\bm{i}}[\exp(\bm{t}^T \bm{z})] \\
\end{split}
\end{equation*}
where $E[\cdot]$ denotes the expectation with respect to $\bm{z}$ and $E_{\bm{i}}[\cdot]$ denotes the expectation with respect to $\bm{z}$ conditioned on $\bm{i}$.

$E_{\bm{i}}[\exp(\bm{t}^T \bm{z})]$ in the last equation equals to the moment generating function of $\mathcal{P}_{\bm{\Sigma_i}}$, so we have 
\begin{equation*}
\begin{split}
M(\bm{t}) &= \sum_{\bm{i} \in [m]^l} p(\bm{i}) \exp \left(\frac{1}{2}\bm{t}^T\bm{\Sigma_i} \bm{t}\right)
\end{split}
\end{equation*}

Substituting \Cref{eq:conditioned-covariance-matrix} into this yields
\begin{equation}
\begin{split}
M(\bm{t}) &= \sum_{\bm{i} \in [m]^l} p(\bm{i}) \exp \left(\frac{1}{2}(1-\gamma)|\bm{t}|^2 + \frac{\gamma}{2l} \sum_{j=1}^l (\bm{t}^T \bm{q}_{i_j})^2 \right) \\
&= \exp \left( \frac{1}{2}(1-\gamma)|\bm{t}|^2\right) \sum_{\bm{i} \in [m]^l} p(\bm{i}) \prod_{j=1}^l \exp \left(\frac{\gamma}{2l} (\bm{t}^T \bm{q}_{i_j})^2 \right)
\end{split}
\label{eq:moment-generating-function}
\end{equation}

We immediately reach the conclusion if $l=1$. When $l>1$, the second multiplier of the last equation in \Cref{eq:moment-generating-function} can be written in the following recursive form:
\begin{equation*}
\begin{split}
& \sum_{\bm{i} \in [m]^l} p(\bm{i}) \prod_{j=1}^l \exp \left(\frac{\gamma}{2l} (\bm{t}^T \bm{q}_{i_j})^2 \right) \\
& = \sum_{\substack{\bm{i} \in [m]^{l-1} \\ i_l \in[m]}} p(\bm{i},i_l) \prod_{j=1}^{l} \exp \left(\frac{\gamma}{2l} (\bm{t}^T \bm{q}_{i_j})^2 \right) \\
& = \sum_{\substack{\bm{i} \in [m]^{l-1} \\ i_l \in[m]}} p(\bm{i}) \prod_{j=1}^{l-1} \exp \left(\frac{\gamma}{2l} (\bm{t}^T \bm{q}_{i_j})^2 \right) p(i_l) \exp \left(\frac{\gamma}{2l} (\bm{t}^T \bm{q}_{i_l})^2 \right) \\
& = \left(\sum_{\bm{i} \in [m]^{l-1}} p(\bm{i}) \prod_{j=1}^{l-1} \exp \left(\frac{\gamma}{2l} (\bm{t}^T \bm{q}_{i_j})^2 \right) \right) \cdot \\
& \;\;\;\;\left(\sum_{i_l\in[m]} p(i_l) \exp \left(\frac{\gamma}{2l} (\bm{t}^T \bm{q}_{i_l})^2 \right)\right)
\end{split}
\end{equation*}

Repeat this for $l-1$ times and we obtain
\begin{equation*}
\begin{split}
& \sum_{\bm{i} \in [m]^l} p(\bm{i}) \prod_{j=1}^l \exp \left(\frac{\gamma}{2l} (\bm{t}^T \bm{q}_{i_j})^2 \right) \\
& = \prod_{j=1}^l \left(\sum_{i_j\in[m]} p(i_j) \exp \left(\frac{\gamma}{2l} (\bm{t}^T \bm{q}_{i_j})^2 \right)\right) \\
& = \prod_{j=1}^l \left(\sum_{j\in[m]} \alpha_j \exp \left(\frac{\gamma}{2l} (\bm{t}^T \bm{q}_{j})^2 \right)\right)
\end{split}
\end{equation*}

Substitute this into \Cref{eq:moment-generating-function} and the target form of the moment generating function is reached.

Then, we prove that $M(\bm{t})$ uniquely determines $\mathcal{P}_m$ by showing there exists a neighborhood of $\bm{0}_n$ in which $M(\bm{t})$ is finite.

Let $\delta$ be a positive scalar. Then, for any $\bm{t}$ satisfying $|\bm{t}| \le \delta$, we have
\begin{equation*}
\begin{split}
M(\bm{t}) & \le \exp\left(\frac{1}{2}(1-\gamma)\delta^2\right) \left(\sum_{j=1}^m \alpha_j \exp\left(\frac{\gamma}{2l}\delta^2|\bm{q}_j|^2\right)\right)^l \\ 
& \le \exp\left(\frac{1}{2}(1-\gamma)\delta^2\right) \sum_{j=1}^m \alpha_j \exp\left(\frac{\gamma}{2}\delta^2|\bm{q}_j|^2\right) \\ 
\end{split}
\end{equation*}
where the second inequality is derived from the Jensen's inequality. This gives a positive radius of convergence for the moment generating function, which completes the proof. 
\end{proof}

\section{Proof of Theorem 1}
\begin{proof}
Denote the moment generating function of $\mathcal{P}_a$ by $U(\bm{t})$. 
With \Cref{eq:parameter-probability-i,eq:covariance-non-recursively-rank-1-update}, we have
\begin{equation*}
\begin{split}
U(\bm{t}) & = \exp \left( \frac{1}{2}\bm{t}^T \left( (1-c_a)^m \bm{I}_n + c_a \sum_{j=1}^m (1-c_a)^{m-j} \bm{q}_j \bm{q}_j^T \right)\bm{t} \right) \\
& = \exp \left( \frac{1}{2}\bm{t}^T \left( (1-\gamma) \bm{I}_n + \gamma \sum_{j=1}^m \alpha_j \bm{q}_j \bm{q}_j^T \right)\bm{t} \right)
\end{split}
\end{equation*}

Using the identity of \textit{zeroth weighted power mean} (i.e., $\lim \limits_{r \rightarrow 0} (\omega_1 x_1^r + \cdots + \omega_n x_n^r)^\frac{1}{r} = x_1^{\omega_1}\cdots x_n^{\omega_n}$ for $\sum_{i=1}^n \omega_i=1$ and $\omega_1,\cdots,\omega_n>0$), we have

\begin{equation*}
\begin{split}
& \lim \limits_{l \rightarrow \infty} M(\bm{t}) \\
& = \lim \limits_{l \rightarrow \infty} \exp\left(\frac{1}{2}(1-\gamma)|\bm{t}|^2\right) \left(\sum_{j=1}^m \alpha_j \exp\left(\frac{\gamma}{2l}(\bm{t}^T \bm{q}_j)^2\right)\right)^l \\
& = \lim \limits_{l \rightarrow \infty} \exp\left(\frac{1}{2}(1-\gamma)|\bm{t}|^2\right) \left(\sum_{j=1}^m \alpha_j \left(\exp\left(\frac{\gamma}{2}(\bm{t}^T \bm{q}_j)^2\right)\right)^{\frac{1}{l}}\right)^l \\
& = \exp\left(\frac{1}{2}(1-\gamma)|\bm{t}|^2\right) \prod_{j=1}^m  \left(\exp\left(\frac{\gamma}{2}(\bm{t}^T \bm{q}_j)^2\right)\right)^{\alpha_j} \\
& = \exp\left(\frac{1}{2}(1-\gamma)|\bm{t}|^2\right) \exp\left(\sum_{j=1}^m  \frac{\gamma}{2} \alpha_j (\bm{t}^T \bm{q}_j)^2\right) \\
& = \exp\left(\frac{1}{2}\bm{t}^T\left((1-\gamma) \bm{I}_n + \gamma\sum_{j=1}^m   \alpha_j \bm{q}_j \bm{q}_j^T\right)\bm{t}\right) \\
& = U(\bm{t})
\end{split}
\end{equation*}

As $\mathcal{P}_m$ can be fully characterized by its moment generating function (see \Cref{lemma:moment-generating-function-for-FMS}), we can conclude that its limiting distribution is the same as $\mathcal{P}_a$. This completes the proof. 

\end{proof}

\section{Proof of Theorem 2}
\begin{proof}
According to~\Cref{eq:pdf-z}, the covariance matrix of $\mathcal{P}_m$ can be written as the expectation of the covariance matrix of $\mathcal{P}_{\bm{\Sigma_i}}$:
\begin{equation*}
\begin{split}
Var[\bm{z}] & = E[\bm{z}\bm{z}^T] = \int \bm{z}\bm{z}^T p(\bm{z}) d\bm{z} \\
& = \int \bm{z}\bm{z}^T \sum_{\bm{i}\in[m]^l} p(\bm{i})\phi(\bm{z}|\bm{\Sigma_i}) d\bm{z} \\ 
& = \sum_{\bm{i}\in[m]^l} p(\bm{i}) \int \bm{z}\bm{z}^T \phi(\bm{z}|\bm{\Sigma_i}) d\bm{z} \\ 
& = \sum_{\bm{i}\in[m]^l} p(\bm{i}) \bm{\Sigma_i} \\
\end{split}
\end{equation*}
where $Var[\cdot]$ denotes the variance.

By \Cref{eq:conditioned-covariance-matrix}, we have
\begin{equation}
\begin{split}
Var[\bm{z}] & = \sum_{\bm{i}\in[m]^l} p(\bm{i}) \left( (1-\gamma) \bm{I}_n + \frac{\gamma}{l} \sum_{j=1}^l \bm{q}_{i_j} (\bm{q}_{i_j})^T \right) \\
& = (1-\gamma)\bm{I}_n + \frac{\gamma}{l} \sum_{\bm{i}\in[m]^l} p(\bm{i}) \sum_{j=1}^l \bm{q}_{i_j} (\bm{q}_{i_j})^T \\
\end{split}
\label{eq:tmp-1}
\end{equation}
With the same method as in proving \Cref{lemma:moment-generating-function-for-FMS}, we have 

\begin{equation}
\begin{split}
& \sum_{\bm{i}\in[m]^l} p(\bm{i}) \sum_{j=1}^l \bm{q}_{i_j} \bm{q}_{i_j}^T \\
& = \sum_{\substack{\bm{i}\in[m]^{l-1} \\ i_l \in [m]}} p(\bm{i},i_l) \left(\sum_{j=1}^{l-1} \bm{q}_{i_j} \bm{q}_{i_j}^T + \bm{q}_{i_l} \bm{q}_{i_l}^T\right)\\
& = \sum_{\bm{i}\in[m]^{l-1}} p(\bm{i}) \sum_{j=1}^{l-1} \bm{q}_{i_j} \bm{q}_{i_j}^T + \sum_{i_l \in [m]} p(i_l) \bm{q}_{i_l} \bm{q}_{i_l}^T\\
& = ... \\
& = \sum_{i_1 \in [m]} p(i_1) \bm{q}_{i_1} \bm{q}_{i_1}^T + \cdots + \sum_{i_l \in [m]} p(i_l) \bm{q}_{i_l} \bm{q}_{i_l}^T\\
& = l \sum_{j \in [m]} \alpha_j \bm{q}_{j} \bm{q}_{j}^T \\ 
\end{split}
\end{equation}
Substituting this into \Cref{eq:tmp-1} and adopting the parameters in \Cref{eq:parameter-probability-i} lead to the conclusion.

\end{proof}

\section{Proof of Theorem 3}
\begin{proof}
We first show that
\begin{equation}
\frac{M(\bm{t})}{U(\bm{t})} = 1 + O\left(\frac{1}{l}|\bm{t}|^4\right)
\end{equation}

It follows from \Cref{eq:parameter-probability-i} and \Cref{lemma:moment-generating-function-for-FMS} that

\begin{equation*}
\begin{split}
\frac{M(\bm{t})}{U(\bm{t})} & = \left( \frac{\sum_{j=1}^m \alpha_j \exp\left(\frac{\gamma}{2l}(\bm{t}^T\bm{q}_j)^2\right)}{\exp\left(\frac{\gamma}{2l}\sum_{i=1}^m \alpha_i (\bm{t}^T\bm{q}_i)^2\right)} \right)^l \\
& = \left( \sum_{j=1}^m \alpha_j \exp\left(\frac{\gamma}{2l}\left((\bm{t}^T\bm{q}_j)^2 - \sum_{i=1}^m \alpha_i (\bm{t}^T\bm{q}_i)^2\right) \right) \right)^l \\
& = \left( \sum_{j=1}^m \alpha_j \exp\left(\frac{\gamma}{2l}\bm{t}^T\left(\bm{q}_j\bm{q}_j^T - \sum_{i=1}^m \alpha_i \bm{q}_i\bm{q}_i^T\right)\bm{t} \right) \right)^l \\
\end{split}
\end{equation*}

Set $\bm{B}_j = \frac{\gamma}{2}(\bm{q}_j \bm{q}_j^T - \sum_{i=1}^m \alpha_i \bm{q}_i \bm{q}_i^T)$, then  
\begin{equation*}
\frac{M(\bm{t})}{U(\bm{t})} = \left( \sum_{j=1}^m \alpha_j \exp\left(\frac{1}{l}\bm{t}^T \bm{B}_j \bm{t} \right) \right)^l \\
\end{equation*}

Expand $\exp\left(\frac{1}{l}\bm{t}^T \bm{B}_j \bm{t} \right)$ as a Taylor series and we have
\begin{equation*}
\frac{M(\bm{t})}{U(\bm{t})} = \left( \sum_{j=1}^m \alpha_j \left( 1+ \frac{1}{l}\bm{t}^T \bm{B}_j \bm{t} + O\left(\left(\frac{1}{l}\bm{t}^T \bm{B}_j \bm{t}\right)^2\right) \right) \right)^l \\
\end{equation*}

Applying the identity 
$\sum_{j=1}^m \alpha_j \bm{B}_j = \bm{0}_{n\times n}$ (the $n\times n$ matrix of zeros)
and using the fact 
$\bm{t}^T \bm{B}_j \bm{t} = O(|\bm{t}|^2)$, we have

\begin{equation*}
\begin{split}
\frac{M(\bm{t})}{U(\bm{t})} & = \left( 1 + \sum_{j=1}^m \alpha_j \left( \frac{1}{l}\bm{t}^T \bm{B}_j \bm{t} \right)+ O\left(\left(\frac{1}{l}|\bm{t}|^2\right)^2\right)  \right)^l \\
& = \left( 1 + O\left(\frac{1}{l^2}|\bm{t}|^4\right)  \right)^l \\
& = 1 + O\left(\frac{1}{l}|\bm{t}|^4\right)  
\end{split}
\end{equation*}

Introduce a string of $k$ integers $i_1,i_2,\cdots,i_k \in \{1,\cdots,n\}$ to denote the indexes of the elements in the vector $\bm{t}$.
Let $\nabla_k$ be the operator of $k$-th order partial derivative with respective to $t_{i_1},t_{i_2},\cdots,t_{i,k}$, i.e.,

\begin{equation*}
\nabla^k M(\bm{t}) = \frac{\partial^k}{\partial t_{i_1},\partial t_{i_2},\cdots,\partial t_{i_k}} M(\bm{t})
\end{equation*}
Then $\nabla^k M(\bm{0}_n)$ is the $k$-th order moment of $\mathcal{P}_m$.

Let $\nabla_k^r$ be the operator of summing up all $r$-th order partial derivative with respective to the elements of $\bm{t}$, where the indexes are combinations of $i_1,i_2,\cdots,i_k$ taken $r$ at a time. That is,
\begin{equation*}
\nabla_k^r M(\bm{t}) = \sum \limits_{j_1,\cdots,j_r\in \{i_1,\cdots,i_k \}} \frac{\partial^r}{\partial t_{j_1}\cdots\partial t_{j_r} } M(\bm{t}).
\end{equation*}

With the above notations, we have $\nabla_k^k = \nabla^k$ and 

\begin{equation*}
\begin{split}
& \nabla^k M(\bm{t}) = \nabla^k \left(U(\bm{t})\left(1+O\left(\frac{1}{l}|\bm{t}|^4\right)\right) \right)\\
& = \nabla^k U(\bm{t}) +\nabla^k \left(U(\bm{t})O\left(\frac{1}{l}|\bm{t}|^4\right)\right) \\
& = \nabla^k U(\bm{t}) + \sum_{j=k-4}^{k} \left(\nabla^j_k U(\bm{t})\right) O\left(\frac{|\bm{t}|^{4-k+j}}{l}\right) \\
& = \nabla^k U(\bm{t}) + \sum_{j=k-3}^{k} \left(\nabla^j_k U(\bm{t})\right) O\left(\frac{|\bm{t}|^{4-k+j}}{l}\right) \\
& \;\;\;\;\;\;\;\;+ \left(\nabla^{k-4}_k U(\bm{t})\right) O\left(\frac{1}{l}\right)\\
\end{split}
\end{equation*}

Set $\bm{t}=\bm{0}_n$ on both sides and we obtain the $k$-th order moment

\begin{equation*}
\begin{split}
& \nabla^k M(\bm{0}_n) \\
& = \nabla^k U(\bm{0}_n) + \sum_{j=k-3}^{k} \left(\nabla^j_k U(\bm{0}_n)\right) O\left(\frac{|\bm{0}_n|^{4-k+j}}{l}\right) \\
& \;\;\;\;\;\;\;\;+ \left(\nabla^{k-4}_k U(\bm{0}_n)\right) O\left(\frac{1}{l}\right)\\
& = \nabla^k U(\bm{0}_n) + \left(\nabla^{k-4}_k U(\bm{0}_n)\right) O\left(\frac{1}{l}\right)\\
\end{split}
\end{equation*}

$\nabla^{k-4}_k U(\bm{0}_n)$ is finite due to the existence of the moments of $\mathcal{P}_m$. We thus obtain
\begin{equation*}
\nabla^k M(\bm{0}_n) = \nabla^k U(\bm{0}_n) + O\left(\frac{1}{l}\right)
\end{equation*}
and the theorem is proved. 

\end{proof}

\section{Proof of Theorem 4}
\begin{proof}
It follows from \Cref{lemma:moment-generating-function-for-FMS} that the moment generating function of the projected distribution onto any unit vector $\bm{v}$ is
\begin{equation*}
\begin{split}
M(t|\bm{v}) &= \exp\left(\frac{1}{2}(1-\gamma)t^2\right) \left(\sum_{j=1}^m \alpha_j \exp\left(\frac{\gamma}{2l}(\bm{v}^T \bm{q}_j)^2 t^2\right)\right)^l \\
&= \left(\exp\left(\frac{1}{2l}(1-\gamma)t^2\right)\right)^l \\
&\;\;\;\;\; \left(\sum_{j=1}^m \alpha_j \exp\left(\frac{\gamma}{2l}(\bm{v}^T \bm{q}_j)^2 t^2\right)\right)^l \\
&= \left(\sum_{j=1}^m \alpha_j \exp\left(\frac{t^2}{2l}\left((1-\gamma)+\gamma(\bm{v}^T \bm{q}_j)^2 \right)\right)\right)^l \\
&= \left(\sum_{j=1}^m \alpha_j \exp\left(\frac{t^2}{2l} d_j \right)\right)^l \\
\end{split}
\end{equation*}
where we denote $(1-\gamma)+\gamma(\bm{v}^T \bm{q}_j)^2$ by $d_j$ to simplify the notations.
Expand the exponential term as

\begin{equation*}
\exp\left(\frac{t^2}{2l}d_j\right) = 1 + \frac{t^2}{2l}d_j + \frac{t^4}{8l^2}d_j^2 + O(t^6)
\end{equation*}

and we obtain
\begin{equation*}
\begin{split}
M(t|\bm{v}) &= \left(\sum_{j=1}^m \alpha_j \left( 1 + \frac{t^2}{2l}d_j + \frac{t^4}{8l^2}d_j^2 + O(t^6) \right)\right)^l \\
&= \left(1 + t^2\sum_{j=1}^m \alpha_j \frac{d_j}{2l} + t^4\sum_{j=1}^m \alpha_j \frac{d_j^2}{8l^2} + O(t^6)\right)^l \\
&= t^4 \left(\frac{l(l-1)}{2} \cdot \left(\sum_{j=1}^m \alpha_j \frac{d_j}{2l}\right)^2 + l\cdot \sum_{j=1}^m \alpha_j \frac{d_j^2}{8l^2}\right) \\
&\;\;\;\;\;\;\;\; + 1 + At^2 + O(t^6) \\
\end{split}
\end{equation*}
where $A$ is some constant independent of $t$ and the last equality is derived from the multinomial theorem.

The fourth order moment can thus be obtained as

\begin{equation*}
\begin{split}
& \frac{\partial^4}{\partial t^4}M(t|\bm{v}) |_{t=0} \\
&\;\;\;= 24 \left(\frac{l(l-1)}{2}  \left(\sum_{j=1}^m \alpha_j \frac{d_j}{2l}\right)^2 + l \sum_{j=1}^m \alpha_j \frac{d_j^2}{8l^2}\right) \\
&\;\;\;= 3\frac{l-1}{l} \left(\sum_{j=1}^m \alpha_j d_j\right)^2 + 3\frac{1}{l} \sum_{j=1}^m \alpha_j d_j^2
\end{split}
\end{equation*}

On the other hand, it follows from \textit{\Cref{eq:conditioned-covariance-matrix}} that the variance of the projected distribution is $\sum_{j=1}^m \alpha_j d_j$. So the excess kurtosis can be calculated as
\begin{equation*}
\frac{\frac{\partial^4}{\partial t^4}M(t|\bm{v}) |_{t=0}}{\left(\sum_{j=1}^m \alpha_j d_j\right)^2} - 3 =
\frac{3}{l} \left(\frac{\sum_{j=1}^m \alpha_j d_j^2}{\left(\sum_{j=1}^m \alpha_j d_j\right)^2} - 1\right)
\end{equation*}
Applying the Jensen's inequality $$\sum_{j=1}^m\alpha_j d_j^2 \ge \left(\sum_{j=1}^m\alpha_j d_j\right)^2$$ completes the proof.

\end{proof}

\section{Proof of Proposition 1}
\begin{proof}
Set $A_i = \omega_i I\{f(\bm{x}_{i:\lambda}^{(g-1)}) > f(\bm{x}_{i:\lambda}^{(g)})\}$. 
We show that the sequence $\{A_i\}$ has bounded fourth order moment such that the Lyapunov's condition holds and we can apply the central limit theorem to reach the conclusion.

Firstly, it is easy to see that $E[A_i] = \frac{1}{2}\omega_i$ and $Var[A_i]=\frac{1}{4}\omega_i^2$. Then, we check the Lyapunov's condition by investigating the growing rate of the fourth order moment compared with the second order moment
\begin{equation*} 
\begin{split}
T_\mu & = \frac{1}{(\sum_{i=1}^\mu Var[A_i])^2} \sum_{i=1}^\mu E[|A_i-E[A_i]|^4] \\
& = 2\frac{1}{(\sum_{i=1}^\mu \omega_i^2)^2} \sum_{i=1}^\mu \omega_i^4 \\
& = 2\frac{1}{(\sum_{i=1}^\mu (\omega_i')^2)^2} \sum_{i=1}^\mu (\omega_i')^4 \\
\end{split}
\end{equation*}

Then, define a function 
\begin{equation*}
v(x) = (\ln(\mu+0.5)-\ln(x))^2.
\end{equation*}
It is easy to see that $v(x)$ is convex on the range $[1,\mu]$. Then we have, from this convexity,
\begin{equation*}
\begin{split}
\sum_{i=1}^\mu (\omega_i')^2 &= \sum_{i=1}^\mu v(i) \ge \mu v\left(\frac{1+\mu}{2}\right) \\
& = \mu \left(\ln\left(\frac{2\mu+1}{\mu+1}\right)\right)^2
\end{split}
\end{equation*}

Therefore, we have
\begin{equation*}
T_\mu \le 2\frac{\sum_i^\mu (\omega_i')^4}{\mu^2\left(\ln(\frac{2\mu+1}{\mu+1})\right)^4} \le 2\frac{\mu (\omega_1')^4}{\mu^2\left(\ln(\frac{2\mu+1}{\mu+1})\right)^4}
\end{equation*}
and thus $\lim_ {\mu \rightarrow \infty} T_\mu = 0$. This indicates that the rate of growth of the fourth order moment is limited, i.e., the Lyapunov's condition holds. Finally, apply the central limit theorem and we get

\begin{equation*}
\frac{\sum_{i=1}^\mu \left(A_i-E[A_i]\right)}{\sqrt{\sum_{i=1}^\mu Var[A_i]}} \xrightarrow{\text{d}} \mathcal{N}(0,1)
\end{equation*}
which is equivalent to the conclusion.
\end{proof}

\section{Detailed numerical results associated with \Cref{tab:1000-dimensional-summary,tab:scability-ranks,tab:CEC2010}}

\Cref{tab:1000-dimensional-summary-detailed,tab:scability-summary,tab:CEC2010-summary} provides respectively the detailed numerical results associated with \Cref{tab:1000-dimensional-summary,tab:scability-ranks,tab:CEC2010} in the main manuscript.

\begin{table*}[htbp]
	\centering
	\scriptsize
	\caption{Median results on the 1000-dimensional basic test problems in terms of the number of function evaluations required to reach the accuracy $10^{-8}$. The best and the second best results for each test instance are shown with dark and light gray background, respectively.}
	\setlength{\tabcolsep}{6pt}
	\label{tab:1000-dimensional-summary-detailed}
	\begin{threeparttable}
\begin{tabular}{cccccccc}
\toprule
  & sep-CMA & CMA-ES & LM-MA & LM-CMA & Rm-ES & SDA-ES & MMES \\
\midrule
$\fElli$ & \cellcolor[rgb]{ .627,  .627,  .627} $4.22E+05$ $\circ$ & $1.87E+07$ $\bullet$ & $1.41E+07$ $\bullet$ & \cellcolor[rgb]{ .816,  .816,  .816} $1.07E+07$ $\circ$ & $1.70E+07$ $\bullet$ & $1.40E+07$ $\bullet$ & $1.24E+07$ \\
$\fRosen$ & $1.90E+07$ $\bullet$ & $1.99E+07$ $\bullet$ & $1.09E+07$ $\dagger$ & $1.53E+07$ $\bullet$ & $1.05E+07$ $\dagger$ & \cellcolor[rgb]{ .627,  .627,  .627} $7.63E+06$ $\circ$ & \cellcolor[rgb]{ .816,  .816,  .816} $1.01E+07$ \\
$\fDiscus$ & \cellcolor[rgb]{ .627,  .627,  .627} $1.66E+05$ $\circ$ & $7.08E+06$ $\bullet$ & $4.10E+06$ $\bullet$ & $3.08E+06$ $\bullet$ & $1.98E+06$ $\bullet$ & $2.42E+06$ $\bullet$ & \cellcolor[rgb]{ .816,  .816,  .816} $1.62E+06$ \\
$\fCigar$ & $2.07E+05$ $\bullet$ & $3.53E+05$ $\bullet$ & $1.76E+06$ $\bullet$ & \cellcolor[rgb]{ .816,  .816,  .816} $1.96E+05$ $\dagger$ & \cellcolor[rgb]{ .627,  .627,  .627} $1.81E+05$ $\circ$ & $2.09E+05$ $\bullet$ & $1.97E+05$ \\
$\fDiffPow$ & \cellcolor[rgb]{ .627,  .627,  .627} $2.42E+05$ $\circ$ & $1.80E+06$ $\bullet$ & $5.65E+05$ $\circ$ & \cellcolor[rgb]{ .816,  .816,  .816} $5.10E+05$ $\circ$ & $8.93E+05$ $\bullet$ & $7.21E+05$ $\bullet$ & $5.88E+05$ \\
$\fRotElli$ & N/A $\bullet$ & $1.86E+07$ $\bullet$ & $1.39E+07$ $\bullet$ & \cellcolor[rgb]{ .627,  .627,  .627} $1.14E+07$ $\circ$ & $1.71E+07$ $\bullet$ & $1.40E+07$ $\bullet$ & \cellcolor[rgb]{ .816,  .816,  .816} $1.24E+07$ \\
$\fRotRosen$ & N/A $\bullet$ & $3.56E+07$ $\bullet$ & \cellcolor[rgb]{ .816,  .816,  .816} $8.50E+06$ $\circ$ & $1.77E+07$ $\bullet$ & $1.25E+07$ $\dagger$ & \cellcolor[rgb]{ .627,  .627,  .627} $8.09E+06$ $\circ$ & $1.15E+07$ \\
$\fRotDiscus$ & N/A $\bullet$ & $6.99E+06$ $\bullet$ & $4.09E+06$ $\bullet$ & $2.56E+06$ $\bullet$ & \cellcolor[rgb]{ .816,  .816,  .816} $1.96E+06$ $\bullet$ & $2.42E+06$ $\bullet$ & \cellcolor[rgb]{ .627,  .627,  .627} $1.62E+06$ \\
$\fRotCigar$ & N/A $\bullet$ & $3.53E+05$ $\bullet$ & $1.74E+06$ $\bullet$ & \cellcolor[rgb]{ .816,  .816,  .816} $1.97E+05$ $\dagger$ & \cellcolor[rgb]{ .627,  .627,  .627} $1.81E+05$ $\circ$ & $2.09E+05$ $\bullet$ & $1.98E+05$ \\
$\fRotDiffPow$ & N/A $\bullet$ & $1.83E+06$ $\bullet$ & \cellcolor[rgb]{ .627,  .627,  .627} $5.64E+05$ $\circ$ & \cellcolor[rgb]{ .816,  .816,  .816} $5.83E+05$ $\circ$ & $8.90E+05$ $\bullet$ & $7.16E+05$ $\bullet$ & $5.92E+05$ \\
\midrule
$\bullet$ / $\circ$ / $\dagger$ & 7 / 3 / 0 & 10 / 0 / 0 & 6 / 3 / 1 & 4 / 4 / 2 & 6 / 2 / 2 & 8 / 2 / 0 &  \\
\bottomrule
\end{tabular}%

		\begin{tablenotes}
			\item[1] ``$\bullet$'' indicates that MMES significantly outperforms the peer algorithm at a 0.05 significance level by the Wilcoxon rank sum test, whereas ``$\circ$'' indicates the opposite. If no significant difference is detected, it will be marked by the symbol ``$\dagger$''. They have the same meanings in other tables.
			\item[2] The median result is marked by ``N/A'' if the algorithm fails to reach the target accuracy in all independent runs.
		\end{tablenotes}
	\end{threeparttable}
\end{table*}

\begin{table*}[htbp]
	\centering
	\scriptsize
	\caption{Median results on the 2500-, 5000-, 7500-, and 10000-dimensional basic test problems in terms of the number of function evaluations required to reach the accuracy $10^{-8}$. The best and the second best results for each test instance are shown with dark and light gray background, respectively.}
	\setlength{\tabcolsep}{8pt}
	\label{tab:scability-summary}
\begin{tabular}{ccccccc}
\toprule
  & $n$ & LM-MA & LM-CMA & Rm-ES & SDA-ES & MMES \\
\midrule
$\fElli$ & 2500 & $3.75E+07$ $\bullet$ & \cellcolor[rgb]{ .627,  .627,  .627} $2.70E+07$ $\circ$ & $4.44E+07$ $\bullet$ & $3.05E+07$ $\bullet$ & \cellcolor[rgb]{ .816,  .816,  .816} $3.02E+07$  \\
  & 5000 & $7.76E+07$ $\bullet$ & \cellcolor[rgb]{ .627,  .627,  .627} $5.55E+07$ $\circ$ & $9.34E+07$ $\bullet$ & $6.05E+07$ $\bullet$ & \cellcolor[rgb]{ .816,  .816,  .816} $5.94E+07$  \\
  & 7500 & $1.17E+08$ $\bullet$ & \cellcolor[rgb]{ .816,  .816,  .816} $9.25E+07$ $\bullet$ & $1.45E+08$ $\bullet$ & $9.45E+07$ $\bullet$ & \cellcolor[rgb]{ .627,  .627,  .627} $8.80E+07$  \\
  & 10000 & $1.60E+08$ $\bullet$ & $1.46E+08$ $\bullet$ & $2.00E+08$ $\bullet$ & \cellcolor[rgb]{ .816,  .816,  .816} $1.30E+08$ $\bullet$ & \cellcolor[rgb]{ .627,  .627,  .627} $1.17E+08$  \\
\midrule
$\fRosen$ & 2500 & $3.87E+07$ $\dagger$ & $6.49E+07$ $\bullet$ & \cellcolor[rgb]{ .816,  .816,  .816} $3.32E+07$ $\dagger$ & \cellcolor[rgb]{ .627,  .627,  .627} $3.16E+07$ $\dagger$ & $3.77E+07$  \\
  & 5000 & $9.52E+07$ $\dagger$ & $2.02E+08$ $\bullet$ & \cellcolor[rgb]{ .627,  .627,  .627} $8.10E+07$ $\dagger$ & \cellcolor[rgb]{ .816,  .816,  .816} $8.35E+07$ $\dagger$ & $9.17E+07$  \\
  & 7500 & \cellcolor[rgb]{ .816,  .816,  .816} $1.35E+08$ $\dagger$ & $2.98E+08$ $\bullet$ & \cellcolor[rgb]{ .627,  .627,  .627} $1.22E+08$ $\dagger$ & $1.45E+08$ $\dagger$ & $1.39E+08$  \\
  & 10000 & \cellcolor[rgb]{ .816,  .816,  .816} $1.89E+08$ $\dagger$ & $4.01E+08$ $\bullet$ & $2.10E+08$ $\dagger$ & \cellcolor[rgb]{ .627,  .627,  .627} $1.63E+08$ $\circ$ & $1.94E+08$  \\
\midrule
$\fDiscus$ & 2500 & $6.51E+06$ $\bullet$ & $6.75E+06$ $\bullet$ & $6.17E+06$ $\bullet$ & \cellcolor[rgb]{ .816,  .816,  .816} $4.46E+06$ $\bullet$ & \cellcolor[rgb]{ .627,  .627,  .627} $2.88E+06$  \\
  & 5000 & $9.45E+06$ $\bullet$ & $3.31E+07$ $\bullet$ & $1.37E+07$ $\bullet$ & \cellcolor[rgb]{ .816,  .816,  .816} $7.08E+06$ $\bullet$ & \cellcolor[rgb]{ .627,  .627,  .627} $4.20E+06$  \\
  & 7500 & $1.19E+07$ $\bullet$ & $5.84E+07$ $\bullet$ & $2.05E+07$ $\bullet$ & \cellcolor[rgb]{ .816,  .816,  .816} $9.20E+06$ $\bullet$ & \cellcolor[rgb]{ .627,  .627,  .627} $5.23E+06$  \\
  & 10000 & $1.37E+07$ $\bullet$ & $7.95E+07$ $\bullet$ & $2.70E+07$ $\bullet$ & \cellcolor[rgb]{ .816,  .816,  .816} $1.11E+07$ $\bullet$ & \cellcolor[rgb]{ .627,  .627,  .627} $6.57E+06$  \\
\midrule
$\fCigar$ & 2500 & $4.07E+06$ $\bullet$ & $5.38E+05$ $\bullet$ & \cellcolor[rgb]{ .627,  .627,  .627} $4.30E+05$ $\circ$ & $5.05E+05$ $\bullet$ & \cellcolor[rgb]{ .816,  .816,  .816} $4.67E+05$  \\
  & 5000 & $7.92E+06$ $\bullet$ & $1.03E+06$ $\bullet$ & \cellcolor[rgb]{ .627,  .627,  .627} $8.69E+05$ $\circ$ & $9.81E+05$ $\bullet$ & \cellcolor[rgb]{ .816,  .816,  .816} $9.10E+05$  \\
  & 7500 & $1.17E+07$ $\bullet$ & $1.70E+06$ $\bullet$ & \cellcolor[rgb]{ .627,  .627,  .627} $1.31E+06$ $\circ$ & $1.47E+06$ $\bullet$ & \cellcolor[rgb]{ .816,  .816,  .816} $1.33E+06$  \\
  & 10000 & $1.56E+07$ $\bullet$ & $2.50E+06$ $\bullet$ & \cellcolor[rgb]{ .627,  .627,  .627} $1.70E+06$ $\circ$ & $1.96E+06$ $\bullet$ & \cellcolor[rgb]{ .816,  .816,  .816} $1.76E+06$  \\
\midrule
$\fDiffPow$ & 2500 & $1.86E+06$ $\bullet$ & \cellcolor[rgb]{ .627,  .627,  .627} $1.59E+06$ $\circ$ & $2.86E+06$ $\bullet$ & $2.44E+06$ $\bullet$ & \cellcolor[rgb]{ .816,  .816,  .816} $1.74E+06$  \\
  & 5000 & $4.69E+06$ $\bullet$ & \cellcolor[rgb]{ .627,  .627,  .627} $3.91E+06$ $\circ$ & $6.99E+06$ $\bullet$ & $6.20E+06$ $\bullet$ & \cellcolor[rgb]{ .816,  .816,  .816} $3.98E+06$  \\
  & 7500 & $7.91E+06$ $\bullet$ & \cellcolor[rgb]{ .627,  .627,  .627} $6.36E+06$ $\circ$ & $1.18E+07$ $\bullet$ & $1.06E+07$ $\bullet$ & \cellcolor[rgb]{ .816,  .816,  .816} $6.43E+06$  \\
  & 10000 & $1.23E+07$ $\bullet$ & \cellcolor[rgb]{ .627,  .627,  .627} $9.07E+06$ $\circ$ & $1.75E+07$ $\bullet$ & $1.52E+07$ $\bullet$ & \cellcolor[rgb]{ .816,  .816,  .816} $9.16E+06$  \\
\midrule
$\bullet$ / $\circ$ / $\dagger$ &   & 16 / 0 / 4 & 14 / 6 / 0 & 12 / 4 / 4 & 16 / 1 / 3 &  \\
\bottomrule
\end{tabular}%

\end{table*}

\begin{table*}[htbp]
	\centering
	\scriptsize
	\caption{Median of the objective values obtained on the CEC'2010 test problems. The best and the second best results for each test instance are shown with dark and light gray background, respectively.}
	\setlength{\tabcolsep}{12pt}
	\label{tab:CEC2010-summary}
\begin{tabular}{ccccccc}
\toprule
  & DECC-G & MA-SW & MOS & CCPSO2 & DECC-DG & MMES \\
\midrule
$f_{1}$ & $3.65E-07$ $\circ$ & \cellcolor[rgb]{ .816,  .816,  .816} $1.50E-14$ $\circ$ & \cellcolor[rgb]{ .627,  .627,  .627} $0.00E+00$ $\circ$ & $7.80E-01$ $\circ$ & $1.42E+02$ $\circ$ & $6.38E+03$  \\
$f_{2}$ & $1.32E+03$ $\bullet$ & $7.90E+02$ $\bullet$ & \cellcolor[rgb]{ .816,  .816,  .816} $1.95E+02$ $\circ$ & \cellcolor[rgb]{ .627,  .627,  .627} $4.25E+00$ $\circ$ & $4.46E+03$ $\bullet$ & $3.99E+02$  \\
$f_{3}$ & $1.13E+00$ $\bullet$ & \cellcolor[rgb]{ .816,  .816,  .816} $6.11E-13$ $\bullet$ & $1.29E+00$ $\bullet$ & $4.16E-03$ $\bullet$ & $1.66E+01$ $\bullet$ & \cellcolor[rgb]{ .627,  .627,  .627} $0.00E+00$  \\
$f_{4}$ & $2.56E+13$ $\bullet$ & $3.54E+11$ $\bullet$ & \cellcolor[rgb]{ .816,  .816,  .816} $1.88E+10$ $\bullet$ & $1.45E+12$ $\bullet$ & $5.08E+12$ $\bullet$ & \cellcolor[rgb]{ .627,  .627,  .627} $3.72E+09$  \\
$f_{5}$ & $2.49E+08$ $\bullet$ & $2.31E+08$ $\bullet$ & $6.86E+08$ $\bullet$ & $3.76E+08$ $\bullet$ & \cellcolor[rgb]{ .816,  .816,  .816} $1.52E+08$ $\bullet$ & \cellcolor[rgb]{ .627,  .627,  .627} $1.10E+07$  \\
$f_{6}$ & $4.85E+06$ $\bullet$ & \cellcolor[rgb]{ .627,  .627,  .627} $1.60E+00$ $\circ$ & $1.98E+07$ $\bullet$ & $1.97E+07$ $\bullet$ & \cellcolor[rgb]{ .816,  .816,  .816} $1.64E+01$ $\circ$ & $2.13E+01$  \\
$f_{7}$ & $7.19E+08$ $\bullet$ & \cellcolor[rgb]{ .816,  .816,  .816} $9.04E+01$ $\circ$ & \cellcolor[rgb]{ .627,  .627,  .627} $0.00E+00$ $\circ$ & $2.67E+06$ $\bullet$ & $9.20E+03$ $\circ$ & $7.71E+05$  \\
$f_{8}$ & $8.82E+07$ $\bullet$ & $3.43E+06$ $\bullet$ & \cellcolor[rgb]{ .627,  .627,  .627} $2.74E-01$ $\circ$ & $2.00E+07$ $\bullet$ & $1.62E+07$ $\bullet$ & \cellcolor[rgb]{ .816,  .816,  .816} $9.75E+05$  \\
$f_{9}$ & $4.32E+08$ $\bullet$ & $1.40E+07$ $\bullet$ & \cellcolor[rgb]{ .816,  .816,  .816} $8.83E+06$ $\bullet$ & $1.14E+08$ $\bullet$ & $5.52E+07$ $\bullet$ & \cellcolor[rgb]{ .627,  .627,  .627} $7.10E+03$  \\
$f_{10}$ & $1.01E+04$ $\bullet$ & \cellcolor[rgb]{ .816,  .816,  .816} $2.07E+03$ $\bullet$ & $7.83E+03$ $\bullet$ & $5.14E+03$ $\bullet$ & $4.47E+03$ $\bullet$ & \cellcolor[rgb]{ .627,  .627,  .627} $4.27E+02$  \\
$f_{11}$ & $2.54E+01$ $\dagger$ & $3.75E+01$ $\bullet$ & $1.99E+02$ $\bullet$ & $1.98E+02$ $\bullet$ & \cellcolor[rgb]{ .627,  .627,  .627} $1.02E+01$ $\circ$ & \cellcolor[rgb]{ .816,  .816,  .816} $2.02E+01$  \\
$f_{12}$ & $9.92E+04$ $\bullet$ & $3.50E-06$ $\bullet$ & \cellcolor[rgb]{ .627,  .627,  .627} $0.00E+00$ $\dagger$ & $2.78E+04$ $\bullet$ & $2.58E+03$ $\bullet$ & \cellcolor[rgb]{ .627,  .627,  .627} $0.00E+00$  \\
$f_{13}$ & $3.56E+03$ $\bullet$ & \cellcolor[rgb]{ .816,  .816,  .816} $1.07E+03$ $\bullet$ & $1.18E+03$ $\bullet$ & $1.36E+03$ $\bullet$ & $5.06E+03$ $\bullet$ & \cellcolor[rgb]{ .627,  .627,  .627} $1.60E+01$  \\
$f_{14}$ & $9.98E+08$ $\bullet$ & $3.09E+07$ $\bullet$ & \cellcolor[rgb]{ .816,  .816,  .816} $1.85E+07$ $\bullet$ & $3.42E+08$ $\bullet$ & $3.46E+08$ $\bullet$ & \cellcolor[rgb]{ .627,  .627,  .627} $8.99E+03$  \\
$f_{15}$ & $1.18E+04$ $\bullet$ & \cellcolor[rgb]{ .816,  .816,  .816} $2.72E+03$ $\bullet$ & $1.54E+04$ $\bullet$ & $1.04E+04$ $\bullet$ & $5.86E+03$ $\bullet$ & \cellcolor[rgb]{ .627,  .627,  .627} $4.19E+02$  \\
$f_{16}$ & $7.32E+01$ $\bullet$ & $9.44E+01$ $\bullet$ & $3.97E+02$ $\bullet$ & $3.97E+02$ $\bullet$ & \cellcolor[rgb]{ .627,  .627,  .627} $7.50E-13$ $\circ$ & \cellcolor[rgb]{ .816,  .816,  .816} $3.83E+01$  \\
$f_{17}$ & $3.09E+05$ $\bullet$ & $1.26E+00$ $\bullet$ & \cellcolor[rgb]{ .816,  .816,  .816} $4.83E-05$ $\bullet$ & $8.99E+04$ $\bullet$ & $4.02E+04$ $\bullet$ & \cellcolor[rgb]{ .627,  .627,  .627} $0.00E+00$  \\
$f_{18}$ & $3.03E+04$ $\bullet$ & \cellcolor[rgb]{ .816,  .816,  .816} $1.19E+03$ $\bullet$ & $3.55E+03$ $\bullet$ & $3.10E+03$ $\bullet$ & $1.47E+10$ $\bullet$ & \cellcolor[rgb]{ .627,  .627,  .627} $6.10E+01$  \\
$f_{19}$ & $1.13E+06$ $\bullet$ & $2.85E+05$ $\bullet$ & \cellcolor[rgb]{ .816,  .816,  .816} $3.40E+04$ $\bullet$ & $1.51E+06$ $\bullet$ & $1.75E+06$ $\bullet$ & \cellcolor[rgb]{ .627,  .627,  .627} $1.60E-06$  \\
$f_{20}$ & $4.22E+03$ $\bullet$ & $1.06E+03$ $\bullet$ & \cellcolor[rgb]{ .816,  .816,  .816} $7.26E+02$ $\bullet$ & $2.10E+03$ $\bullet$ & $6.53E+10$ $\bullet$ & \cellcolor[rgb]{ .627,  .627,  .627} $6.55E+02$  \\
\midrule
$\bullet$ / $\circ$ / $\dagger$ & 18 / 1 / 1 & 17 / 3 / 0 & 15 / 4 / 1 & 18 / 2 / 0 & 15 / 5 / 0 &  \\
\bottomrule
\end{tabular}%

\end{table*}

\section{Additional experiments for verifying the impact of mixing strength}
While in the main manuscript we have investigated the performance of MMES with mixing strength set to different small constants (i.e., $l \in \{2, 4, 8, 16, 32\}$), here we consider another setting, $l=m$. That is, totally $m$ vectors are randomly selected from $\bm{q}_1,\cdots,\bm{q}_m$ for constructing the mixture distribution $\mathcal{P}_m$. Please note that since the selection is with replacement this setting does not guarantee that every $\bm{q}$ vector is chosen exactly once. It neither produces the target distribution $\mathcal{P}_a$, as the latter can only be achieved in the case of $l \rightarrow \infty$ (see \Cref{theorem:FMS-convergence-to-gaussian}). The aim of this section is to verify the impact of mixing strength when set to a relatively large value and to provide more evidences to our statements about its practical setting.

\subsection{Impact on the algorithm efficiency}
\begin{figure}[htbp]
	\centering
	\includegraphics[width=0.4\textwidth]{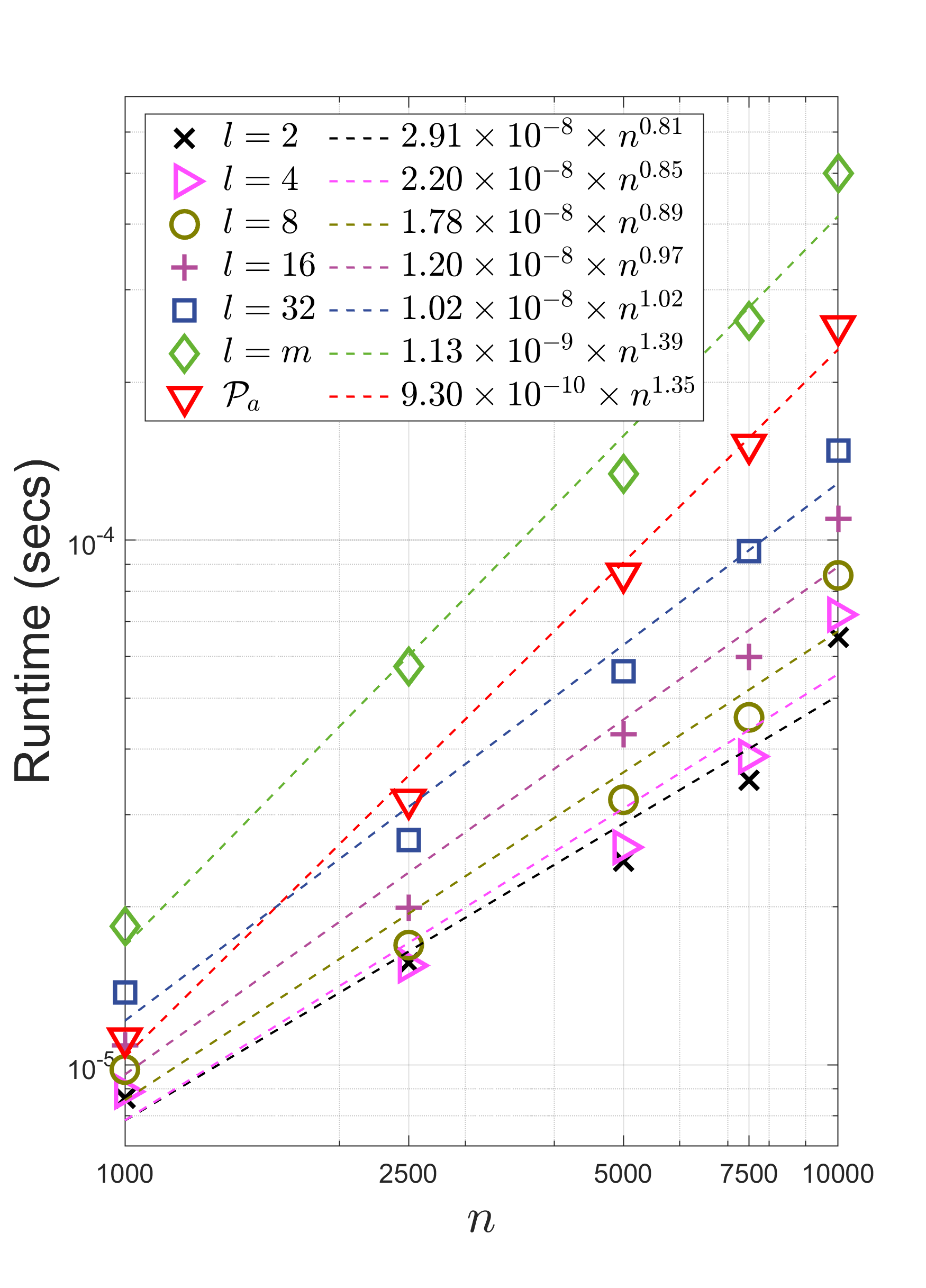} 
	\caption{Runtime results measured for different mixing strengths. The instance of MMES where the mixture distribution is directly replaced with the target Gaussian distribution is also considered and is denoted by $\mathcal{P}_a$. The dashed lines are obtained by the log-log linear regression where the order of $n$ estimates the asymptotic order of runtime growth.}
	\label{fig:running-time-different-l}
\end{figure}

We firstly show the runtime of MMES versus different $l$ values, where runtime is defined and measured in the same way as in \Cref{ss:runtime}. To reflect the asymptotic performance, we plot for each $l$ a log-log regression line by fitting the experimental results.  \Cref{fig:running-time-different-l} shows that the runtime of MMES increases consistently as $l$ increases. For $l\in \{2,4,8,16,32\}$ the runtime scales approximately linearly because the mixing strengths are constant. Choosing $l=m$ leads to an obvious jump in the order of the runtime (i.e., $1.39$ as shown by the regression line). It is because the parameter $m$ is set to $O(\sqrt{n})$, so the runtime increases super-linearly with the increasing $n$. In general, these observations coincide with the $O(ln)$ time complexity of MMES. We also consider another setting where the solutions are directly drawn from the target distribution, denoted by $\mathcal{P}_a$ in the plot. This setting leads to a super-linearly increasing runtime but is significantly faster than $l=m$, probably because the sampling can be performed within a simple matrix-vector multiplication. However, it requires nearly triple the time compared to the setting $l=4$, demonstrating the time efficiency improvement of the proposed mixture sampling method.

\begin{table*}[htbp]
	\centering
	\scriptsize
	\caption{Runtime results obtained on $\fElli$, with the execution time of the function evaluations excluded. Better results are shown with darker gray background.}
	\renewcommand\arraystretch{1.4}
	\setlength{\tabcolsep}{5.5pt}
	\label{tab:runtime-summary}
\begin{tabular}{cccccccccccccc}
\toprule
\multirow{2}[0]{*}{$n$} & \multirow{2}[0]{*}{sep-CMA} & \multirow{2}[0]{*}{CMA-ES} & \multirow{2}[0]{*}{LM-MA} & \multirow{2}[0]{*}{LM-CMA} & \multirow{2}[0]{*}{Rm-ES} & \multirow{2}[0]{*}{SDA-ES} & \multicolumn{7}{c}{MMES} \\
\cmidrule{8-14}
      &       &       &       &       &       &       & $l=2$ & $l=4$ & $l=8$ & $l=16$ & $l=32$ & $l=m$ & $\mathcal{P}_a$ \\
\midrule
1000  & \cellcolor[rgb]{ .627,  .627,  .627}8.55E-06 & \cellcolor[rgb]{ .98,  .98,  .98}6.38E-04 & \cellcolor[rgb]{ .953,  .953,  .953}2.48E-05 & \cellcolor[rgb]{ .922,  .922,  .922}2.26E-05 & \cellcolor[rgb]{ .686,  .686,  .686}8.87E-06 & \cellcolor[rgb]{ .776,  .776,  .776}1.01E-05 & \cellcolor[rgb]{ .659,  .659,  .659}8.63E-06 & \cellcolor[rgb]{ .718,  .718,  .718}8.89E-06 & \cellcolor[rgb]{ .745,  .745,  .745}9.80E-06 & \cellcolor[rgb]{ .804,  .804,  .804}1.09E-05 & \cellcolor[rgb]{ .863,  .863,  .863}1.37E-05 & \cellcolor[rgb]{ .894,  .894,  .894}1.84E-05 & \cellcolor[rgb]{ .835,  .835,  .835}1.12E-05 \\
2500  & \cellcolor[rgb]{ .745,  .745,  .745}1.96E-05 & \cellcolor[rgb]{ .98,  .98,  .98}4.92E-03 & \cellcolor[rgb]{ .922,  .922,  .922}5.04E-05 & \cellcolor[rgb]{ .894,  .894,  .894}4.90E-05 & \cellcolor[rgb]{ .718,  .718,  .718}1.90E-05 & \cellcolor[rgb]{ .804,  .804,  .804}2.01E-05 & \cellcolor[rgb]{ .659,  .659,  .659}1.57E-05 & \cellcolor[rgb]{ .627,  .627,  .627}1.55E-05 & \cellcolor[rgb]{ .686,  .686,  .686}1.69E-05 & \cellcolor[rgb]{ .776,  .776,  .776}1.99E-05 & \cellcolor[rgb]{ .835,  .835,  .835}2.68E-05 & \cellcolor[rgb]{ .953,  .953,  .953}5.74E-05 & \cellcolor[rgb]{ .863,  .863,  .863}3.20E-05 \\
5000  & \cellcolor[rgb]{ .718,  .718,  .718}3.70E-05 & \cellcolor[rgb]{ .98,  .98,  .98}2.39E-02 & \cellcolor[rgb]{ .894,  .894,  .894}9.98E-05 & \cellcolor[rgb]{ .922,  .922,  .922}1.15E-04 & \cellcolor[rgb]{ .776,  .776,  .776}4.33E-05 & \cellcolor[rgb]{ .804,  .804,  .804}4.60E-05 & \cellcolor[rgb]{ .627,  .627,  .627}2.45E-05 & \cellcolor[rgb]{ .659,  .659,  .659}2.60E-05 & \cellcolor[rgb]{ .686,  .686,  .686}3.20E-05 & \cellcolor[rgb]{ .745,  .745,  .745}4.27E-05 & \cellcolor[rgb]{ .835,  .835,  .835}5.62E-05 & \cellcolor[rgb]{ .953,  .953,  .953}1.34E-04 & \cellcolor[rgb]{ .863,  .863,  .863}8.64E-05 \\
7500  & \cellcolor[rgb]{ .718,  .718,  .718}5.44E-05 & \cellcolor[rgb]{ .98,  .98,  .98}5.32E-02 & \cellcolor[rgb]{ .894,  .894,  .894}1.53E-04 & \cellcolor[rgb]{ .922,  .922,  .922}1.56E-04 & \cellcolor[rgb]{ .804,  .804,  .804}6.22E-05 & \cellcolor[rgb]{ .745,  .745,  .745}5.83E-05 & \cellcolor[rgb]{ .627,  .627,  .627}3.48E-05 & \cellcolor[rgb]{ .659,  .659,  .659}3.87E-05 & \cellcolor[rgb]{ .686,  .686,  .686}4.60E-05 & \cellcolor[rgb]{ .776,  .776,  .776}5.99E-05 & \cellcolor[rgb]{ .835,  .835,  .835}9.54E-05 & \cellcolor[rgb]{ .953,  .953,  .953}2.62E-04 & \cellcolor[rgb]{ .863,  .863,  .863}1.52E-04 \\
10000 & \cellcolor[rgb]{ .718,  .718,  .718}8.96E-05 & \cellcolor[rgb]{ .98,  .98,  .98}6.12E-02 & \cellcolor[rgb]{ .922,  .922,  .922}4.62E-04 & \cellcolor[rgb]{ .863,  .863,  .863}2.47E-04 & \cellcolor[rgb]{ .745,  .745,  .745}9.55E-05 & \cellcolor[rgb]{ .776,  .776,  .776}1.08E-04 & \cellcolor[rgb]{ .627,  .627,  .627}6.53E-05 & \cellcolor[rgb]{ .659,  .659,  .659}7.21E-05 & \cellcolor[rgb]{ .686,  .686,  .686}8.57E-05 & \cellcolor[rgb]{ .804,  .804,  .804}1.10E-04 & \cellcolor[rgb]{ .835,  .835,  .835}1.48E-04 & \cellcolor[rgb]{ .953,  .953,  .953}5.01E-04 & \cellcolor[rgb]{ .894,  .894,  .894}2.56E-04 \\
\bottomrule
\end{tabular}%

\end{table*}

\subsection{Impact on the scalability}
We then assess the impact of mixing strength on the scalability performance of MMES. The experiment adopts the same setting as in \Cref{ss:sensitiveness-to-mixing-strength}. However, setting $l$ to $m$ leads to a super-linearly increase of the execution time and so we have to modify the stopping criteria such that MMES can stop in a reasonable time budget. Concretely, the target solution accuracy $\epsilon_{acc}$ for terminating MMES is set to $10^5, 1$, and $10^{-5}$ for $\fElli,\fDiscus$, and $\fDiffPow$, respectively. The $\fCigar$ function is relatively simple, so we use the same setting $\epsilon_{acc} = 10^{-8}$ as in \Cref{ss:sensitiveness-to-mixing-strength}.

\begin{figure}[tb]
	\centering
	\subfloat[$\fDiscus, \epsilon_{acc} = 1$]{\includegraphics[width=0.24\textwidth]{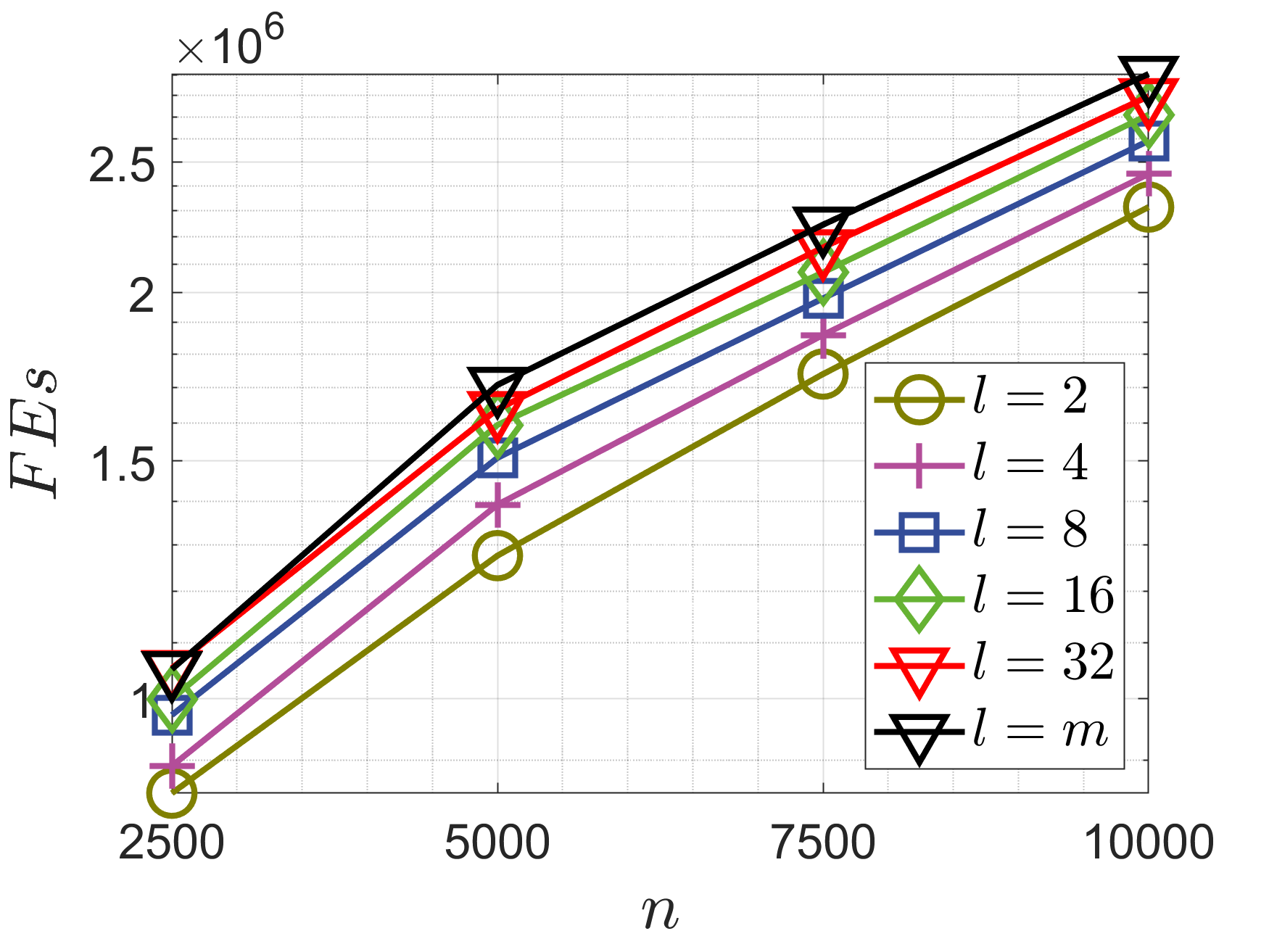}}
	\subfloat[$\fElli, \epsilon_{acc} = 10^5$]{\includegraphics[width=0.24\textwidth]{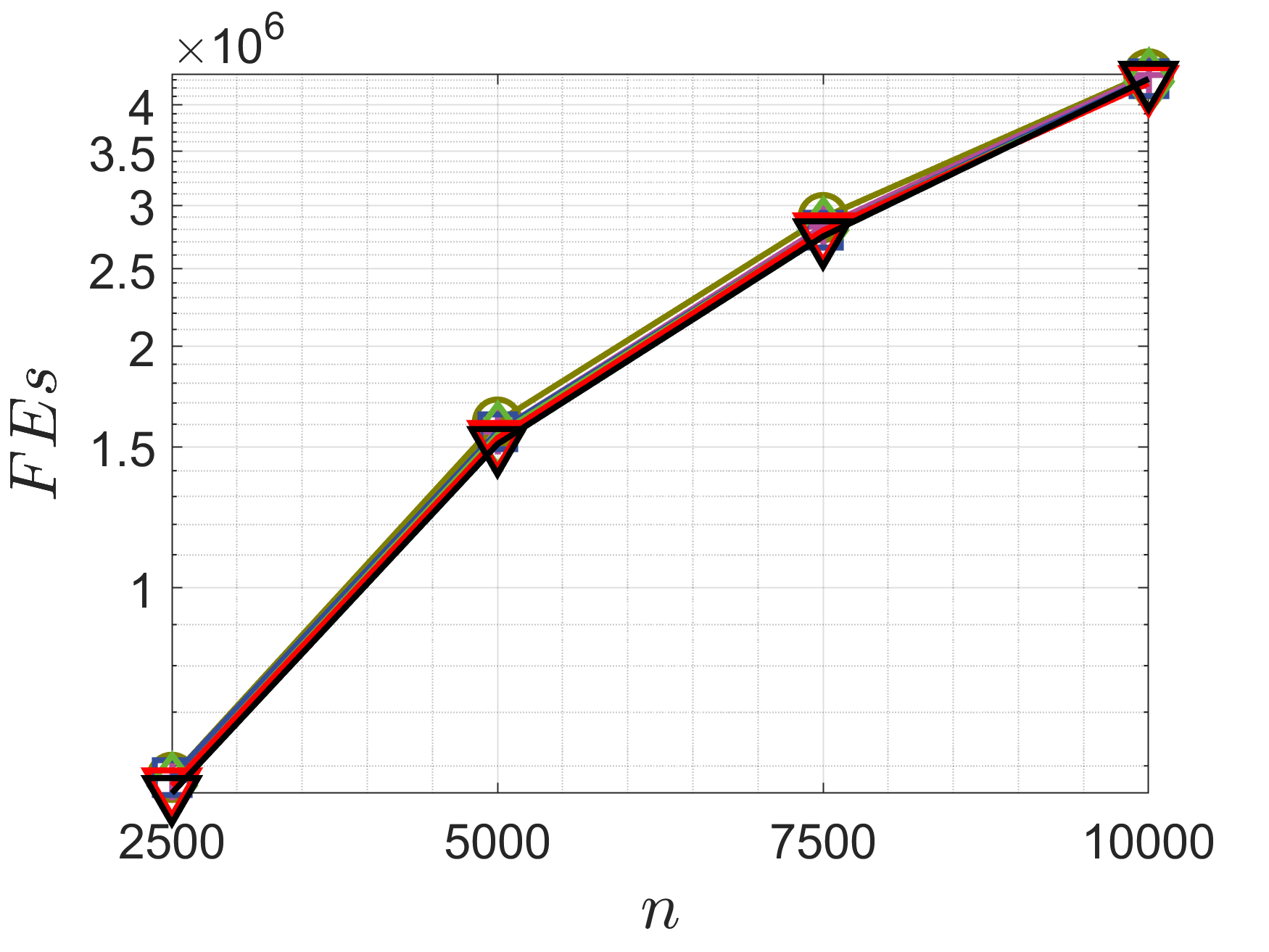}}
	\hfil
	\subfloat[$\fCigar, \epsilon_{acc} = 10^{-8}$]{\includegraphics[width=0.24\textwidth]{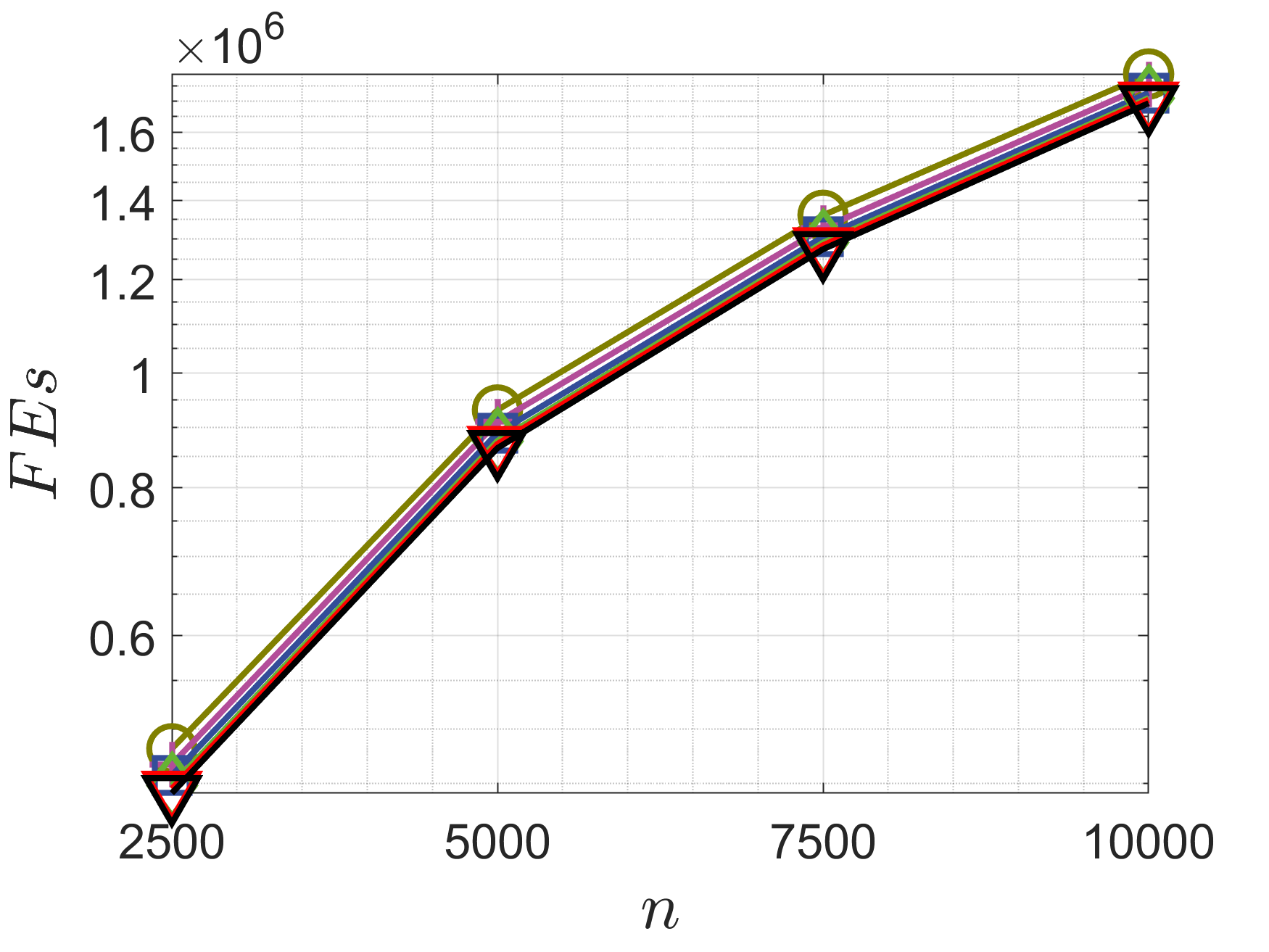}}
	\subfloat[$\fDiffPow, \epsilon_{acc} = 10^{-5}$]{\includegraphics[width=0.24\textwidth]{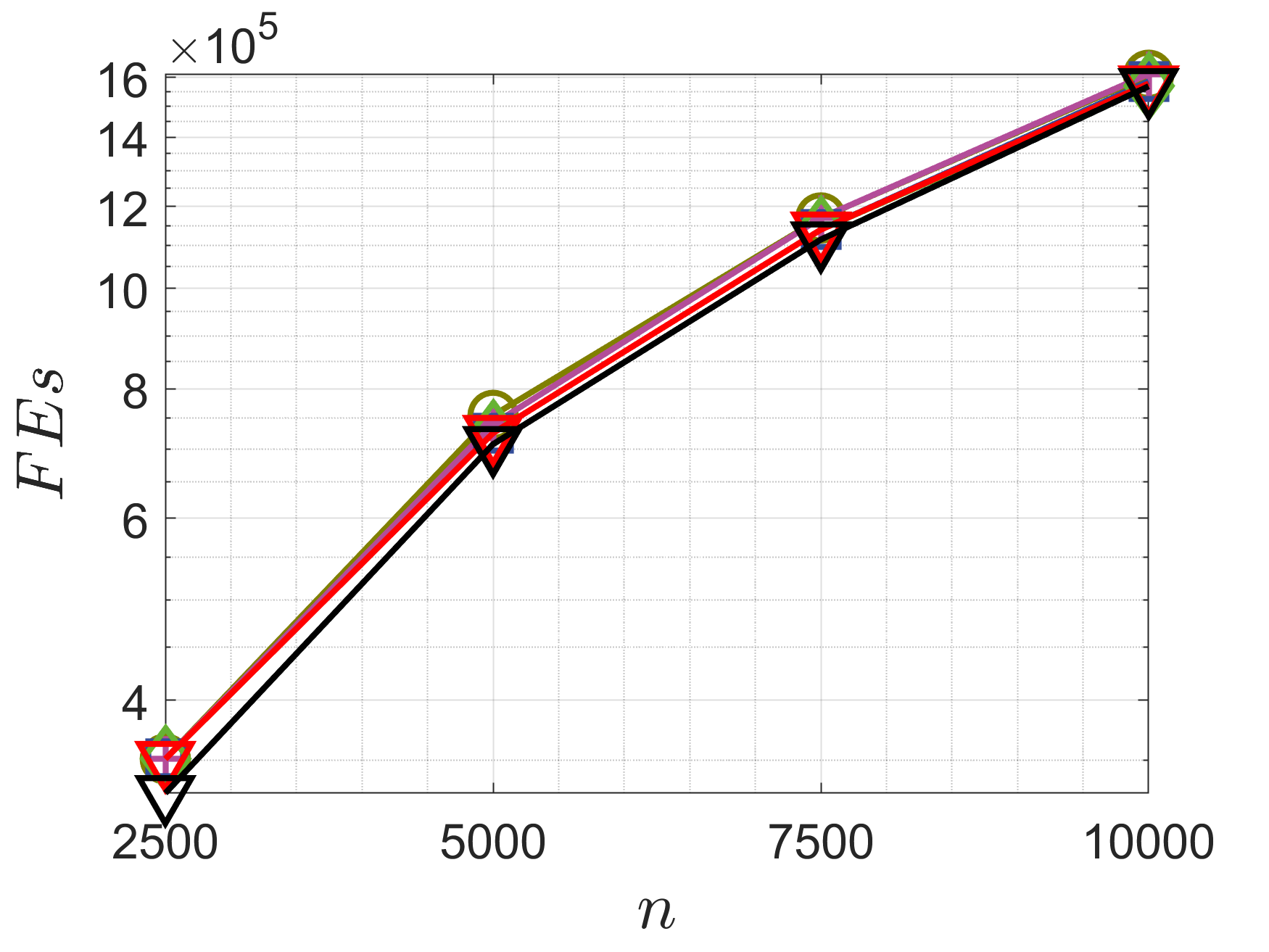}}
	\caption{Sensitiveness of MMES to different settings of $l$. The curves present the number of function evaluations required to reach the accuracy $\epsilon_{acc}$.}
	\label{fig:additional-parameter-test-scalability}
\end{figure}

\Cref{fig:additional-parameter-test-scalability} shows how mixing strength influences the effectiveness of MMES in decreasing the objective function. The presented results measure the number of function evaluations required to achieve the target accuracy. On $\fElli$ and $\fCigar$, the performance improves as $l$ increases; but these improvements are quite small. For example, in the 10000-dimensional case for both $\fElli$ and $\fCigar$, increasing $l$ from 2 to $m$ saves no more than 5\% function evaluations. Using $l=m$ leads to a moderate improvement on 2500-dimensional $\fDiscus$, but this improvement becomes insignificant for higher dimensions. On $\fDiscus$, contrarily, the increase of $l$ causes an obvious performance degradation; the setting $l=m$ leads to nearly 25\% more function evaluations than $l=2$ does in the 10000-dimensional instance. Linking these observations and those in \Cref{fig:running-time-different-l}, we conclude that a small constant value for the mixing strength is reasonable and can serve as a good trade-off between performance and efficiency.

\subsection{Impact on the rotational invariance}
The proposed FMS method is obviously invariant against rotations since it relies only on basic linear operations. Therefore, any rotational transformations applied to the decision space will not influence the performance of MMES, if the initial distribution can be transformed correspondingly. In \Cref{ss:rotational-invariance} we have demonstrated that, with $l=4$, this statement holds in practice and the non-invariant initialization has little influence on the algorithm performance. One may, however, argue that with a larger $l$ MMES may require more computational effect to offset the possible influence caused by the non-invariant initialization. This accords with the intuition that with a large $l$ the search distribution of MMES in the early stage may behave like a Gaussian which is non-invariantly initialized. Below we will show that this is not the case and MMES still possesses the rotational invariance even when $l$ is large.

\begin{figure}[tb]
	\centering
	\subfloat[Ellipsoid]{\includegraphics[width=0.24\textwidth]{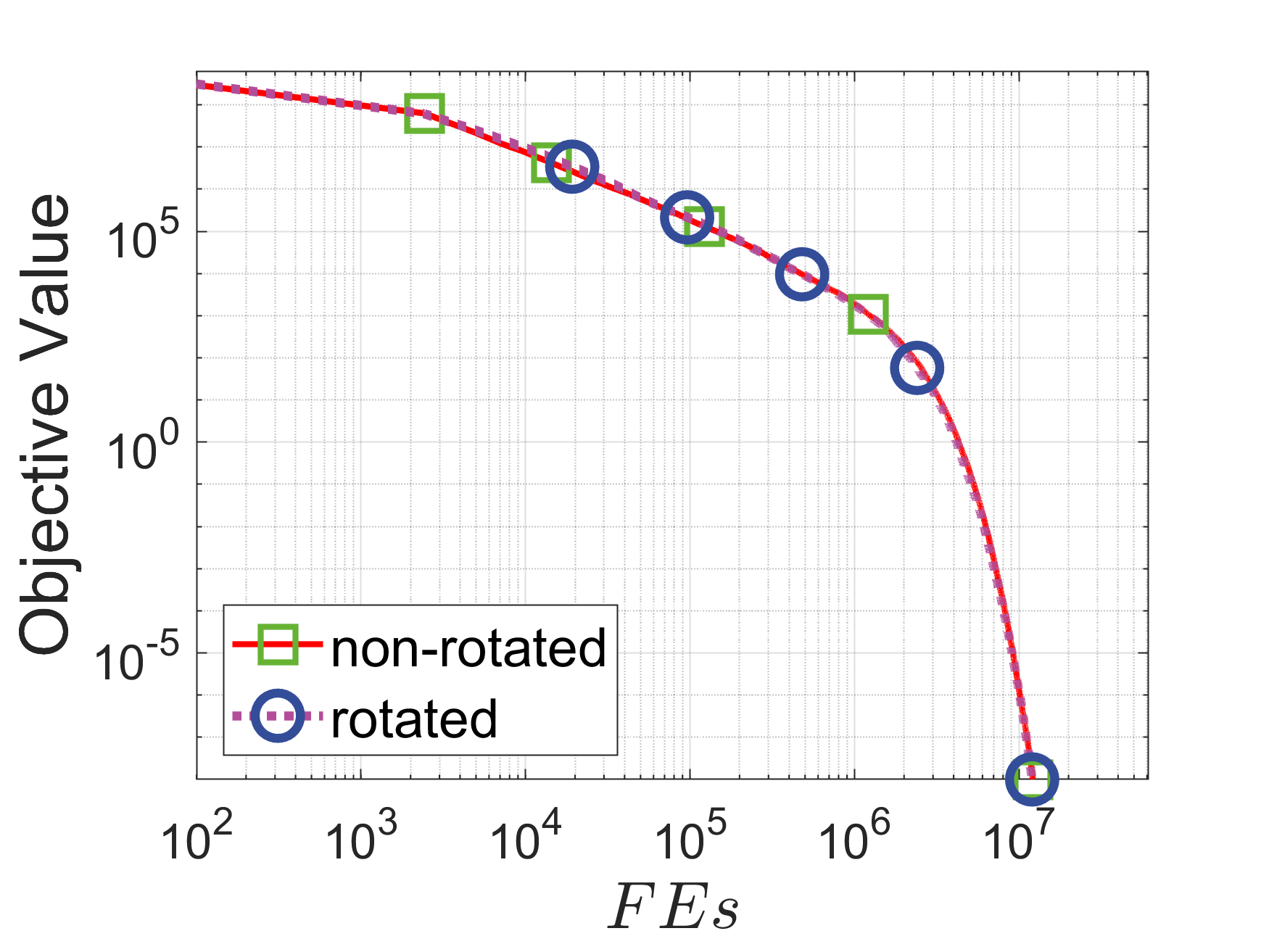}} 
	\hfil
	\subfloat[Rosenbrock]{\includegraphics[width=0.24\textwidth]{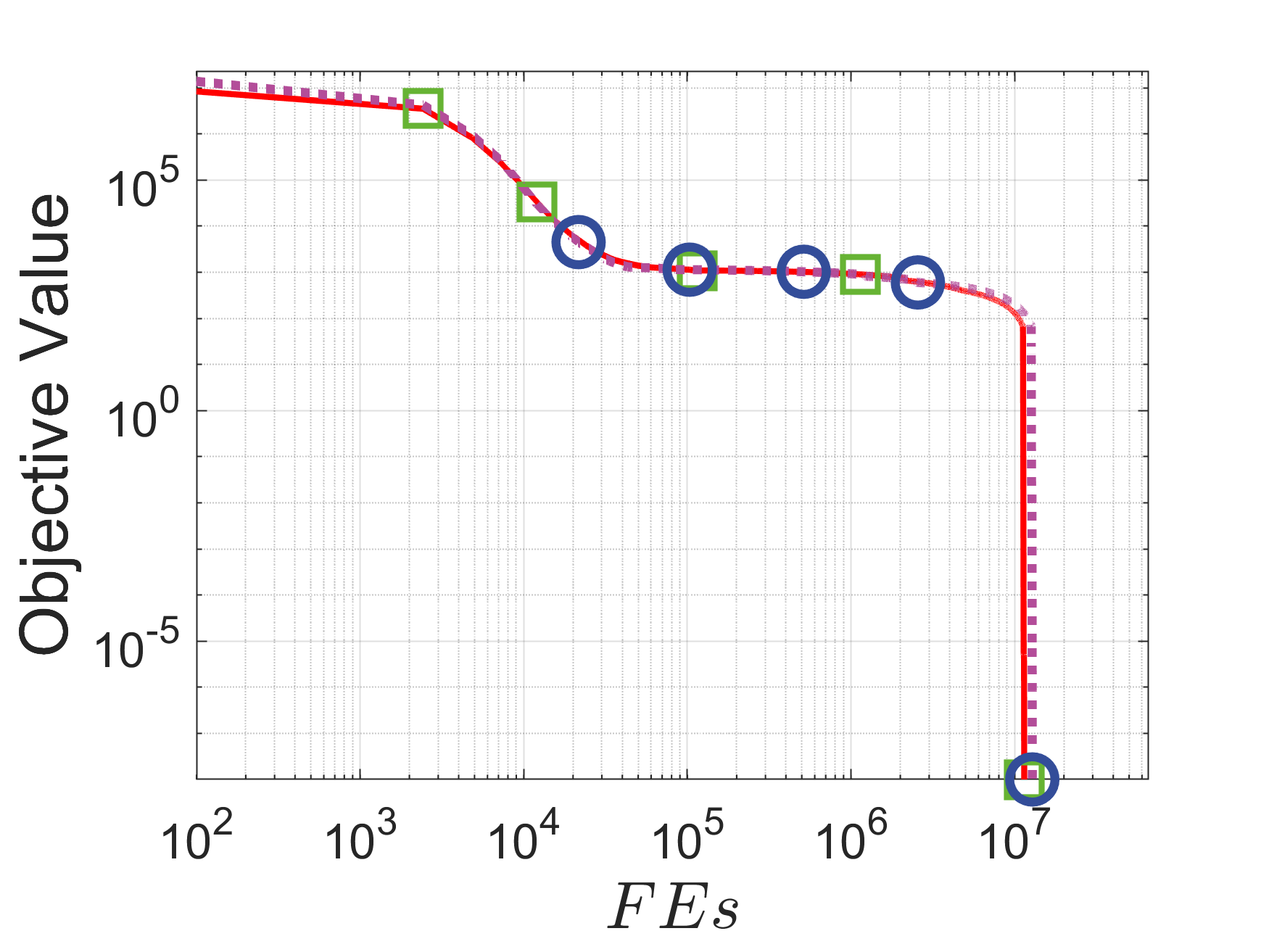}} 
	\subfloat[Discus]{\includegraphics[width=0.24\textwidth]{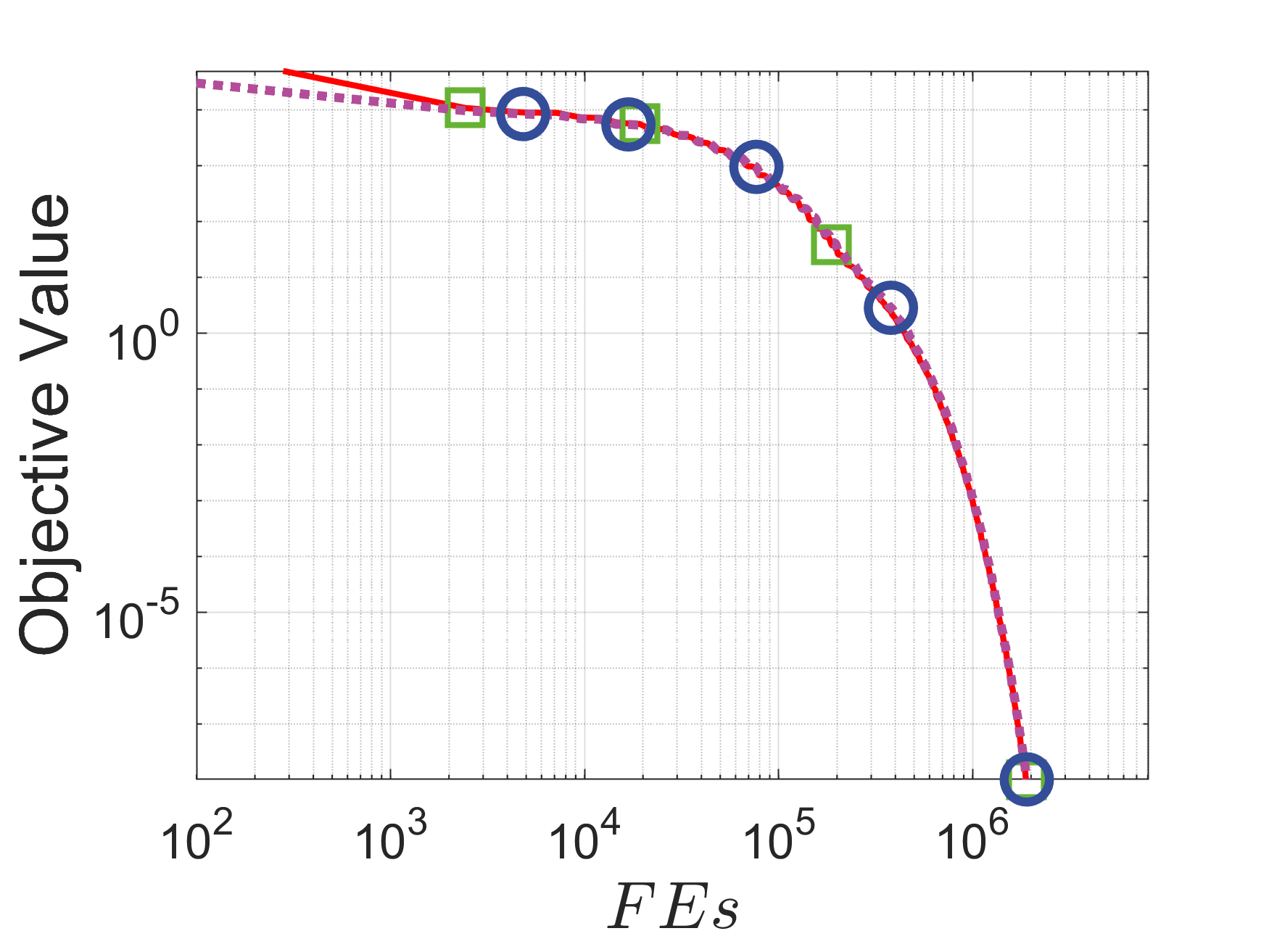}} 
	\hfil
	\subfloat[Cigar]{\includegraphics[width=0.24\textwidth]{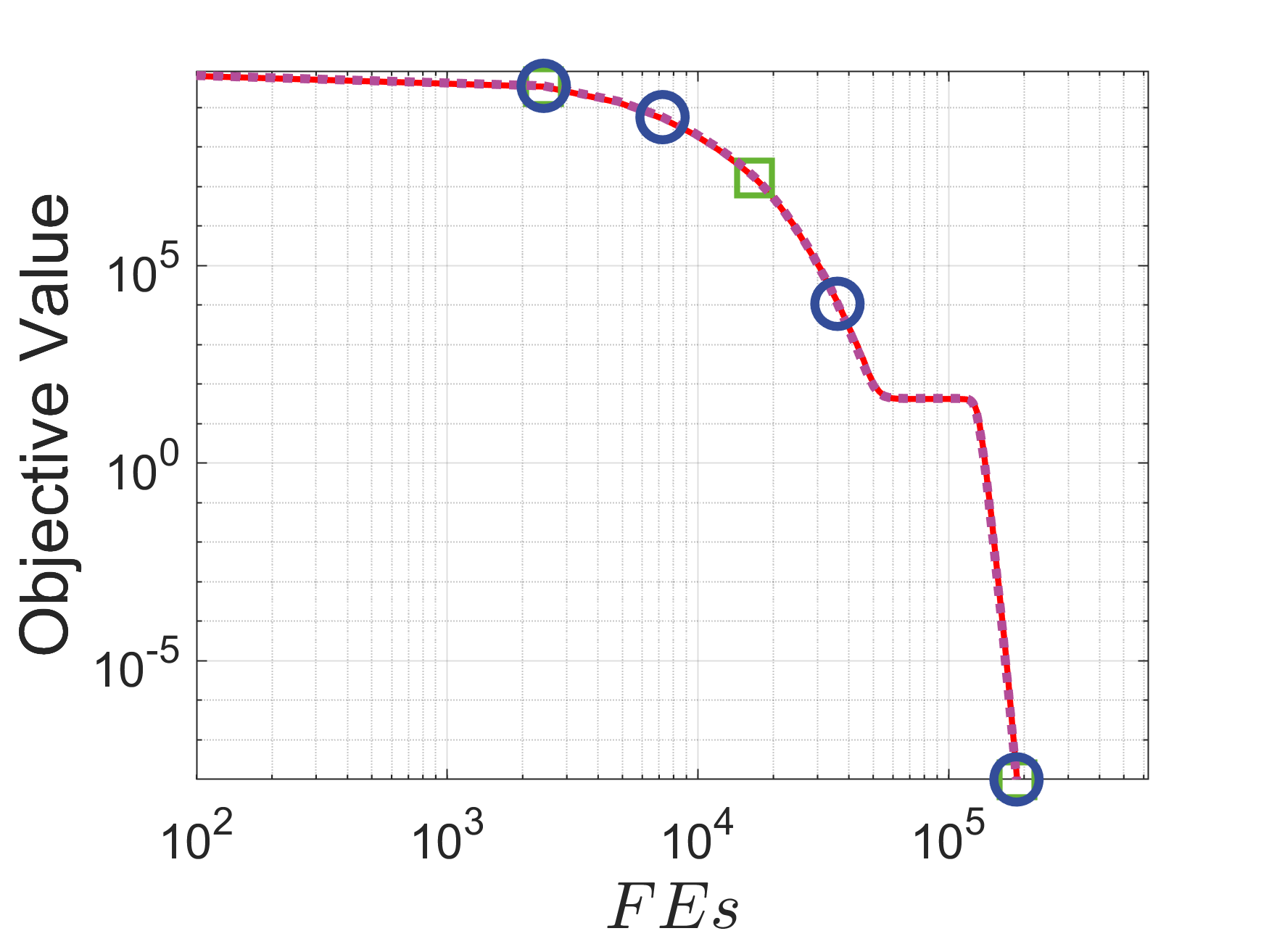}} 
	\subfloat[Different Powers]{\includegraphics[width=0.24\textwidth]{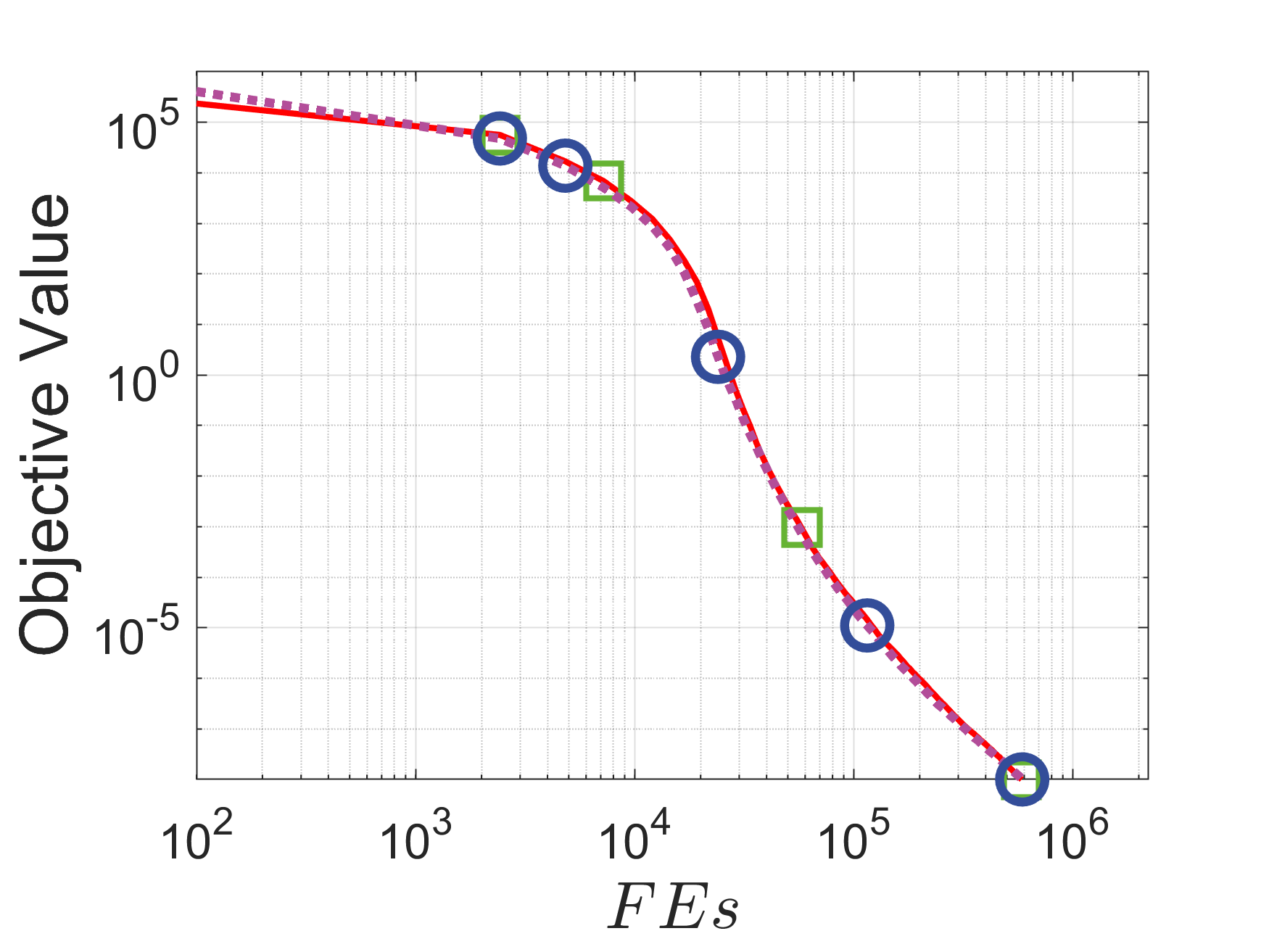}} 
	\caption{MMES with $l=m$ on the 1000-dimensional non-rotated and the corresponding rotated problems, shown by evolutionary trajectories.}
	\label{fig:MMES_msm_rotational_invariance}
\end{figure}

The experiment considers the setting $l=m$. We perform MMES on the 1000-dimensional non-rotated problems $\fElli,\fRosen,\fDiscus,\fCigar,\fDiffPow$ and their rotated versions $\fRotElli, \fRotElli,\fRotDiscus,\fRotCigar,\fRotDiffPow$ and then compare the corresponding convergence curves. As shown in \Cref{fig:MMES_msm_rotational_invariance}, MMES behaves almost the same on a non-rotated problem and its rotated counterpart. That is, increasing $l$ does not impact the rotational invariance of MMES. One possible reason is that, by initializing the $\bm{q}$ vectors to zeros, MMES imposes a strict limit to the degrees of freedom in the initial distribution which in turn reduces its impact on the algorithm performance.

\subsection{Relation to the problem}
The results in \Cref{fig:additional-parameter-test-scalability} show that, even when the time budget is not limited, no one value of the mixing strength $l$ is uniformly preferable. A larger $l$ does not necessary improve performance, but may, in fact, degrade it. Note that this does not contradict the theoretic analyses provided in \Cref{sec:FMS} since $l$ only influences the approximation to the Gaussian distribution $\mathcal{P}_a$ while the latter is not guaranteed to be the best distribution for guiding the search. The strict dependence of the optimal $l$ on the problem type is unclear; but below we carry out a primary study which may provide practical hints for choosing the mixing strength. 

\paragraph{\textbf{Irrelevance of approximation accuracy and algorithm performance}} It should be noted firstly that a better approximation to the Gaussian distribution does not lead to a better performance. We state here two reasons for which the Gaussian distribution is so widely used in modern ESs, with the aim of illustrating that the problem characteristics are usually not taken into consideration in the design of Gaussian based ESs.

The first reason is that the Gaussian distribution is preferable for black-box optimization as it has maximal entropy among all continuous distributions~\cite{hansen_evolution_2015}. The use of the Gaussian distribution is mainly to achieve uniformly good performance and have no explicit relation to specific characteristics of the problems. According to the No Free Lunch theorem~\cite{wolpert_no_1997}, one can always find a certain type of problems on which ESs with a non-Gaussian distribution works better than with a Gaussian distribution. Thus, using a worse approximation to achieve a better performance does not come as a surprise.

The second reason is more concrete and limited to the class of quadratic problems. Existing studies suggested that the standard ES with an isotropic Gaussian exhibits linear convergence on quadratic problems and the convergence rate is inversely related to the condition number of the Hessian~\cite{akimoto_quality_2018}. Modern ESs generally adapt a Gaussian distribution to the problem landscape which then acts as a pre-conditioner to eliminate the adverse effect of ill-conditioning~\cite{beyer_convergence_2014}. However, the usefulness of the Gaussian distribution is heuristic based and often relies on the assumption that the variable correlations can be full captured by the maintained covariance matrix. In real scenes of large-scale optimization, there is no theoretical guarantee that the Gaussian distribution works the best when the correlations can only be partially learned. 

\paragraph{\textbf{An example for which a worse approximation may work better}} We give an example to show that the approximation distribution $\mathcal{P}_m$ can work better than the target distribution $\mathcal{P}_a$. Consider a problem $f(\bm{x}) = \bm{x}^T \bm{H} \bm{x}$ to be minimized from a starting point $\bm{x}_0$ sufficiently far from the optimum $\bm{0}$, where $\bm{H}$ is the symmetric positive definitive Hessian. Denote $\bm{u}$ as the eigenvector of $\bm{H}$ corresponding to the smallest eigenvalue. We omit the mutation strength for simplicity and focus on analyzing the samples in a single step. The quality of a sample $\bm{y}$ is dominated by the distance of its projection on $\bm{u}$ to $\bm{0}$, given by $|\bm{u}^T \bm{y}|$. We further assume $\bm{u} \notin span(\bm{q}_1,\cdots,\bm{q}_n)$, which simulates the situation when the probability model fails to capture the most promising search direction. Now consider two samples $\bm{y}_a$ and $\bm{y}_m$ sampled from $\mathcal{P}_a$ and $\mathcal{P}_m$, respectively.
From the properties of multivariate Gaussian distribution we have $\bm{u}^T \bm{y}_a \sim \mathcal{N}(\bm{u}^T \bm{x}_0,\bm{u}^T \bm{C}_a \bm{u})$ which is reduced to $\mathcal{N}(\bm{u}^T \bm{x}_0,(1-c_a)^m)$. Regarding $\bm{y}_m$ we have $E[\bm{u}^T \bm{y}_m] = \bm{u}^T \bm{x}_0$ and $\mathbb{V}[\bm{u}^T \bm{y}_m] = (1-c_a)^m$, where the first equality holds trivially and the second is according to \Cref{theorem:second-order-moment-for-FMS}. That is, the improvements made by $\bm{y}_a$ and $\bm{y}_m$, along the direction of $\bm{u}$, have the same first- and second- order moments. Despite this similarity, $\bm{u}^T \bm{y}_a$ has zero excess kurtosis since it is normally distributed while $\bm{u}^T \bm{y}_m$ has positive excess kurtosis, as proved by \Cref{theorem:non-negative-excess-kurtosis}. 
This means there exists a region around $\bm{0}$ in which the density of $\mathcal{P}_m$ is higher while simultaneously decaying slower than $\mathcal{P}_a$, and therefore, we can find a positive $\epsilon$ such that $P\{-\epsilon <\bm{u}^T \bm{y}_a<\epsilon\} < P\{-\epsilon <\bm{u}^T \bm{y}_m<\epsilon\}$. This indicates $\mathcal{P}_m$ is more likely to produce samples close to the optimum than $\mathcal{P}_a$.

The above example shows that $\mathcal{P}_m$ allows producing longer jumps on promising directions that have not been captured by the search distribution. This also accounts for the situation where there exists more promising search directions than the direction vectors used in the search distribution. A concrete example reflecting this situation is the Discus function $\fDiscus$. The Hessian of this problem has two distinct eigenvalues: 2 of multiplicity $n-1$ and $2\times 10^6$ of multiplicity 1. Thus, it has $n-1$ promising directions required to be explored. However, most large-scale variants of CMA-ES only utilize $m \ll n$ direction vectors and will in no way explore these directions simultaneously. The use of $\mathcal{P}_m$ will benefit faster exploration on these directions, which explains why MMES performs better than other Gaussian based ESs. In addition, \Cref{theorem:non-negative-excess-kurtosis} states that the decay rate of the density of $\mathcal{P}_m$ is proportional to the mixing strength. This explains why on $\fDiscus$ MMES works better with a small $l$ than with a large $l$.

\paragraph{\textbf{Numerical simulations}}
In the next we give empirical evidences to the above discussion. The aim of the performed experiment is to verify it is the ability of producing long jumps allowing MMES to perform better than other algorithms on $\fDiscus$. To this end, we modify the sampling operation of MMES which we restate below
\begin{equation}
	\bm{z} = \sqrt{1-\gamma} \bm{z}_0 + \sqrt{\frac{\gamma}{l}} \sum_{j=1}^l \bm{q}_{i_j} z_j.
\end{equation}
In this experiment, the $\bm{q}$ vectors, the mixing components $z_1,\cdots,z_m$, and the hyper-parameter $\gamma$ remain unchanged. $\bm{z}_0$ is the $n$-dimensional isotropic Gaussian in MMES but here we replace it with two other types of random variable. The first is a random vector with each element drawn from a t-distribution with 5 degrees of freedom, followed by a proper normalization to have a unit variance. The resultant distribution has the same variance as that of MMES but will possess a slightly larger excess kurtosis than MMES. The second is a random vector drawn from the Rademacher distribution~\cite{loshchilov_lm-cma:_2017}. The Rademacher distribution mimics the statistical property of Gaussian in high-dimensional space but has a negative excess kurtosis. The MMES variants with t-distribution and Rademacher distribution are denoted by MMES-T and MMES-R, respectively. Intuition suggests that, on $\fDiscus$, due to the different behaviors in producing long jumps, MMES-T will perform better than MMES while MMES-R will perform worse.

\begin{figure}[tb]
	\centering
	\subfloat[$\fElli$]{\includegraphics[width=0.24\textwidth]{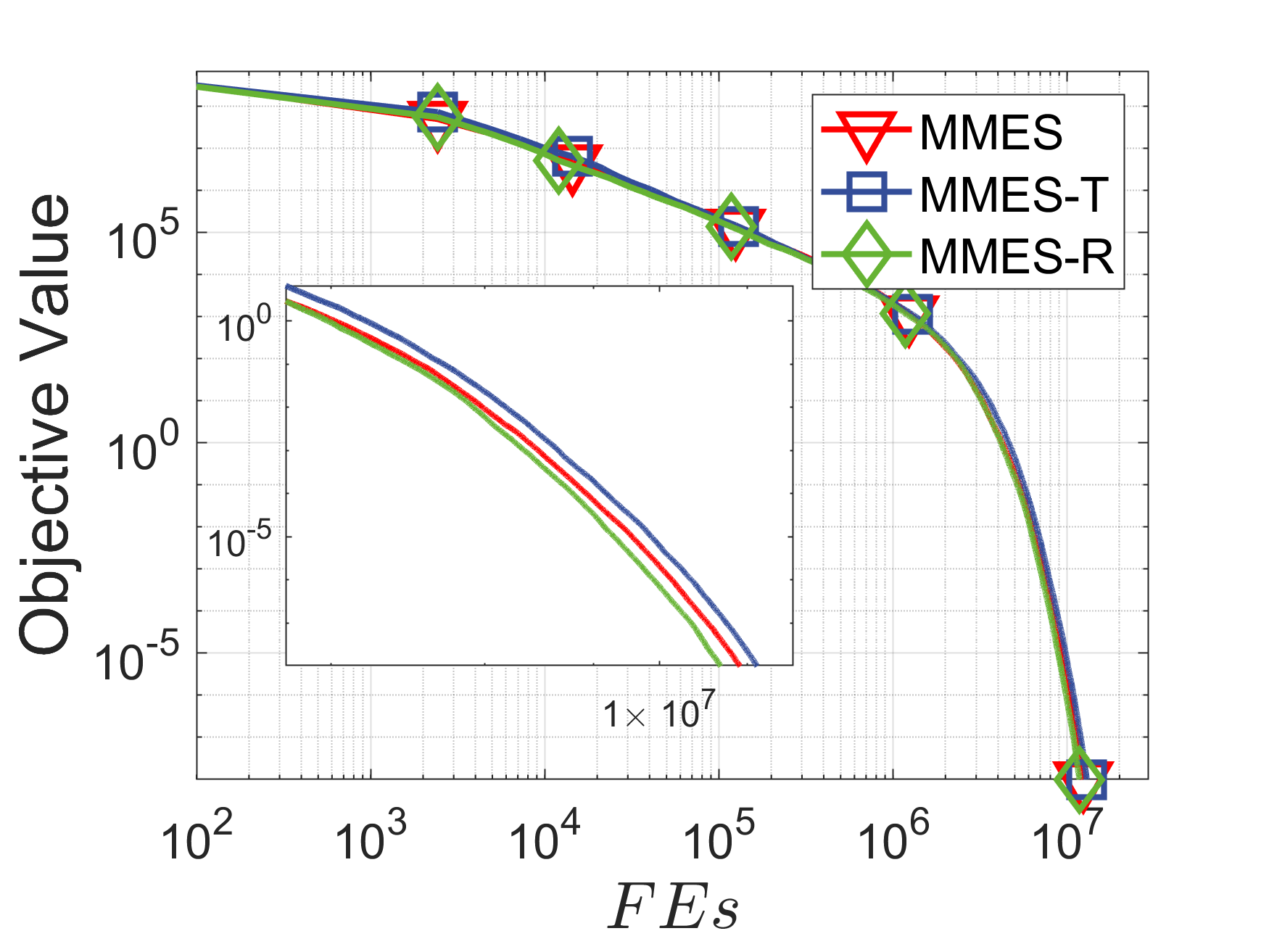} }
	\subfloat[$\fDiscus$]{\includegraphics[width=0.24\textwidth]{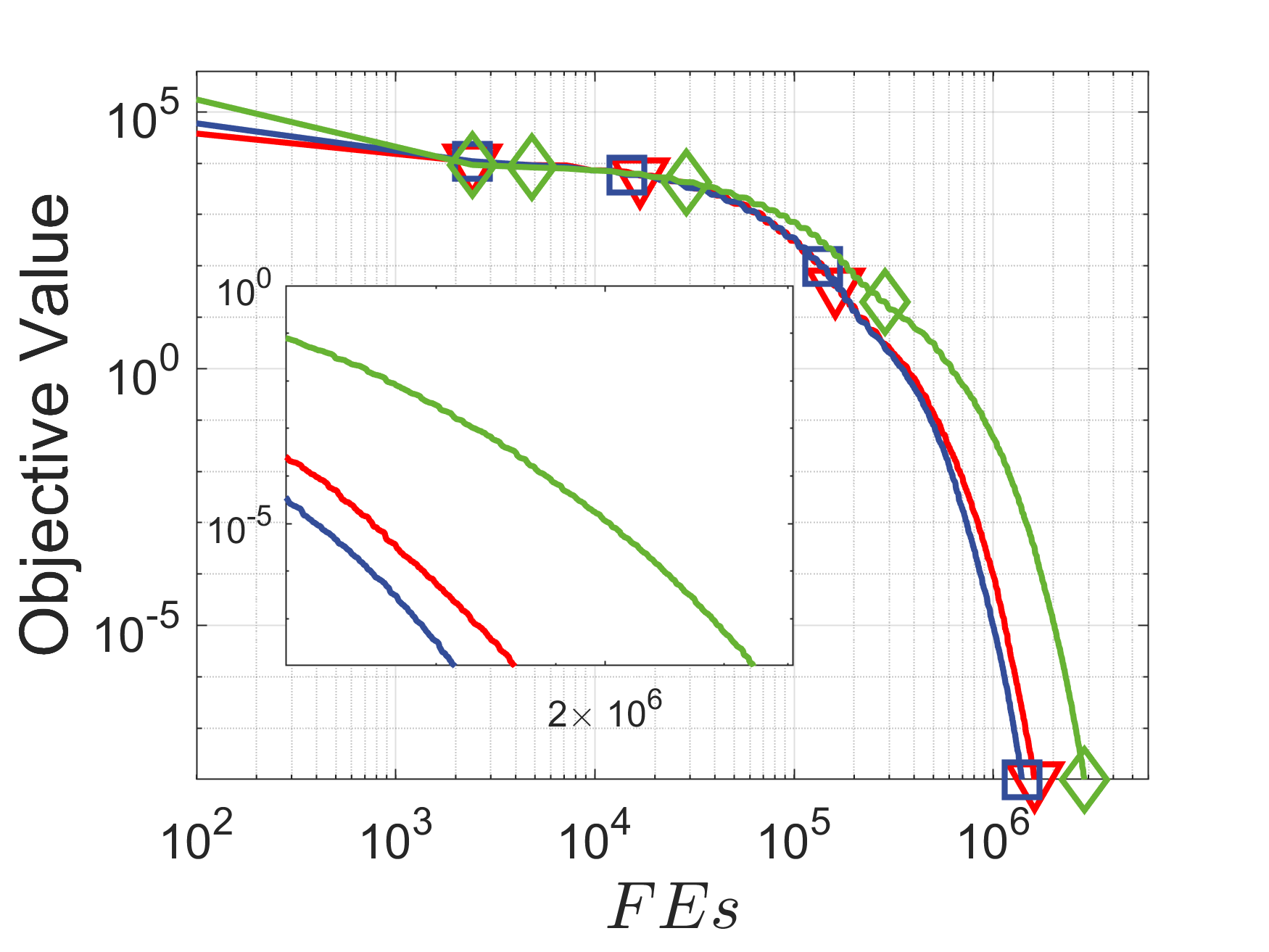} }
	\hfil
	\subfloat[$\fCigar$]{\includegraphics[width=0.24\textwidth]{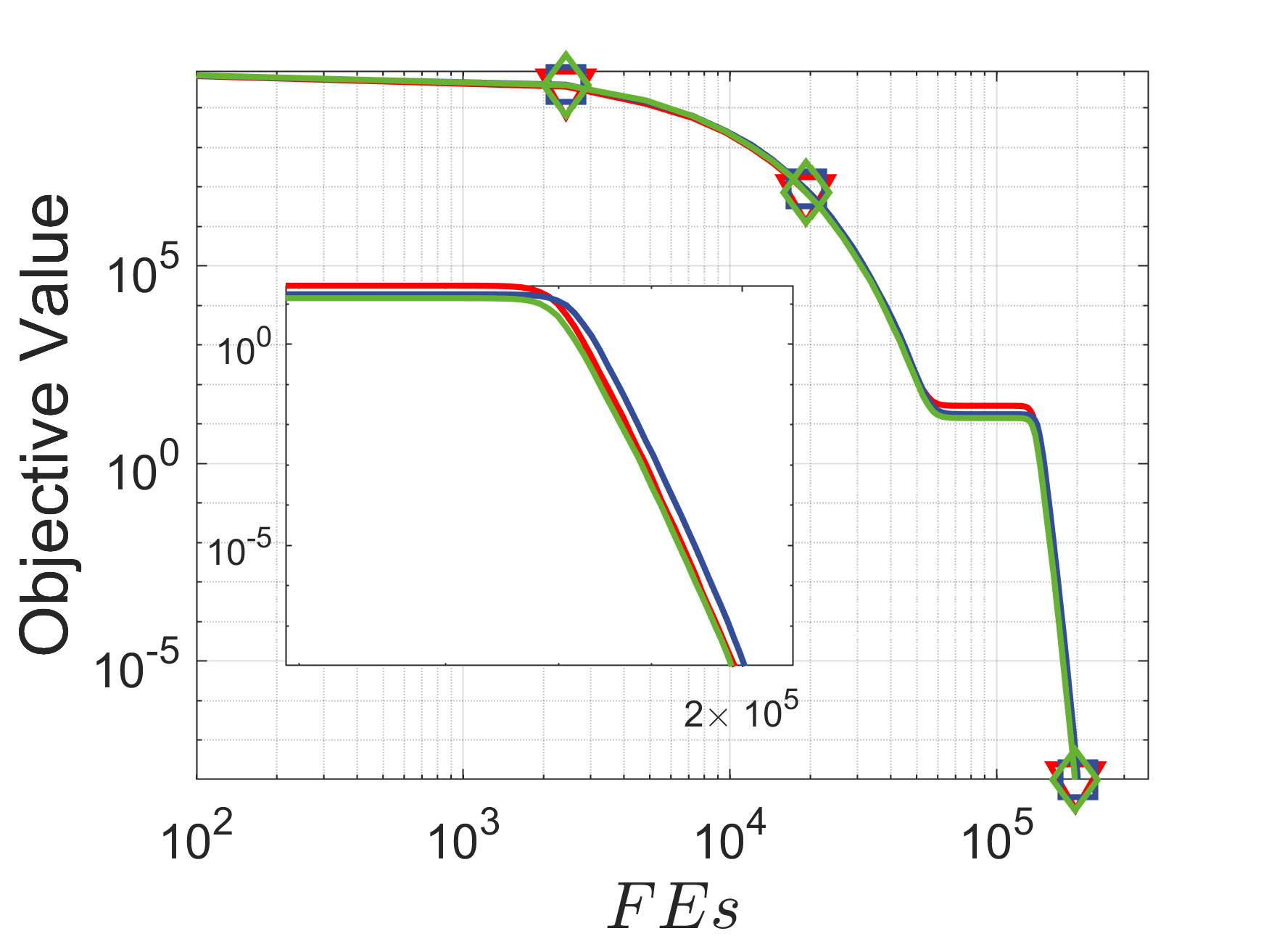} }
	\subfloat[$\fDiffPow$]{\includegraphics[width=0.24\textwidth]{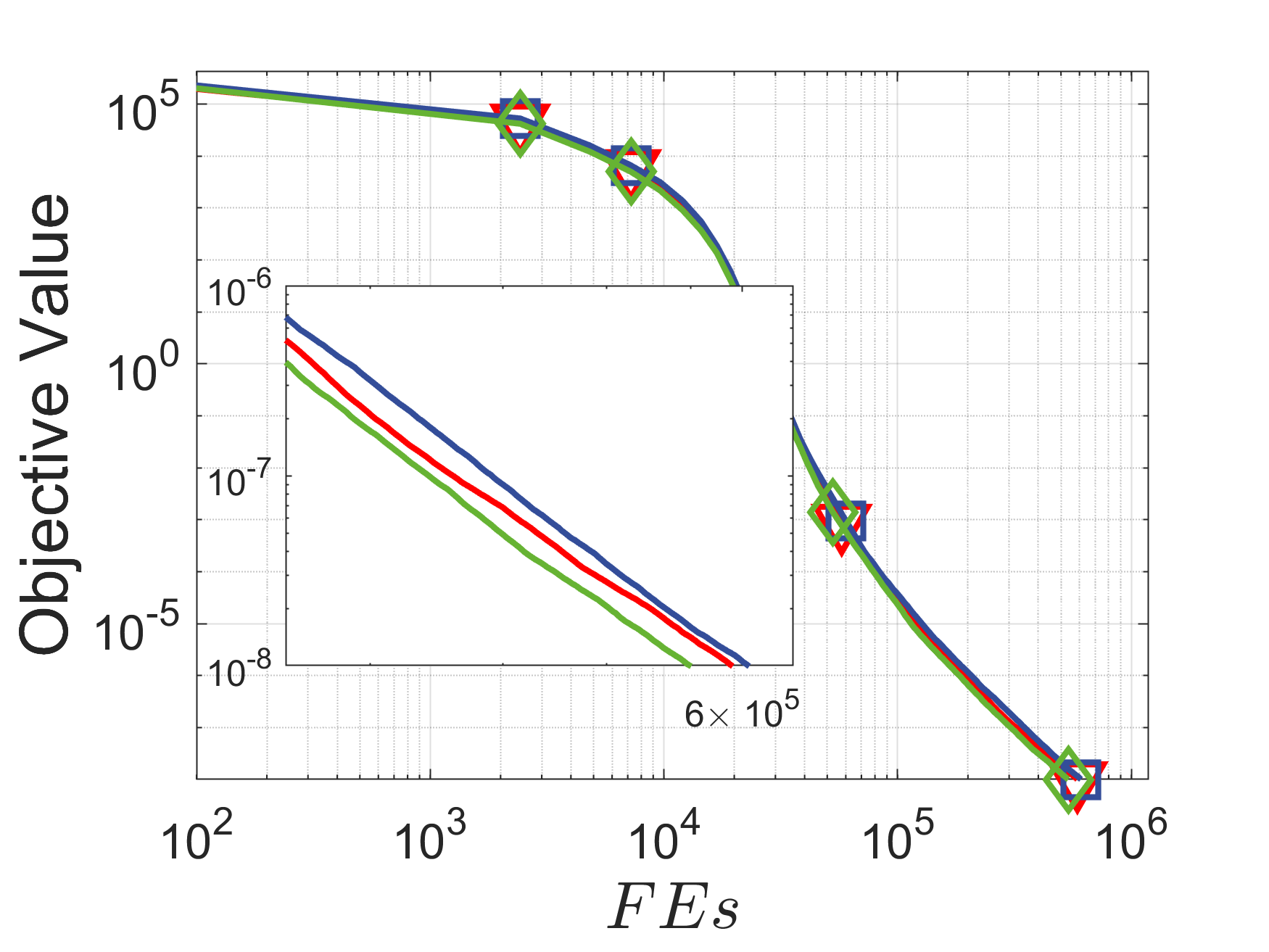} }
	\caption{MMES with different distributions on 1000-dimensional non-rotated problems. The mixing strength $l=4$. Plots have been zoomed in to show the small difference in convergence curves.}
	\label{fig:MMES-different-distributions-1000d}
\end{figure}

\Cref{fig:MMES-different-distributions-1000d} plots for MMES and its two variants the convergence curves on four 1000-dimensional problems. As expected, the t-distribution leads to a better performance on $\fDiscus$ while the Rademacher distribution degrades the performance significantly. On the contrary, the $\fCigar$ problem has only one promising search direction and is less influenced by the distribution type, so the two variants show no difference from MMES. These observations confirm our statement about the relation between distribution type and landscape characteristics. $\fElli$ and $\fDiffPow$ can be regarded as between the above two extreme cases: the Hessian has widely spread eigenvalues. A properly designed ES usually capture some of the directions that contribute mostly to decreasing the objective value. Producing long jumps may degrade the exploitation ability, which illustrates the better performance of MMES-R and the worse performance of MMES-T. Nevertheless, the difference in convergence behaviors of these three competitors is insignificant, probably due to that all their distributions have densities decaying exponentially fast.

\paragraph{\textbf{Further discussion}}
The above experiments show that a small mixing strength enhances the exploration ability by increasing the likelihood of producing long jumps. In fact, similar studies can be found in the field of multimodal optimization. In \cite{yao_evolutionary_1999,sanyang_heavy_2015,schaul_high_2011} the authors used long-tailed distributions, such as the t-distribution and the Cauchy distribution, to prevent from being trapped into local optima. The difficulty they were to solve, albeit in a different context, share an essential similarity to large-scale optimization. Specifically, the difficulty in handling multimodality for a Gaussian based ES (or EDA) comes from the fact that the Hessian of local landscape may be unbounded but the covariance matrix is bounded by design. On the other hand, the difficulty for solving large-scale optimization is due to the Hessian may have many sensitive directions (i.e., the directions of eigenvectors corresponding to smallest eigenvalues) while the covariance matrix has a low-rank structure and hence lacks the ability of capturing all these directions. In a unified view, the above two difficulties can be considered as caused by the inconsistence in characteristics between the problem landscape and the probability distribution, or more concretely, the inconsistence in the principal components of the inverse of Hessian and the covariance matrix. Experiments presented above show that using a small mixing strength efficiently handles this inconsistence issue while causing no significant adverse effect in other situations. 

\subsection{Practical hints for mixing strength selection}
When applying MMES in practice we suggest always choosing a small constant value for the mixing strength, say $l < 10$. The benefit, as revealed by the experiments, is threefold: 1) it improves the performance in coping with the inconsistence between the characteristics of problem landscape and probability distribution; 2) it is robust on other types of problems where the aforementioned inconsistence is insignificant; 3) it reduces the computational time. Manually tuning $l$ for certain problems may be useless, as the improved performance may not compensate for the increased execution time. Thus, it is recommended using MMES with a fix and small $l$ in all application scenarios.

\section{Experiment on the CEC'2013 test set}
To further investigate the performance of MMES in handling complicated problems, we carry out more experiments using the recent CEC'2013 large-scale test set~\cite{li_benchmark_2013}.
\subsection{Test Problems}
The CEC'2013 test set contains three fully separable problems ($f_1$ to $f_3$), eight partially non-separable problems ($f_4$ to $f_{11}$), three overlapping problems ($f_{12}$ to $f_{14}$), and one fully non-separable problem ($f_{15}$). 
Among the partially non-separable problems, $f_{8}$ to $f_{11}$ consist of several fully non-separable subcomponents and no correlation exists between different subcomponents. $f_3$ to $f_7$ are respectively identical to $f_8$ to $f_{11}$ except that one of the subcomponents is made fully separable. Hence, they exhibit more separability than $f_8$ to $f_{11}$ do.
The overlapping problems are constructed in a similar way; the only difference is that the adjacent subcomponents are not disjoint but have overlapping variables. Generally, this test set is more challenging than the CEC'2010 test set in that 1) the subcomponent partitioning is non-uniform and imbalanced and 2) the local landscape near the optimum is asymmetric and irregular.

\subsection{Experimental Settings}
The same algorithms in the previous experiment with the CEC'2010 test set, namely DECC-G~\cite{yang_large_2008}, MA-SW~\cite{molina_memetic_2011}, MOS~\cite{latorre_large_2013}, CCPSO2~\cite{li_cooperatively_2012}, and DECC-DG~\cite{omidvar_cooperative_2014}, are selected as competitors. Their corresponding results are from~\cite{yang_level-based_2017,latorre_large_2013}, measured in the standard settings for the CEC competitions~\cite{li_benchmark_2013}. MMES is configured as described in Section IV-A. The final results are obtained from 25 independent runs with a fixed budget of $3\times 10^6$ function evaluations, which is also according to the guideline of the CEC competitions.

\subsection{Numerical Results}
\Cref{tab:cec13} presents the median results of MMES and the five competitors. MMES ranks first and has a significantly better overall performance than the three CC based algorithms, as suggested by the multi-comparison. We see that MMES is robust to the non-separability of the problem, and can produce the best or the second best results once the problems are not fully separable. The two memetic algorithms (i.e., MA-SW and MOS), on the contrary, appear to show clear preference for solving fully separable problems (e.g., $f_1$) or partially separable problems that involve separable subcomponents (e.g., $f_7$ and $f_8$).

The above observations regarding MMES largely coincide with what we have obtained on the CEC'2010 test set. It is probably due to the implicit similarities between the CEC'2010 test set and the 2013 one. In deed, the major distinction between the two sets is the non-uniform and overlapping scheme in partitioning the subcomponent, which is in essence equivalent to an affine transformation on the decision space. Therefore, MMES behaves similarly on these two sets due to its inherent affine invariance. 

The CEC'2013 test set also introduces local irregularity and asymmetry to render deceptive landscape for multimodal problems, causing difficulties for algorithms that exploit the search space symmetrically. Nevertheless, their influence on the performance of MMES appears to be marginal. For example, MMES can produce competitive results on all instances of the Ackley's problem ($f_3, f_6$, and $f_{10}$), regardless of its local irregularity and asymmetry. One possible reason is that the landscape of this problem has a strong global structure while MMES with restarts has enhanced ability in global exploration. The difficulties caused by the local irregularity and asymmetry can also be compensated by the capacity in handling variable correlations. This can be verified by observing that, on the Rastrigin's problem, MMES performs relatively poor on the separable instance $f_2$ while becoming much better on the partially non-separable instances $f_5$ and $f_9$. We may conclude that the local irregularity and asymmetry are not the crucial factors in determining the hardness of the test problems and their impact on MMES seems to be insignificant.

\begin{table*}[htbp]
	\centering
	\scriptsize
	\caption{Median of the objective values obtained on the CEC'2013 test problems. The best and the second best results for each test instance are shown with dark and light gray background, respectively.}
	\setlength{\tabcolsep}{12pt}
	\label{tab:cec13}
	\begin{threeparttable}
\begin{tabular}{ccccccc}
\toprule
  & DECC-G & MA-SW & MOS & CCPSO2 & DECC-DG & MMES \\
\midrule
$f_{1}$ & $2.06E-06$ $\circ$ & \cellcolor[rgb]{ .816,  .816,  .816} $7.90E-21$ $\circ$ & \cellcolor[rgb]{ .627,  .627,  .627} $0.00E+00$ $\circ$ & $2.79E+01$ $\circ$ & $6.03E+02$ $\circ$ & $1.05E+04$  \\
$f_{2}$ & $1.30E+03$ $\bullet$ & \cellcolor[rgb]{ .816,  .816,  .816} $6.85E+02$ $\circ$ & $8.36E+02$ $\circ$ & \cellcolor[rgb]{ .627,  .627,  .627} $3.48E+01$ $\circ$ & $1.28E+04$ $\bullet$ & $8.78E+02$  \\
$f_{3}$ & $2.02E+01$ $\bullet$ & $2.03E+01$ $\bullet$ & \cellcolor[rgb]{ .816,  .816,  .816} $9.10E-13$ $\bullet$ & $2.00E+01$ $\bullet$ & $2.14E+01$ $\bullet$ & \cellcolor[rgb]{ .627,  .627,  .627} $0.00E+00$  \\
$f_{4}$ & $2.00E+11$ $\bullet$ & $5.19E+09$ $\bullet$ & \cellcolor[rgb]{ .816,  .816,  .816} $1.56E+08$ $\bullet$ & $3.20E+10$ $\bullet$ & $7.33E+10$ $\bullet$ & \cellcolor[rgb]{ .627,  .627,  .627} $7.85E+07$  \\
$f_{5}$ & $8.44E+06$ $\bullet$ & \cellcolor[rgb]{ .816,  .816,  .816} $1.74E+06$ $\bullet$ & $6.79E+06$ $\bullet$ & $1.30E+07$ $\bullet$ & $5.81E+06$ $\bullet$ & \cellcolor[rgb]{ .627,  .627,  .627} $8.49E+05$  \\
$f_{6}$ & $1.06E+06$ $\bullet$ & $1.05E+06$ $\bullet$ & \cellcolor[rgb]{ .816,  .816,  .816} $1.39E+05$ $\bullet$ & $1.05E+06$ $\bullet$ & $1.06E+06$ $\bullet$ & \cellcolor[rgb]{ .627,  .627,  .627} $5.70E+03$  \\
$f_{7}$ & $1.04E+09$ $\bullet$ & $2.98E+06$ $\bullet$ & \cellcolor[rgb]{ .627,  .627,  .627} $1.62E+04$ $\circ$ & $1.29E+08$ $\bullet$ & $4.25E+08$ $\bullet$ & \cellcolor[rgb]{ .816,  .816,  .816} $2.06E+04$  \\
$f_{8}$ & $7.90E+15$ $\bullet$ & $1.26E+14$ $\bullet$ & \cellcolor[rgb]{ .816,  .816,  .816} $8.08E+12$ $\bullet$ & $8.16E+14$ $\bullet$ & $2.89E+15$ $\bullet$ & \cellcolor[rgb]{ .627,  .627,  .627} $5.47E+11$  \\
$f_{9}$ & $5.86E+08$ $\bullet$ & \cellcolor[rgb]{ .627,  .627,  .627} $1.07E+08$ $\circ$ & $3.87E+08$ $\bullet$ & $3.63E+09$ $\bullet$ & $4.95E+08$ $\bullet$ & \cellcolor[rgb]{ .816,  .816,  .816} $1.26E+08$  \\
$f_{10}$ & $9.30E+07$ $\bullet$ & $9.34E+07$ $\bullet$ & \cellcolor[rgb]{ .627,  .627,  .627} $1.18E+06$ $\circ$ & $9.29E+07$ $\bullet$ & $9.45E+07$ $\bullet$ & \cellcolor[rgb]{ .816,  .816,  .816} $1.25E+06$  \\
$f_{11}$ & $1.26E+11$ $\bullet$ & $4.79E+08$ $\bullet$ & \cellcolor[rgb]{ .816,  .816,  .816} $4.48E+07$ $\bullet$ & $9.38E+11$ $\bullet$ & $3.81E+10$ $\bullet$ & \cellcolor[rgb]{ .627,  .627,  .627} $2.19E+06$  \\
$f_{12}$ & $4.19E+03$ $\bullet$ & $1.34E+03$ $\bullet$ & \cellcolor[rgb]{ .627,  .627,  .627} $2.46E+02$ $\circ$ & $2.10E+03$ $\bullet$ & $1.68E+11$ $\bullet$ & \cellcolor[rgb]{ .816,  .816,  .816} $8.56E+02$  \\
$f_{13}$ & $8.67E+09$ $\bullet$ & $9.72E+08$ $\bullet$ & \cellcolor[rgb]{ .816,  .816,  .816} $3.30E+06$ $\bullet$ & $3.21E+09$ $\bullet$ & $2.08E+10$ $\bullet$ & \cellcolor[rgb]{ .627,  .627,  .627} $2.39E+05$  \\
$f_{14}$ & $1.28E+11$ $\bullet$ & $5.11E+09$ $\bullet$ & \cellcolor[rgb]{ .816,  .816,  .816} $2.42E+07$ $\bullet$ & $5.98E+10$ $\bullet$ & $1.56E+10$ $\bullet$ & \cellcolor[rgb]{ .627,  .627,  .627} $8.11E+06$  \\
$f_{15}$ & $1.13E+07$ $\bullet$ & $7.95E+06$ $\bullet$ & \cellcolor[rgb]{ .816,  .816,  .816} $2.38E+06$ $\bullet$ & $2.72E+06$ $\bullet$ & $9.52E+06$ $\bullet$ & \cellcolor[rgb]{ .627,  .627,  .627} $1.54E+06$  \\
\midrule
$\bullet$ / $\circ$ / $\dagger$ & 14 / 1 / 0 & 12 / 3 / 0 & 10 / 5 / 0 & 13 / 2 / 0 & 14 / 1 / 0 &  \\
\midrule
Avg Rank & 5.1 & 3.03 & 2 & 4.03 & 5.03 & 1.8 \\
\midrule
$p$-Value & 0 & 1 & 1 & 0.0158 & 0 &  \\
\bottomrule
\end{tabular}%

		\begin{tablenotes}
			\item[*] ``Avg Rank'' denotes the ranking results averaged over all problems according to the Friedman test.
			\item[*] ``$p$-Value'' denotes the significance of difference between the averaged ranks of MMES and the pair algorithms, corrected by the Bonferroni procedure. 
		\end{tablenotes}
	\end{threeparttable}
\end{table*}

\section{Practical Application on Sparse Spectral Clustering}
In this section, we test the effectiveness of MMES on sparse spectral clustering, an emerging application in unsupervised learning.
\subsection{Problem Descriptions}
Spectral clustering (SC) is a pioneered clustering method that has numerous applications in the field of data mining~\cite{von_luxburg_tutorial_2007}. It has a easy-to-compute solution and facilitates exploiting the underlying data structure. Given a set of $N_f$-dimensional data points $\bm{x}_1,\bm{x}_2,\cdots,\bm{x}_{N_s}$ and a matrix $\bm{W}\in R^{N_s \times N_s}$ with the entry $W_{i,j}$ measuring some notion of similarity between $\bm{x}_i$ and $\bm{x}_j$, SC aims to find a partition of the data points that minimizes the between-cluster similarities and maximizes the within-cluster similarities. The partition result is usually stored in an indicator matrix $\bm{U}\in R^{N_s\times N_c}$, where $N_c$ is the number of clusters and $U_{i,j}$ indicates the possibility of $\bm{x}_i$ belonging to the $j$-th cluster. The $i$-th row of $\bm{U}$ then represents a low-dimensional embedding of $\bm{x}_i$ and the final partition result can be obtained by performing a simple clustering (e.g., $k$-means) in the embedding space.

In the standard implementation of SC, the matrix $\bm{U}$ can be obtained by solving $\min \limits_{\bm{U}^T\bm{U}=\bm{I}_{N_c}} Tr[\bm{U}^T\bm{L}\bm{U}]$ where $\bm{L} \in R^{N_s\times N_s}$ is the normalized Laplacian determined by $\bm{W}$ and $Tr[\cdot]$ denotes the matrix trace operation. A recent study~\cite{lu_convex_2016} suggested that, in a perfect clustering, the covariance matrix of the data embeddings is a permutation of the identity matrix, thereby being sparse in nature. This implies that imposing additional sparsity constraint on $\bm{U} \bm{U}^T$ may improve the clustering performance, leading to the sparse SC problem in the following form: 

\begin{equation}
	\begin{aligned}
		\min \limits_{\bm{U} \in R^{N_s\times N_c}} \;\;&Tr[\bm{U}^T\bm{L}\bm{U}] + \beta |\bm{U}\bm{U}^T|_1 \\
		s.t. \;\;&\bm{U}^T\bm{U}=\bm{I}_{N_c}.
	\end{aligned}
	\label{eq:SSC-objective}
\end{equation}
where $|\cdot|_1$ is the $l_1$ norm and $\beta$ is a coefficient trading off the objective of SC and the sparsity constraint. The sparse SC in~\Cref{eq:SSC-objective} does not have a closed-form solution and poses difficulties in two aspects: 1) the search region is not convex and 2) the objective function is not differentiable. The non-convexity and non-differentiability motivate us to use MMES in solving the sparse SC problem.


\subsection{Adaption of MMES in Riemannian Settings}
Special treatment is required in handling the orthogonal constraint in~\Cref{eq:SSC-objective} since MMES is proposed for unconstrained optimization. One can realize that the sparse SC is actually defined on a Riemannian manifold termed as the Grassmann manifold~\cite{edelman_geometry_1998}. On the other hand, as mentioned in Section IV-C, all genetic operations of MMES are linear. This indicates that MMES is still applicable as long as all its operations are appropriately defined on the tangent spaces of the manifold.

In this experiment, we choose the retraction based framework~\cite{absil_projection-like_2012} to handle the Grassmann manifold constraint. Retraction is an approximation of the geodesic which maps a matrix from the tangent space to the manifold. With retraction, MMES may operate on the tangent spaces as if on the manifold, due to the local homeomorphism property. To work with retraction, we modify MMES as follows:

\begin{itemize}
	\item At the iteration $g$, choose the tangent space corresponding to the mean $\bm{m}^{(g)}$ as the workspace. Sample a subset of direction vectors with the probability distribution $\mathcal{P}_{\bm{i}}$ and then parallel translate them to this workspace.
	\item In the chosen workspace, perform the same sampling and updating operations as in the standard MMES, except for changing the objective function from $f$ to $f\circ r$ where $r$ is the retraction operator.
	\item At the end of the iteration, retract the new mean $\bm{m}^{(g+1)}$ onto the manifold to enforce feasibility.
\end{itemize}

The above modifications make sure that 1) each iteration of MMES works in a certain tangent space and 2) the probability model reconstructed from the direction vectors are well-defined in that space. The specification of the tangent space is to preserve the unbiasedness of ESs. That is, under random selection, the mean will stay unchanged in expectation and the Riemannian MMES will degenerate to a Euclidean one which runs in a fixed tangent space. The Riemannian version of MMES is implemented in the Manopt toolbox~\cite{boumal_manopt_2013}.

\subsection{Comparative Algorithm}
We select the recently proposed Grassmann manifold optimization assisted sparse spectral clustering (GSC)~\cite{wang_grassmannian_2017} as a competitor. GSC incorporates a Riemannian trust region framework to improve the global exploration ability on the highly non-convex landscape. Each subproblem in GSC is solved by the truncated Newton method; the termination criteria of the inner conjugate gradient iterations are chosen to achieve superlinear convergence. Unlike MMES, GSC is configured as a white-box solver. However, since the objective function of~\Cref{eq:SSC-objective} is not differentiable, the subgradient is supplied instead in calculating the Riemannian gradient. In this work, GSC is also implemented with the Manopt toolbox.

\subsection{Experimental Settings}
The experiment is performed on 18 real-world data sets consisting of numeric data, figures, and text. All these sets are collected in the scikit-learn toolbox~\cite{pedregosa_scikit-learn:_2011} and are briefly summarized in~\Cref{tab:SSC-data-sets}. We construct the similarity matrix $\bm{W}$ using the $k$-nearest neighbor graph with $k=5$ and the local similarity is measured using the Gaussian similarity function. The coefficient $\beta$ in~\Cref{eq:SSC-objective} is usually problem-dependent, so we choose it from a candidate set $\{0.01, 0.005, 0.001, 0.0005, 0.0001, 0.00005, 0.00001\}$.

The GSC terminates when the number of iterations exceeds 300 or the norm of gradient is smaller than $10^{-6}$. Other hyperparameters are set according to the literature~\cite{wang_grassmannian_2017} and the default settings of the Manopt toolbox. For MMES, we stop the algorithm when the mutation strength is smaller than $10^{-6}$ or the number of function evaluations exceeds $3\times 10^5$. Other settings are the same as in previous experiments. Both MMES and GSC are run 20 times independently and the median results are reported.

\begin{table}[htb]
	\centering
	\caption{Data Sets for Sparse Spectral Clustering}
	\setlength{\tabcolsep}{5pt}
	\label{tab:SSC-data-sets}
	\begin{tabular}{llll}
		\toprule
		Data sets & Samples ($N_s$) & Features ($N_f$) & Classes ($N_c$)\\
		\midrule
		orlraws10P & 100 & 10304 & 10 \\
		pixraw10P & 100 & 10000 & 10 \\
		Prostate-GE & 102 & 5966 & 2 \\
		TOX-171 & 171 & 5748 & 4 \\
		warpAR10P & 130 & 2400 & 10 \\
		Yale & 165 & 1024 & 15 \\
		ALLAML & 72 & 7129 & 2 \\
		warpPIE10P & 210 & 2420 & 10 \\
		GLI-85 & 85 & 22283 & 2 \\
		GLIOMA & 50 & 4434 & 4 \\
		leukemia & 72 & 7070 & 2 \\
		lymphoma & 96 & 4026 & 9 \\
		nci9 & 60 & 9712 & 9 \\
		SMK-CAN-187 & 187 & 19993 & 2 \\
		arcene & 200 & 10000 & 2 \\
		Carcinom & 174 & 9182 & 11 \\
		CLL-SUB-111 & 111 & 11340 & 3 \\
		colon & 62 & 2000 & 2 \\
		\bottomrule
	\end{tabular}%
\end{table}%

\subsection{Numerical Results}
The median results of the objective values are reported in~\Cref{tab:SSC_obj}. It is found that on the majority of the test instances MMES is obviously superior to GSC. An interesting observation is that MMES surpasses GSC for large or moderate $\beta$ values while GSC achieves better results only for small $\beta$ values. The reason is that the objective of the sparse SC is a combination of two subproblems, namely 1) the SC problem $\min Tr[\bm{U}^T \bm{L} \bm{U}]$ which is relative easy due to the smoothness and 2) the sparsity satisfaction problem $\min |\bm{U} \bm{U}^T|_1$ which is much harder due to the non-differentiability. When $\beta$ is small enough, the sparse SC problem will degenerate to be smooth and the trust-region framework of GSC is guaranteed to solve the problem efficiently. Contrarily, when $\beta$ is large, the objective is dominated by its non-differentiable part and the use of subgradient may produce a wrong step deviating from the optimal direction in the subproblem. In this case, MMES exhibits better performance since it does not rely on the differentiability of the objective function.

\begin{table*}[htbp]
	\centering
	\scriptsize
	\caption{Median objective values of the sparse spectral clustering performed on 18 real-world data sets. The better results are shown with dark background.}
	\setlength{\tabcolsep}{5pt}
	\label{tab:SSC_obj}
\begin{tabular}{ccccrrrrc}
\toprule
  & $\beta$   & MMES & GSC &   &   & $\beta$  & MMES & GSC \\
\midrule
orlraws10P & 0.01 & \cellcolor[rgb]{ .627,  .627,  .627} $1.27018E+00$  & $1.55786E+00$ $\bullet$ &   & GLIOMA & 0.01 & \cellcolor[rgb]{ .627,  .627,  .627} $6.84202E-01$  & $7.66707E-01$ $\bullet$ \\
  & 0.005 & \cellcolor[rgb]{ .627,  .627,  .627} $7.30588E-01$  & $8.06160E-01$ $\bullet$ &   &   & 0.005 & \cellcolor[rgb]{ .627,  .627,  .627} $4.72311E-01$  & $4.74266E-01$ $\bullet$ \\
  & 0.001 & \cellcolor[rgb]{ .627,  .627,  .627} $3.20149E-01$  & $3.20612E-01$ $\bullet$ &   &   & 0.001 & \cellcolor[rgb]{ .627,  .627,  .627} $2.67244E-01$  & $2.67246E-01$ $\bullet$ \\
  & 0.0005 & \cellcolor[rgb]{ .627,  .627,  .627} $2.65394E-01$  & $2.65443E-01$ $\bullet$ &   &   & 0.0005 & \cellcolor[rgb]{ .627,  .627,  .627} $2.38068E-01$  & $2.38068E-01$ $\bullet$ \\
  & 0.0001 & \cellcolor[rgb]{ .627,  .627,  .627} $2.20797E-01$  & $2.20798E-01$ $\bullet$ &   &   & 0.0001 & $2.14111E-01$  & \cellcolor[rgb]{ .627,  .627,  .627} $2.14111E-01$ $\dagger$ \\
  & 0.00005 & \cellcolor[rgb]{ .627,  .627,  .627} $2.15156E-01$  & $2.15157E-01$ $\bullet$ &   &   & 0.00005 & $2.11078E-01$  & \cellcolor[rgb]{ .627,  .627,  .627} $2.11078E-01$ $\dagger$ \\
  & 0.00001 & $2.10628E-01$  & \cellcolor[rgb]{ .627,  .627,  .627} $2.10628E-01$ $\circ$ &   &   & 0.00001 & $2.08645E-01$  & \cellcolor[rgb]{ .627,  .627,  .627} $2.08645E-01$ $\circ$ \\
\midrule
pixraw10P & 0.01 & \cellcolor[rgb]{ .627,  .627,  .627} $1.04366E+00$  & $1.38800E+00$ $\bullet$ &   & leukemia & 0.01 & \cellcolor[rgb]{ .627,  .627,  .627} $4.53164E-01$  & $5.66178E-01$ $\bullet$ \\
  & 0.005 & \cellcolor[rgb]{ .627,  .627,  .627} $5.56284E-01$  & $6.88924E-01$ $\bullet$ &   &   & 0.005 & \cellcolor[rgb]{ .627,  .627,  .627} $2.99835E-01$  & $3.29962E-01$ $\bullet$ \\
  & 0.001 & \cellcolor[rgb]{ .627,  .627,  .627} $1.20428E-01$  & $1.22549E-01$ $\bullet$ &   &   & 0.001 & \cellcolor[rgb]{ .627,  .627,  .627} $1.01505E-01$  & $1.01570E-01$ $\bullet$ \\
  & 0.0005 & \cellcolor[rgb]{ .627,  .627,  .627} $7.04549E-02$  & $7.04629E-02$ $\dagger$ &   &   & 0.0005 & \cellcolor[rgb]{ .627,  .627,  .627} $6.72585E-02$  & $6.72590E-02$ $\bullet$ \\
  & 0.0001 & $3.00985E-02$  & \cellcolor[rgb]{ .627,  .627,  .627} $3.00967E-02$ $\circ$ &   &   & 0.0001 & $3.74675E-02$  & \cellcolor[rgb]{ .627,  .627,  .627} $3.74675E-02$ $\circ$ \\
  & 0.00005 & $2.50245E-02$  & \cellcolor[rgb]{ .627,  .627,  .627} $2.50236E-02$ $\dagger$ &   &   & 0.00005 & $3.35935E-02$  & \cellcolor[rgb]{ .627,  .627,  .627} $3.35935E-02$ $\dagger$ \\
  & 0.00001 & $2.09593E-02$  & \cellcolor[rgb]{ .627,  .627,  .627} $2.09592E-02$ $\circ$ &   &   & 0.00001 & $3.04703E-02$  & \cellcolor[rgb]{ .627,  .627,  .627} $3.04703E-02$ $\circ$ \\
\midrule
Prostate-GE & 0.01 & \cellcolor[rgb]{ .627,  .627,  .627} $4.21246E-01$  & $6.45708E-01$ $\bullet$ &   & lymphoma & 0.01 & \cellcolor[rgb]{ .627,  .627,  .627} $1.83488E+00$  & $2.08008E+00$ $\bullet$ \\
  & 0.005 & \cellcolor[rgb]{ .627,  .627,  .627} $2.52351E-01$  & $3.83952E-01$ $\bullet$ &   &   & 0.005 & \cellcolor[rgb]{ .627,  .627,  .627} $1.33092E+00$  & $1.39785E+00$ $\bullet$ \\
  & 0.001 & \cellcolor[rgb]{ .627,  .627,  .627} $9.07137E-02$  & $1.02208E-01$ $\bullet$ &   &   & 0.001 & \cellcolor[rgb]{ .627,  .627,  .627} $8.59514E-01$  & $8.59594E-01$ $\bullet$ \\
  & 0.0005 & \cellcolor[rgb]{ .627,  .627,  .627} $5.53935E-02$  & $5.81033E-02$ $\bullet$ &   &   & 0.0005 & \cellcolor[rgb]{ .627,  .627,  .627} $7.93006E-01$  & $7.93014E-01$ $\bullet$ \\
  & 0.0001 & \cellcolor[rgb]{ .627,  .627,  .627} $2.22608E-02$  & $2.22608E-02$ $\bullet$ &   &   & 0.0001 & \cellcolor[rgb]{ .627,  .627,  .627} $7.37019E-01$  & $7.37019E-01$ $\bullet$ \\
  & 0.00005 & $1.65801E-02$  & \cellcolor[rgb]{ .627,  .627,  .627} $1.65801E-02$ $\dagger$ &   &   & 0.00005 & $7.29787E-01$  & \cellcolor[rgb]{ .627,  .627,  .627} $7.29787E-01$ $\circ$ \\
  & 0.00001 & $1.19131E-02$  & \cellcolor[rgb]{ .627,  .627,  .627} $1.19131E-02$ $\circ$ &   &   & 0.00001 & $7.23958E-01$  & \cellcolor[rgb]{ .627,  .627,  .627} $7.23958E-01$ $\circ$ \\
\midrule
TOX-171 & 0.01 & \cellcolor[rgb]{ .627,  .627,  .627} $7.55167E-01$  & $1.22296E+00$ $\bullet$ &   & nci9 & 0.01 & \cellcolor[rgb]{ .627,  .627,  .627} $2.93831E+00$  & $2.99212E+00$ $\bullet$ \\
  & 0.005 & \cellcolor[rgb]{ .627,  .627,  .627} $5.01975E-01$  & $7.34003E-01$ $\bullet$ &   &   & 0.005 & \cellcolor[rgb]{ .627,  .627,  .627} $2.50472E+00$  & $2.51044E+00$ $\bullet$ \\
  & 0.001 & \cellcolor[rgb]{ .627,  .627,  .627} $2.52695E-01$  & $2.76812E-01$ $\bullet$ &   &   & 0.001 & \cellcolor[rgb]{ .627,  .627,  .627} $2.10050E+00$  & $2.10054E+00$ $\bullet$ \\
  & 0.0005 & \cellcolor[rgb]{ .627,  .627,  .627} $1.89093E-01$  & $1.89843E-01$ $\dagger$ &   &   & 0.0005 & \cellcolor[rgb]{ .627,  .627,  .627} $2.04605E+00$  & $2.04605E+00$ $\bullet$ \\
  & 0.0001 & \cellcolor[rgb]{ .627,  .627,  .627} $1.13652E-01$  & $1.13657E-01$ $\bullet$ &   &   & 0.0001 & \cellcolor[rgb]{ .627,  .627,  .627} $2.00196E+00$  & $2.00196E+00$ $\bullet$ \\
  & 0.00005 & \cellcolor[rgb]{ .627,  .627,  .627} $1.03212E-01$  & $1.03212E-01$ $\bullet$ &   &   & 0.00005 & $1.99642E+00$  & \cellcolor[rgb]{ .627,  .627,  .627} $1.99642E+00$ $\circ$ \\
  & 0.00001 & $9.46874E-02$  & \cellcolor[rgb]{ .627,  .627,  .627} $9.46874E-02$ $\circ$ &   &   & 0.00001 & $1.99198E+00$  & \cellcolor[rgb]{ .627,  .627,  .627} $1.99198E+00$ $\circ$ \\
\midrule
warpAR10P & 0.01 & \cellcolor[rgb]{ .627,  .627,  .627} $2.27204E+00$  & $2.59466E+00$ $\bullet$ &   & SMK-CAN-187 & 0.01 & \cellcolor[rgb]{ .627,  .627,  .627} $5.36344E-01$  & $1.08540E+00$ $\bullet$ \\
  & 0.005 & \cellcolor[rgb]{ .627,  .627,  .627} $1.71013E+00$  & $1.76380E+00$ $\bullet$ &   &   & 0.005 & \cellcolor[rgb]{ .627,  .627,  .627} $3.03356E-01$  & $6.30792E-01$ $\bullet$ \\
  & 0.001 & \cellcolor[rgb]{ .627,  .627,  .627} $1.07139E+00$  & $1.07173E+00$ $\bullet$ &   &   & 0.001 & \cellcolor[rgb]{ .627,  .627,  .627} $1.25723E-01$  & $1.76207E-01$ $\bullet$ \\
  & 0.0005 & \cellcolor[rgb]{ .627,  .627,  .627} $9.77633E-01$  & $9.77656E-01$ $\dagger$ &   &   & 0.0005 & \cellcolor[rgb]{ .627,  .627,  .627} $9.39594E-02$  & $9.75341E-02$ $\bullet$ \\
  & 0.0001 & $8.96073E-01$  & \cellcolor[rgb]{ .627,  .627,  .627} $8.96073E-01$ $\dagger$ &   &   & 0.0001 & \cellcolor[rgb]{ .627,  .627,  .627} $4.09850E-02$  & $4.09850E-02$ $\bullet$ \\
  & 0.00005 & $8.85327E-01$  & \cellcolor[rgb]{ .627,  .627,  .627} $8.85324E-01$ $\circ$ &   &   & 0.00005 & $3.07207E-02$  & \cellcolor[rgb]{ .627,  .627,  .627} $3.07207E-02$ $\circ$ \\
  & 0.00001 & $8.76608E-01$  & \cellcolor[rgb]{ .627,  .627,  .627} $8.76602E-01$ $\circ$ &   &   & 0.00001 & $2.21933E-02$  & \cellcolor[rgb]{ .627,  .627,  .627} $2.21933E-02$ $\circ$ \\
\midrule
Yale & 0.01 & \cellcolor[rgb]{ .627,  .627,  .627} $4.15550E+00$  & $4.65828E+00$ $\bullet$ &   & arcene & 0.01 & \cellcolor[rgb]{ .627,  .627,  .627} $3.38271E-01$  & $9.02705E-01$ $\bullet$ \\
  & 0.005 & \cellcolor[rgb]{ .627,  .627,  .627} $3.16563E+00$  & $3.32454E+00$ $\bullet$ &   &   & 0.005 & \cellcolor[rgb]{ .627,  .627,  .627} $1.96669E-01$  & $4.29830E-01$ $\bullet$ \\
  & 0.001 & \cellcolor[rgb]{ .627,  .627,  .627} $2.15737E+00$  & $2.16496E+00$ $\dagger$ &   &   & 0.001 & \cellcolor[rgb]{ .627,  .627,  .627} $8.17728E-02$  & $1.27606E-01$ $\bullet$ \\
  & 0.0005 & \cellcolor[rgb]{ .627,  .627,  .627} $2.00948E+00$  & $2.01013E+00$ $\dagger$ &   &   & 0.0005 & \cellcolor[rgb]{ .627,  .627,  .627} $5.80182E-02$  & $7.81465E-02$ $\bullet$ \\
  & 0.0001 & $1.88850E+00$  & \cellcolor[rgb]{ .627,  .627,  .627} $1.88847E+00$ $\circ$ &   &   & 0.0001 & \cellcolor[rgb]{ .627,  .627,  .627} $1.28815E-02$  & $1.46339E-02$ $\bullet$ \\
  & 0.00005 & $1.87310E+00$  & \cellcolor[rgb]{ .627,  .627,  .627} $1.87305E+00$ $\circ$ &   &   & 0.00005 & \cellcolor[rgb]{ .627,  .627,  .627} $6.07431E-03$  & $7.27536E-03$ $\bullet$ \\
  & 0.00001 & $1.86068E+00$  & \cellcolor[rgb]{ .627,  .627,  .627} $1.86066E+00$ $\circ$ &   &   & 0.00001 & \cellcolor[rgb]{ .627,  .627,  .627} $1.29086E-03$  & $1.31193E-03$ $\dagger$ \\
\midrule
ALLAML & 0.01 & \cellcolor[rgb]{ .627,  .627,  .627} $4.72313E-01$  & $5.85180E-01$ $\bullet$ &   & Carcinom & 0.01 & \cellcolor[rgb]{ .627,  .627,  .627} $2.28349E+00$  & $2.89216E+00$ $\bullet$ \\
  & 0.005 & \cellcolor[rgb]{ .627,  .627,  .627} $2.88056E-01$  & $3.46486E-01$ $\bullet$ &   &   & 0.005 & \cellcolor[rgb]{ .627,  .627,  .627} $1.60182E+00$  & $1.83127E+00$ $\bullet$ \\
  & 0.001 & \cellcolor[rgb]{ .627,  .627,  .627} $1.10831E-01$  & $1.11142E-01$ $\bullet$ &   &   & 0.001 & \cellcolor[rgb]{ .627,  .627,  .627} $8.36383E-01$  & $8.61114E-01$ $\bullet$ \\
  & 0.0005 & \cellcolor[rgb]{ .627,  .627,  .627} $7.57444E-02$  & $7.57451E-02$ $\bullet$ &   &   & 0.0005 & \cellcolor[rgb]{ .627,  .627,  .627} $7.34524E-01$  & $7.36845E-01$ $\bullet$ \\
  & 0.0001 & \cellcolor[rgb]{ .627,  .627,  .627} $4.57507E-02$  & $4.57507E-02$ $\bullet$ &   &   & 0.0001 & \cellcolor[rgb]{ .627,  .627,  .627} $6.46545E-01$  & $6.46570E-01$ $\bullet$ \\
  & 0.00005 & \cellcolor[rgb]{ .627,  .627,  .627} $4.19256E-02$  & $4.19256E-02$ $\bullet$ &   &   & 0.00005 & \cellcolor[rgb]{ .627,  .627,  .627} $6.35026E-01$  & $6.35028E-01$ $\dagger$ \\
  & 0.00001 & $3.88493E-02$  & \cellcolor[rgb]{ .627,  .627,  .627} $3.88493E-02$ $\circ$ &   &   & 0.00001 & $6.25711E-01$  & \cellcolor[rgb]{ .627,  .627,  .627} $6.25710E-01$ $\circ$ \\
\midrule
warpPIE10P & 0.01 & \cellcolor[rgb]{ .627,  .627,  .627} $1.43408E+00$  & $2.47986E+00$ $\bullet$ &   & CLL-SUB-111 & 0.01 & \cellcolor[rgb]{ .627,  .627,  .627} $5.87469E-01$  & $8.68369E-01$ $\bullet$ \\
  & 0.005 & \cellcolor[rgb]{ .627,  .627,  .627} $8.35624E-01$  & $1.22085E+00$ $\bullet$ &   &   & 0.005 & \cellcolor[rgb]{ .627,  .627,  .627} $3.96819E-01$  & $5.00142E-01$ $\bullet$ \\
  & 0.001 & \cellcolor[rgb]{ .627,  .627,  .627} $2.87847E-01$  & $3.22830E-01$ $\bullet$ &   &   & 0.001 & $1.17443E-01$  & \cellcolor[rgb]{ .627,  .627,  .627} $1.13898E-01$ $\dagger$ \\
  & 0.0005 & \cellcolor[rgb]{ .627,  .627,  .627} $1.82623E-01$  & $1.85795E-01$ $\dagger$ &   &   & 0.0005 & \cellcolor[rgb]{ .627,  .627,  .627} $5.62751E-02$  & $5.63471E-02$ $\bullet$ \\
  & 0.0001 & $8.80997E-02$  & \cellcolor[rgb]{ .627,  .627,  .627} $8.18349E-02$ $\circ$ &   &   & 0.0001 & \cellcolor[rgb]{ .627,  .627,  .627} $1.29430E-02$  & $1.29441E-02$ $\bullet$ \\
  & 0.00005 & $7.24649E-02$  & \cellcolor[rgb]{ .627,  .627,  .627} $6.72412E-02$ $\circ$ &   &   & 0.00005 & \cellcolor[rgb]{ .627,  .627,  .627} $7.49886E-03$  & $7.49932E-03$ $\bullet$ \\
  & 0.00001 & $6.00220E-02$  & \cellcolor[rgb]{ .627,  .627,  .627} $5.64446E-02$ $\circ$ &   &   & 0.00001 & $3.12265E-03$  & \cellcolor[rgb]{ .627,  .627,  .627} $3.12263E-03$ $\circ$ \\
\midrule
GLI-85 & 0.01 & \cellcolor[rgb]{ .627,  .627,  .627} $4.41813E-01$  & $6.33051E-01$ $\bullet$ &   & colon & 0.01 & \cellcolor[rgb]{ .627,  .627,  .627} $5.33337E-01$  & $6.14039E-01$ $\bullet$ \\
  & 0.005 & \cellcolor[rgb]{ .627,  .627,  .627} $3.20914E-01$  & $4.22478E-01$ $\bullet$ &   &   & 0.005 & \cellcolor[rgb]{ .627,  .627,  .627} $3.03551E-01$  & $3.17157E-01$ $\bullet$ \\
  & 0.001 & \cellcolor[rgb]{ .627,  .627,  .627} $1.25968E-01$  & $1.26024E-01$ $\bullet$ &   &   & 0.001 & \cellcolor[rgb]{ .627,  .627,  .627} $8.59321E-02$  & $8.59333E-02$ $\bullet$ \\
  & 0.0005 & \cellcolor[rgb]{ .627,  .627,  .627} $8.47801E-02$  & $8.47805E-02$ $\bullet$ &   &   & 0.0005 & \cellcolor[rgb]{ .627,  .627,  .627} $5.61061E-02$  & $5.61061E-02$ $\bullet$ \\
  & 0.0001 & \cellcolor[rgb]{ .627,  .627,  .627} $4.76805E-02$  & $4.76805E-02$ $\bullet$ &   &   & 0.0001 & \cellcolor[rgb]{ .627,  .627,  .627} $3.15915E-02$  & $3.15915E-02$ $\bullet$ \\
  & 0.00005 & $4.28831E-02$  & \cellcolor[rgb]{ .627,  .627,  .627} $4.28831E-02$ $\circ$ &   &   & 0.00005 & \cellcolor[rgb]{ .627,  .627,  .627} $2.84815E-02$  & $2.84815E-02$ $\dagger$ \\
  & 0.00001 & $3.90254E-02$  & \cellcolor[rgb]{ .627,  .627,  .627} $3.90254E-02$ $\circ$ &   &   & 0.00001 & $2.59862E-02$  & \cellcolor[rgb]{ .627,  .627,  .627} $2.59862E-02$ $\circ$ \\
\midrule
$\bullet$ / $\circ$ / $\dagger$ &   &   & 38 / 16 / 9 &   &   &   &   & 44 / 12 / 7 \\
\bottomrule
\end{tabular}%

\end{table*}

\end{document}